\documentclass{article}
\PassOptionsToPackage{numbers, compress}{natbib}
\usepackage[main, final]{neurips_2026}

\PassOptionsToPackage{table}{xcolor} 
\usepackage[utf8]{inputenc}
\usepackage[T1]{fontenc}
\usepackage[hyphens]{url}
\usepackage{graphicx}
\usepackage{array}
\usepackage{booktabs}
\usepackage{multirow}
\usepackage{amsmath,amssymb,amsfonts}
\usepackage{amsthm}
\usepackage{mathtools}
\newtheorem{proposition}{Proposition}
\usepackage{subcaption}
\usepackage{wrapfig}
\usepackage[most]{tcolorbox}          
\usepackage{nicefrac}
\usepackage{microtype}
\usepackage{float}                    
\usepackage{xcolor}
\usepackage{arydshln}                 
\ADLinactivate                           
\usepackage{pifont}
\usepackage{hyperref}
\hypersetup{pdftitle={Calibrating LLMs with Semantic-level Reward},pdfauthor={Fengfei Yu, Ruijia Niu, Dongxia Wu, Yian Ma, Rose Yu}}
\providecommand{\cmark}{\ding{51}}
\providecommand{\xmark}{\ding{55}}

\title{Calibrating LLMs with Semantic-level Reward}

\author{%
  Fengfei Yu\textsuperscript{1}\,\thanks{Equal contribution.\quad \textsuperscript{\(\dagger\)}Corresponding author: Rose Yu (\texttt{roseyu@ucsd.edu}).\quad \textsuperscript{\(\ddagger\)}Work done during post-doc at Stanford.}\quad
  Ruijia Niu\textsuperscript{1,2}\,\footnotemark[1]\quad
  Dongxia Wu\textsuperscript{3,\(\ddagger\)}\quad
  Yian Ma\textsuperscript{2,1}\quad
  Rose Yu\textsuperscript{1,2,\(\dagger\)}\\
  \textsuperscript{1}Department of Computer Science and Engineering, University of California San Diego\\
  \textsuperscript{2}Hal{\i}c{\i}o\u{g}lu Data Science Institute, University of California San Diego\\
  \textsuperscript{3}Mohamed bin Zayed University of Artificial Intelligence
}

\begin{document}

\maketitle

\begin{abstract}
As large language models (LLMs) are deployed in consequential settings such as medical question answering and legal reasoning, the ability to estimate when their outputs are likely to be correct is essential for safe and reliable use and requires well-calibrated uncertainty. Standard reinforcement learning with verifiable rewards (RLVR) trains models with a binary correctness reward that is indifferent to confidence, providing no penalty for confident but wrong predictions and thereby degrading calibration. Recent work addresses this by training models to produce verbalized confidence scores alongside answers and rewarding agreement with correctness. However, verbalized confidence is tied to surface text and thus varies across textual variations with the same semantic meaning. We propose \textbf{Calibration with Semantic Reward (CSR)}, a framework that calibrates language models directly in semantic space without a verbalized confidence interface. CSR is a training-time policy intervention: instead of eliciting a confidence score, it shapes the policy so that a fixed post-hoc semantic-uncertainty estimator becomes reliable. CSR combines the correctness reward with a novel semantic calibration reward that encourages exploitation among correct rollouts by promoting semantic agreement, and exploration among incorrect ones by discouraging spurious consistency. Across three model families and four open-ended QA datasets, one in-distribution and three out-of-distribution, CSR lowers four-dataset average ECE by up to 57\% and raises average AUROC by up to 12\% over the base model, and by up to 45\% and 23\% over the verbalized-confidence baselines. The code is available at \url{https://github.com/Rose-STL-Lab/Semantic_Calibration}.
\end{abstract}

\section{Introduction}
Reinforcement learning with verifiable rewards (RLVR) has become the dominant training paradigm for reasoning language models, achieving strong performance across mathematical reasoning and multi-step inference~\citep{lambert2025tulu, guo2025deepseek, wen2025reinforcement}. The standard RLVR objective uses a binary correctness reward: a rollout receives a positive signal when the generated answer matches the ground truth and zero otherwise. While effective for improving accuracy, this formulation does not account for model confidence in the generated answer. A model that guesses incorrectly with high confidence receives the same signal as one that abstains. In high-stakes deployment settings such as medical question answering and legal reasoning, a reliable system must not only produce correct answers but also express calibrated uncertainty to support sound downstream decisions. 

Empirically, RL training with binary correctness rewards is known to worsen calibration and increase hallucination rates~\citep{damani2025beyond}. Recent work addresses this by training models to express confidence alongside each answer, as an explicit score or a hedged utterance, and optimizing its agreement with empirical correctness through RL or supervised losses. Methods such as Rewarding Doubt (RD)~\citep{bani2025rewarding}, RLCR~\citep{damani2025beyond}, SaySelf~\citep{xu2024sayself}, ConfTuner~\citep{li2025conftuner}, and LACIE~\citep{stengel2024lacie} operate within this paradigm, attaching the training signal to a verbalized confidence that the model generates when prompted.

\begin{figure*}[t]
\centering
\begin{subfigure}[t]{0.32\linewidth}
\centering
\includegraphics[width=\linewidth]{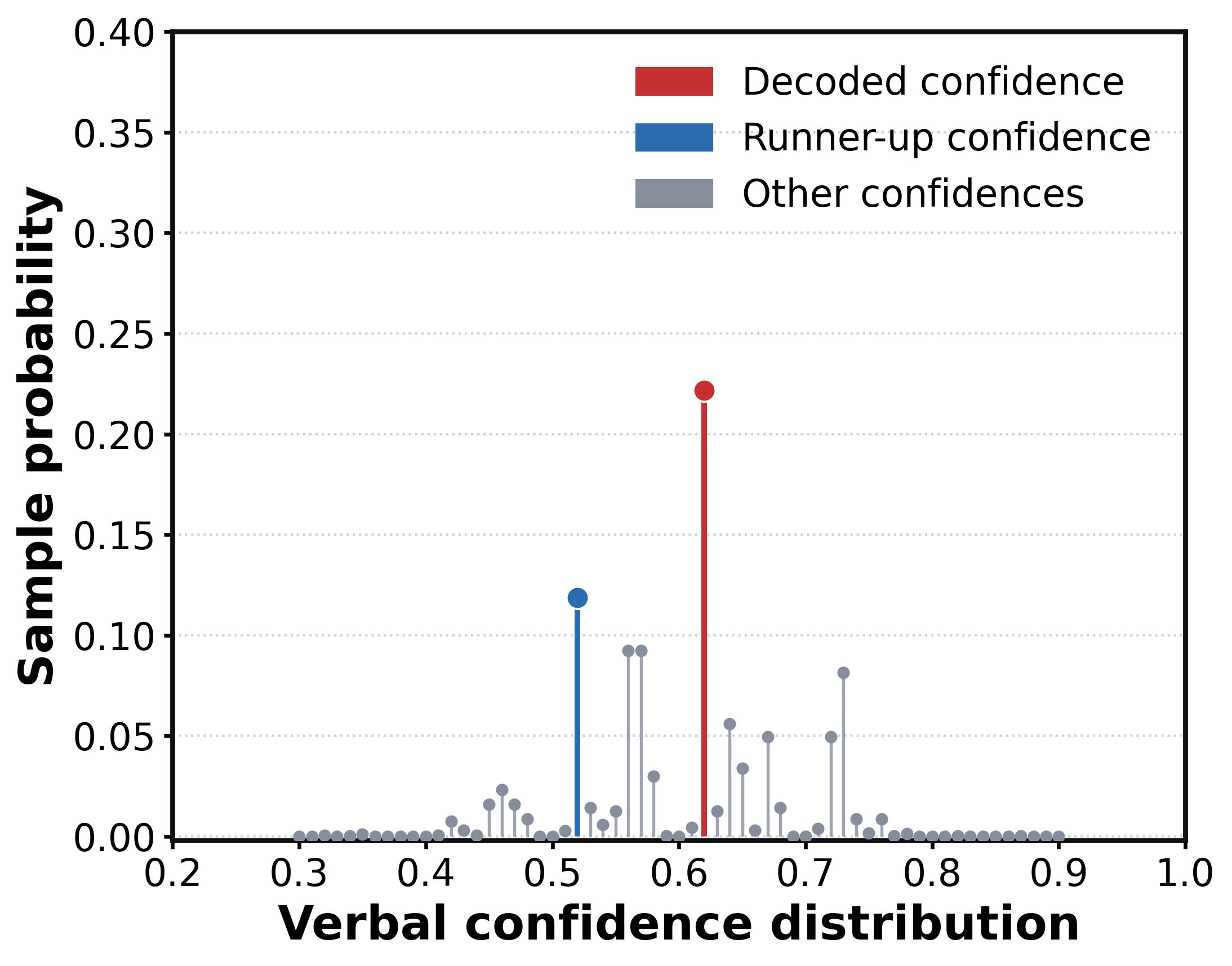}
\caption{Single-sample distribution}\label{fig:fig1_motivation_single}
\end{subfigure}
\hfill
\begin{subfigure}[t]{0.32\linewidth}
\centering
\includegraphics[width=\linewidth]{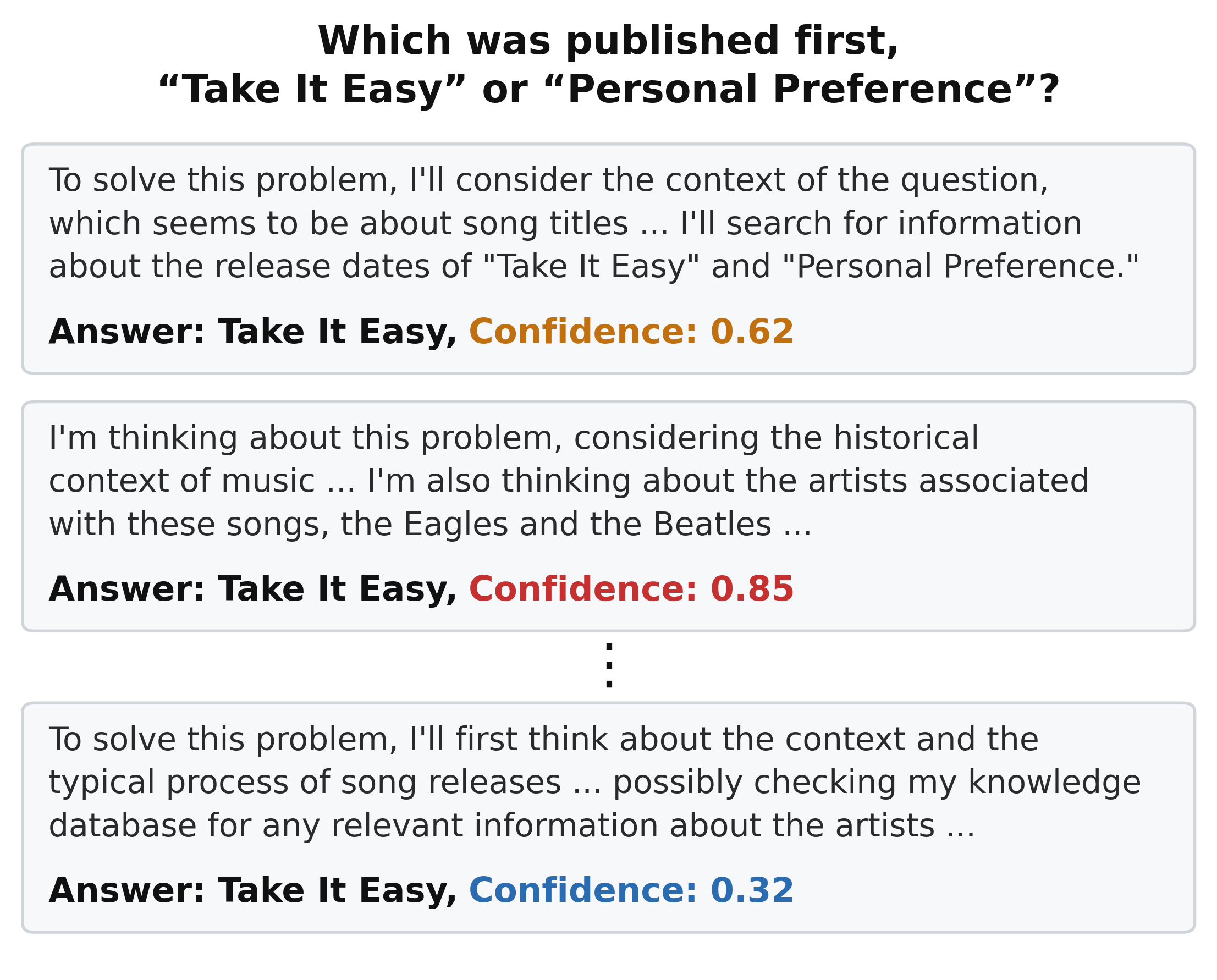}
\caption{Inconsistency}\label{fig:fig1_motivation_cases}
\end{subfigure}
\hfill
\begin{subfigure}[t]{0.32\linewidth}
\centering
\includegraphics[width=\linewidth]{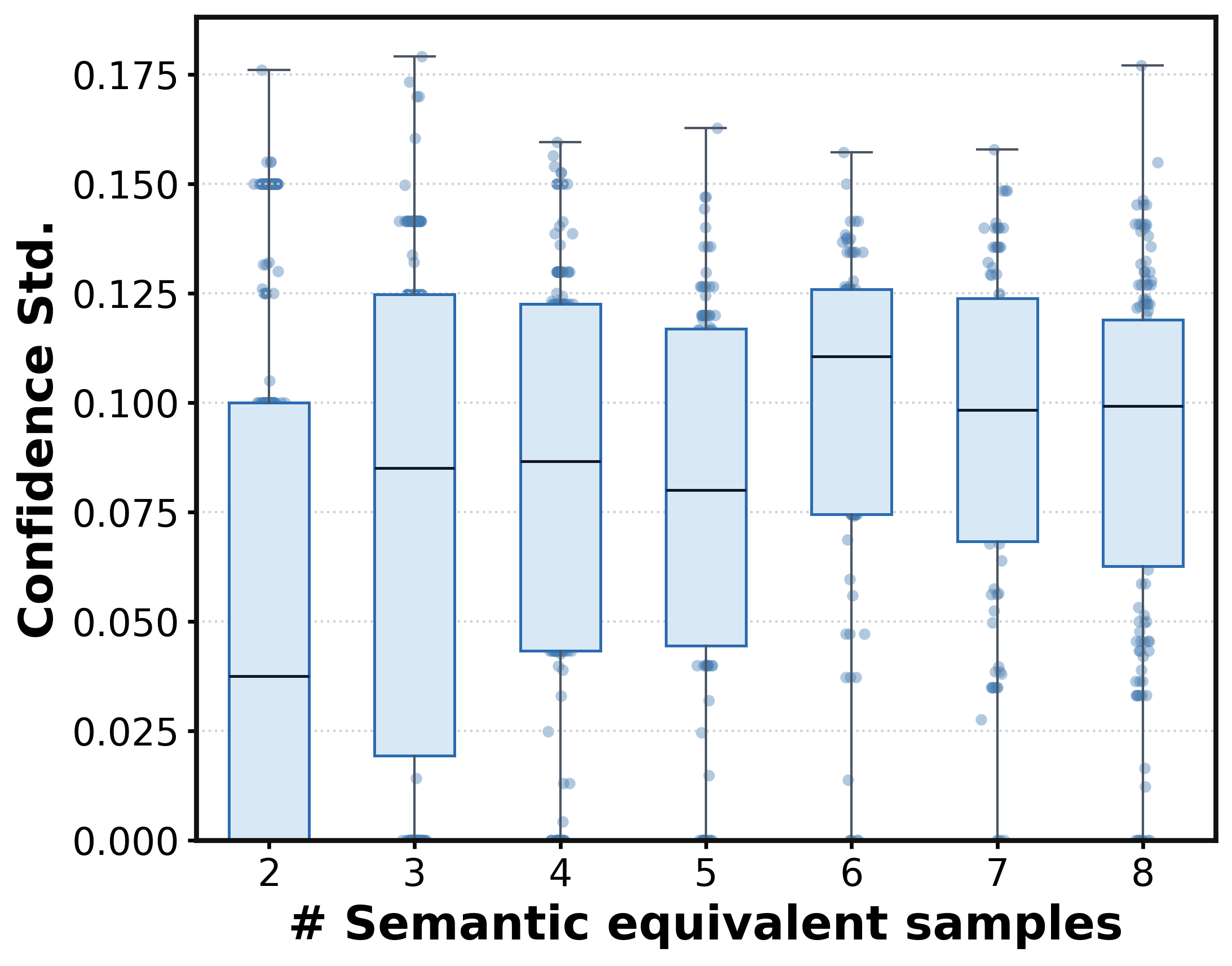}
\caption{Variability}\label{fig:fig1_motivation_variance}
\end{subfigure}
\caption{Verbalized confidence is not always stable. (a)~For a single greedy-decoded answer, the decoded confidence is uncertain, with alternative values appearing with substantial probability. (b)~Multiple responses expressing the same answer yield inconsistent confidence scores. (c)~Within groups of semantically equivalent rollouts, verbalized confidence still varies. All panels use Llama-3.1-8B-Instruct fine-tuned with RLCR\@. Setup details are provided in Appendix~\ref{app:exp_setup}.}
\label{fig:fig1_mismatch_separation}
\end{figure*}

However, existing work has identified verbalized confidence as an unreliable proxy for model uncertainty~\citep{xiong2023can, groot2024overconfidence}. The reported score depends on the answer text and prompt format rather than the distribution of meanings produced under free sampling. Fig.~\ref{fig:fig1_mismatch_separation} illustrates this under RLCR\@. Panel (a) shows that even for a single greedy answer the decoded confidence is not stable. Panel (b) shows that responses giving the same answer report inconsistent scores, so verbalized confidence varies across semantically equivalent outputs. Panel (c) shows the same effect at scale: among rollouts grouped by semantic agreement, confidence still varies widely and fails to track the degree of agreement. Together, these observations suggest that existing methods calibrate a reporting interface rather than the model's underlying uncertainty over answer meanings.

In this work, we propose to calibrate language models directly in semantic space, bypassing the verbalized confidence interface. We frame calibration in terms of \emph{semantic-level uncertainty}: the distribution of meanings the model produces when sampled repeatedly from the same question. Semantic agreement is informative of correctness when correct answers form one concentrated semantic cluster and incorrect answers are dispersed. A correctness reward alone does not produce this structure, because it gives every incorrect rollout the same zero reward whether or not the wrong answers agree with each other. The model can thus learn to repeat one wrong answer, which then appears as confident as a correct one. We introduce \textbf{Calibration with Semantic Reward (CSR)}, a fine-tuning framework that trains the policy toward this structure: alongside the correctness reward, a semantic calibration reward, added with a scheduled weight, rewards each correct rollout for agreeing with its siblings and each incorrect rollout for disagreeing with them. Confidence is then read off the answer distribution post hoc with semantic entropy~\citep{farquhar2024detecting}, without any confidence-reporting prompt, and we show that it becomes both better calibrated (ECE) and more discriminative (AUROC). This confidence can be used directly to abstain on questions the model is likely to get wrong, without changing what the model outputs.

To summarize, our contributions include:
\begin{itemize}
\item \textbf{Semantic calibration as a training target}: instead of eliciting a confidence score, we train the policy so that the confidence derived from its semantic answer distribution becomes better calibrated, providing a principled alternative to verbalized confidence-reporting interfaces.
\item \textbf{CSR}, a reinforcement learning framework that adds to the correctness reward a semantic calibration reward, which rewards agreement among correct rollouts and disagreement among incorrect ones using the GRPO rollout group at no extra sampling cost. An exact decomposition (Proposition~\ref{prop:meanfield}) shows why the correctness term is needed.
\item \textbf{Empirical validation} across three model families showing that CSR reduces four-dataset average ECE by up to 57\% and improves average AUROC by up to 12\% over the base model, and by up to 45\% and 23\% over the verbalized-confidence baselines, while maintaining competitive accuracy and generalizing to out-of-distribution datasets.
\end{itemize}

\section{Related work}
\label{sec:related_work}
\textbf{Sampling-based uncertainty estimation.}
Post-hoc methods read uncertainty off the output distribution without modifying the model or adding a confidence interface. Token-level proxies derive confidence from generation likelihoods~\citep{malinin2021uncertainty}, while sampling-based proxies treat it as dispersion across repeated draws, as in self-consistency~\citep{wang2022self}, black-box similarity measures~\citep{lin2024generating}, and semantic entropy~\citep{farquhar2024detecting}, which clusters rollouts by semantic equivalence and is therefore invariant to paraphrastic variation. Semantic Density~\citep{qiu2024semantic} and SIMBA UQ~\citep{bhattacharjya2025simba} refine sampling-based confidence scores using semantic similarities among sampled responses.

\textbf{Training for uncertainty calibration.}
Verbalized reporting is only \emph{one} interface for exposing confidence and, as discussed in the introduction, an unreliable one~\citep{xiong2023can, yang2024verbalized, groot2024overconfidence}. A line of work trains this interface directly, through RL or supervision~\citep{xu2024sayself, bani2025rewarding, damani2025beyond, li2025conftuner, stengel2024lacie, lovec2025, kapoor2024large}. Closest to us in signal, FUT~\citep{eikema2025faithfully} and the concurrent RLMF~\citep{liu2026metacognitive} derive their target from sampled-response consistency, but still train an \emph{expression} of it, a stated confidence or hedge, on top of a largely preserved answer distribution. CSR supervises no reporting interface at all: it combines each rollout's consistency with its correctness \emph{before} the group-relative advantage is formed (Sec.~\ref{sec:csr_grpo}), so consistency changes rollout credit and reshapes the distribution from which confidence is derived. A method-by-method comparison is provided in Appendix~\ref{app:related_work_full}.

\section{Preliminaries}
\label{sec:prelim}

\textbf{Reinforcement learning with verifiable rewards (RLVR).}
Let $\pi_\theta$ be a language model mapping prompts $x \in \mathcal{X}$ to outputs $Y \in \mathcal{Y}$. Given a dataset $\mathcal{D}$ of question-answer pairs $(x, Y^*)$ and a reward function $R : \mathcal{Y} \times \mathcal{Y} \to \mathbb{R}$, RLVR maximizes the expected reward under the policy~\citep{lambert2025tulu, wen2025reinforcement}:
\begin{equation}
\arg\max_\theta\; \mathbb{E}_{(x,Y^*)\sim\mathcal{D},\; Y\sim\pi_\theta(\cdot\mid x)}\left[R(Y, Y^*)\right].
\label{eq:rlvr_objective}
\end{equation}
The standard choice of $R$ is a binary correctness reward, $1$ if the generated answer $Y$ is judged equivalent to the gold answer $Y^*$ and $0$ otherwise. This reward is effective for improving accuracy but provides no calibration signal, as it is indifferent to the confidence or consistency with which the model produces an answer.

\textbf{Semantic entropy.}
Semantic entropy~\citep{farquhar2024detecting} is a sampling-based method for estimating uncertainty in language model outputs. Given a question $x$, $K$ responses $\{Y^{(1)}, \ldots, Y^{(K)}\}$ are sampled from $\pi_\theta(\cdot \mid x)$ and partitioned into semantic-equivalence classes $\{\mathcal{M}_m\}_{m=1}^{M}$ by a semantic-equivalence judge $J(Y^{(i)}, Y^{(j)}) \in \{0, 1\}$. Since pairwise verdicts need not be transitive, we follow \citet{farquhar2024detecting}: responses are visited in order, and each joins the first existing class whose first member it is judged equivalent to, or otherwise starts a new class. The probability of each semantic class is estimated empirically as
\begin{equation}
\hat{P}(\mathcal{M}_m \mid x) = \frac{1}{K} \sum_{j=1}^{K} \mathbf{1}\!\left[Y^{(j)} \in \mathcal{M}_m\right].
\label{eq:se_prob}
\end{equation}
Semantic entropy is then the Shannon entropy over the class distribution:
\begin{equation}
\mathrm{SE}(x) = -\sum_{m=1}^{M} \hat{P}(\mathcal{M}_m \mid x)\, \log \hat{P}(\mathcal{M}_m \mid x).
\label{eq:se}
\end{equation}
By operating over semantic classes rather than individual tokens or surface forms, $\mathrm{SE}$ is invariant to paraphrastic variation. High $\mathrm{SE}(x)$ indicates semantically diverse outputs and high uncertainty, and low $\mathrm{SE}(x)$ indicates consistent semantic agreement across rollouts. We use $\mathrm{SE}(x)$ as our post-hoc uncertainty estimate and convert it into a \emph{derived semantic confidence} $\hat{s}(x) = \exp\!\left(-\mathrm{SE}(x)\right)$.

\textbf{Definitions.} We follow standard uncertainty quantification (UQ) terminology~\citep{guo2017calibration,geng2024survey,lin2024generating}. $\mathrm{SE}(x)$ is an \emph{uncertainty}, a property of the question's predictive distribution, and $\hat{s}(x)$ is a question-level UQ score oriented toward correctness, which makes it a \emph{confidence}: higher values indicate that the policy's answers are more likely correct. A fixed score is \emph{calibrated} if, among questions with score $c$, the policy's accuracy is $c$, which ECE measures, and \emph{discriminative} if it ranks questions the policy answers correctly above those it does not, which AUROC measures. \emph{Semantic calibration} refers to calibration of $\hat{s}$ in this standard sense. Formal definitions are provided in Appendix~\ref{app:definitions}.

\begin{figure*}[t]
\centering
\includegraphics[width=\linewidth]{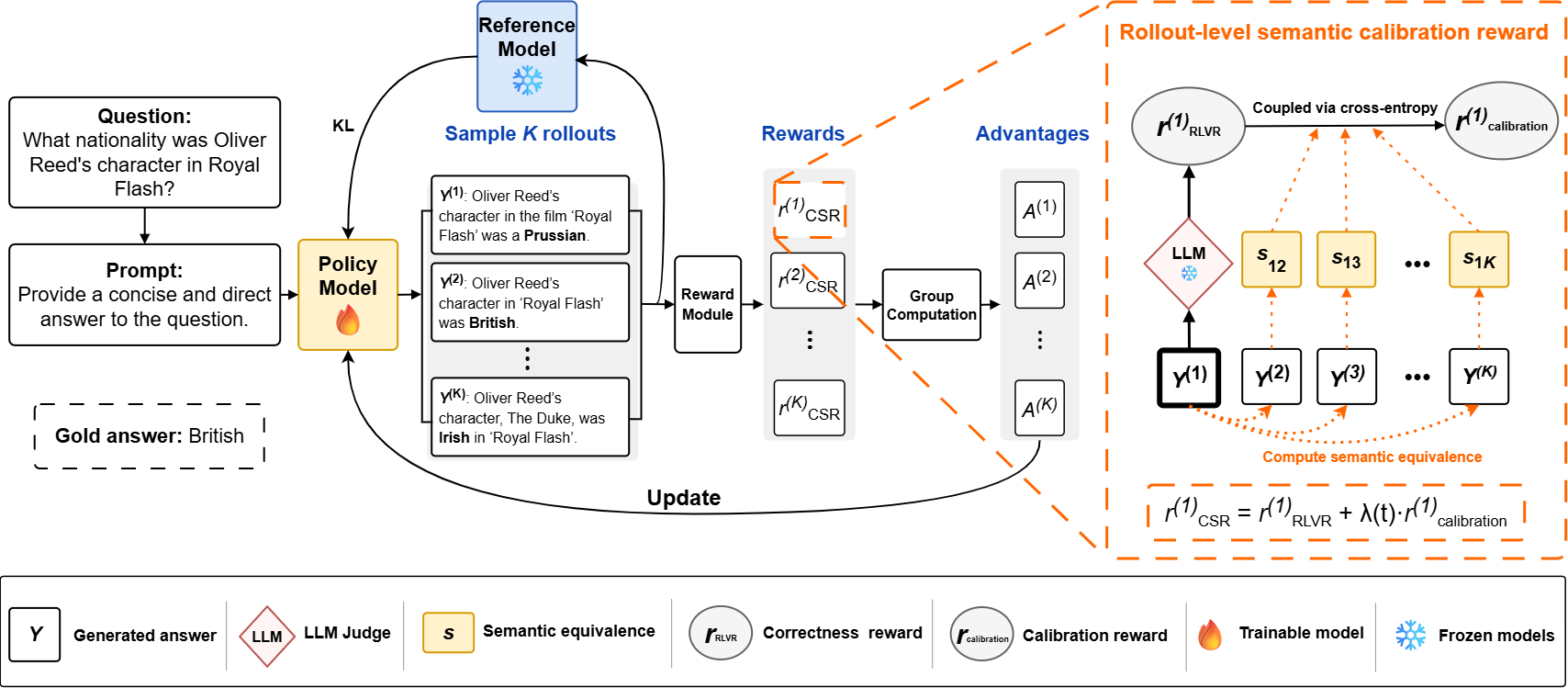}
\caption{Overview of CSR\@. For each question $x$, we draw $K$ rollouts from the policy $\pi_\theta$, score each rollout with a verifiable correctness reward $r_{\mathrm{RLVR}}$ and a semantic calibration reward $r_{\mathrm{calibration}}$ (Eq.~\eqref{eq:rlsr}), and update the policy via group-relative advantages. CSR concentrates correct rollouts into a tight semantic cluster while keeping incorrect rollouts dispersed, so that semantic agreement among sampled outputs becomes an informative signal of correctness.}
\label{fig:overview_method}
\end{figure*}

\section{Calibration with Semantic Reward (CSR)}
\label{sec:method}

We present \emph{Calibration with Semantic Reward} (CSR), a framework that calibrates a language model directly in semantic space without introducing any explicit confidence variable (Fig.~\ref{fig:overview_method}). The central idea is to make \emph{semantic agreement among sampled rollouts informative of correctness}, so that correct answers are supported by consistent rollouts while incorrect answers are not reinforced by spurious agreement. We realize this through two complementary reward signals, a \emph{verifiable correctness reward} that shifts probability mass toward correct semantic modes, and a \emph{semantic calibration reward} (Sec.~\ref{sec:rlsr}) that shapes the agreement structure across rollouts so that it tracks correctness. We combine them through a curriculum learning objective (Sec.~\ref{sec:csr_obj}) and optimize with group-relative policy optimization (Sec.~\ref{sec:csr_grpo}).

\textbf{Setup and notation.} We keep the notation of Sec.~\ref{sec:prelim}, and the gold answer $Y^*$ may be a set of acceptable answers. For a group of $K$ rollouts $Y^{(1)},\ldots,Y^{(K)}\sim\pi_\theta(\cdot\mid x)$ we write $Q_j=J(Y^{(j)},Y^*)$ for the correctness of rollout $j$ and $s_{ji}=J(Y^{(j)},Y^{(i)})$ for the agreement of rollouts $j$ and $i$. The judge $J$ can be instantiated as exact match or thresholded token-level F1 for short factual QA, or as an LLM judge for tasks with paraphrastic variation.

\subsection{Reward definitions}
\label{sec:reward_def}

\textbf{Correctness reward.}\phantomsection\label{sec:rlvr}
The verifiable correctness reward of rollout $j$ is its judged correctness,
\begin{equation}
r_{\mathrm{RLVR}}^{(j)} \;=\; Q_j \;\in\; \{0,1\}.
\label{eq:rlvr}
\end{equation}

\textbf{Semantic calibration reward.}\phantomsection\label{sec:rlsr}
We treat semantic consistency as an \emph{implicit proxy for confidence}: rollouts that concentrate on one meaning signal high semantic confidence, while rollouts that spread across meanings signal low confidence. Calibration in semantic space then amounts to making this implicit confidence track correctness, with consistency high for correct rollouts and low for incorrect ones under the same question.

We turn this principle into a per-rollout \emph{semantic calibration reward} by comparing each rollout's pairwise agreement with the rest of the group against its correctness with respect to the gold answer:
\begin{equation}
r_{\mathrm{calibration}}^{(j)}
=-\frac{1}{K-1}\sum_{i\neq j}\mathrm{CE}(s_{ji},Q_j),
\label{eq:rlsr}
\end{equation}
where $\mathrm{CE}(\cdot,\cdot)$ is binary cross-entropy and the binary agreement $s_{ji}$ is clipped to $[\epsilon,1-\epsilon]$ before the cross-entropy is computed. Each summand is maximized exactly when pairwise agreement matches correctness, so a correct rollout is rewarded for agreeing with its siblings and an incorrect rollout for disagreeing with them. For i.i.d.\ rollouts, Eq.~\eqref{eq:rlsr} thus pushes the policy-level agreement probability $p_j=\mathbb P_{Y'\sim\pi_\theta(\cdot\mid x)}(J(Y^{(j)},Y')=1\mid Y^{(j)})$ toward $Q_j$: correct answers concentrate into one semantic cluster, while incorrect rollouts that agree with one another are penalized relative to dispersed ones, which promotes exploration away from concentrated incorrect modes.
\subsection{CSR objective}
\label{sec:csr_obj}

\textbf{Limitation of the calibration reward alone.} Eq.~\eqref{eq:rlsr} shapes agreement across rollouts 
but does not raise the fraction of correct rollouts. Proposition~\ref{prop:meanfield} makes this exact.

\begin{proposition}[Exact decomposition of the calibration reward]
\label{prop:meanfield}
Let $\{Y^{(j)}\}_{j=1}^K$ be drawn i.i.d.\ from $\pi_\theta(\cdot\mid x)$, let $\alpha=K^{-1}\sum_j Q_j$ be the empirical accuracy of the group, let $\hat p_j=(K-1)^{-1}\sum_{i\neq j}s_{ji}$ be the empirical agreement rate of rollout $j$, let $p_j$ be the policy-level agreement probability defined above, and write $c_\epsilon=\log\frac{1-\epsilon}{\epsilon}$. Then, exactly,
\[
r^{(j)}_{\mathrm{calibration}}=\log\epsilon+c_\epsilon\big[Q_j\hat p_j+(1-Q_j)(1-\hat p_j)\big],
\qquad
\mathbb E\big[\hat p_j\mid Y^{(j)}\big]=p_j,
\]
and the group-average calibration reward decomposes as
\[
\bar r_{\mathrm{calibration}}
=\log\epsilon+c_\epsilon\Big[\alpha\,\bar p_{\mathcal C}+(1-\alpha)\big(1-\bar p_{\mathcal I}\big)\Big],
\]
where $\bar p_{\mathcal C}$ and $\bar p_{\mathcal I}$ average $\hat p_j$ over the correct and the incorrect rollouts (a term whose set is empty is omitted).
Consequently, when $\alpha\to0$, the reward is dominated by the second term and is maximized by driving $p_j\to0$, and when $\alpha\to1$, it is maximized by driving $p_j\to1$. Neither regime directly rewards a change in correctness, so when accuracy is low the calibration reward alone can shape agreement structure while correctness stagnates.
\end{proposition}

Early in training, when $\alpha$ is small, the calibration reward alone therefore favors maximally diverse, mostly incorrect responses, and when $\alpha$ is close to $1$ it favors collapse onto a single answer. In neither regime does it reward switching to a correct answer, so confidence can become discriminative while accuracy stagnates or falls, as in Fig.~\ref{fig:ablation_row}(a). A detailed proof is provided in Appendix~\ref{app:meanfield_proof}.

\textbf{Curriculum learning objective.} Building on the analysis above, we define the reward at step $t$ as
\begin{equation}
r^{(j)}_{\mathrm{CSR}}(t) \;=\; r_{\mathrm{RLVR}}^{(j)} \;+\; \lambda(t)\, r_{\mathrm{calibration}}^{(j)},
\label{eq:csr}
\end{equation}
where $\lambda(t)\geq 0$ controls the strength of the semantic calibration signal at training step $t$. In the implementation, $r_{\mathrm{calibration}}^{(j)}$ enters in its normalized form $Q_j\hat p_j+(1-Q_j)(1-\hat p_j)\in[0,1]$, obtained from Proposition~\ref{prop:meanfield} by dropping the constant $\log\epsilon$ and the scale $c_\epsilon$, so $\lambda(t)$ weights a reward on the same $[0,1]$ scale as $r_{\mathrm{RLVR}}$. The correctness term supplies the accuracy incentive that the calibration term lacks, while the calibration term organizes the agreement structure across rollouts. We set $\lambda(t)$ as a curriculum that increases over training, so that the calibration weight starts below the correctness weight, when accuracy is low and the calibration term cannot raise it, and ends above it as correct rollouts become common.

\subsection{Optimization with GRPO}
\label{sec:csr_grpo}

We optimize Eq.~\eqref{eq:csr} with group-relative policy optimization (GRPO)~\citep{shao2024deepseekmath}, which samples $K$ rollouts per input and computes advantages from their relative rewards within the group. This group is exactly what the calibration reward needs to measure pairwise agreement, so CSR adds no sampling overhead, only the cost of judging the rollouts. Because the calibration reward is added to the correctness reward before the group-relative normalization, it changes which rollouts receive credit within a group: among correct rollouts, more agreement earns a larger advantage, and among incorrect ones, less. We discuss the regularization effect of CSR in the GRPO setting in Appendix~\ref{app:regularization_view}.

\textbf{Alignment with post-hoc estimators.} For a sampling-based online RL algorithm such as GRPO, the rollouts drawn for each question provide a natural environment for the optimization. The sampled answers that define the reward are the same kind of samples that post-hoc estimators read to measure the uncertainty of the answer distribution. The pairwise score of Eq.~\eqref{eq:rlsr} and semantic entropy are two such uncertainty estimates, so training on one is aligned with, but not identical to, evaluating with the other, and the transfer between them is empirical. Because the estimator stays fixed, CSR improves calibration and discrimination by changing the policy, which we test with semantic entropy and four other estimators (Secs.~\ref{sec:main_results} and~\ref{sec:beyond_se}).

\section{Experiments}
\begin{table*}[t]
\centering
\small
\setlength{\tabcolsep}{3.0pt}
\renewcommand{\arraystretch}{1.0}
\caption{Calibration results across three model families and four QA datasets. HotpotQA is in distribution and the other three are out of distribution. Confidence is the derived semantic confidence for Base, RLVR, and CSR and the verbalized confidence for RD and RLCR\@. Average is the four-dataset macro-average, and Tok counts all $K$ rollouts. \textbf{Bold} marks the best and \underline{underline} the second-best value in each column within each model block.}
\label{tab:main_results}
\resizebox{\textwidth}{!}{%
\begin{tabular}{l|l c c c|c c c|c c c|c c c|c c c}
\toprule
\multicolumn{1}{l}{} & & \multicolumn{3}{c}{\textbf{In-distribution}} & \multicolumn{3}{c}{} & \multicolumn{3}{c}{\textbf{Out-of-distribution}} & \multicolumn{3}{c}{} & \multicolumn{3}{c}{}\\
\cmidrule(lr){3-5}\cmidrule(lr){6-14}
\multicolumn{1}{l}{} & & \multicolumn{3}{c}{\textbf{HotpotQA}} & \multicolumn{3}{c}{\textbf{TriviaQA}} & \multicolumn{3}{c}{\textbf{MSMARCO}} & \multicolumn{3}{c}{\textbf{NQ-Open}} & \multicolumn{3}{c}{\textbf{Average}}\\
\cmidrule(lr){3-5}\cmidrule(lr){6-8}\cmidrule(lr){9-11}\cmidrule(lr){12-14}\cmidrule(lr){15-17}
\textbf{Model} & \textbf{Method} & Acc$\uparrow$ & ECE$\downarrow$ & AUROC$\uparrow$ & Acc$\uparrow$ & ECE$\downarrow$ & AUROC$\uparrow$ & Acc$\uparrow$ & ECE$\downarrow$ & AUROC$\uparrow$ & Acc$\uparrow$ & ECE$\downarrow$ & AUROC$\uparrow$ & Tok & ECE$\downarrow$ & AUROC$\uparrow$\\
\midrule
\multirow{5}{*}{Llama}
& Base  & 0.1881 & 0.2949 & 0.7050 & \underline{0.6682} & 0.0772 & \underline{0.8423} & \textbf{0.3581} & \underline{0.1996} & 0.7062 & \underline{0.4494} & \underline{0.1186} & 0.7361 & 839.93 & 0.1726 & 0.7474\\
& RLVR  & \textbf{0.3079} & 0.2567 & 0.8089 & \textbf{0.6850} & 0.0901 & 0.8268 & \underline{0.3514} & 0.2280 & \underline{0.7371} & \textbf{0.4512} & 0.1686 & 0.6957 & 605.45 & 0.1859 & 0.7671\\
& RD    & 0.2570 & 0.1764 & \underline{0.8316} & 0.6600 & \underline{0.0746} & 0.7996 & 0.2820 & 0.3302 & 0.7298 & 0.3540 & 0.2081 & \underline{0.7464} & 249.01 & 0.1973 & \underline{0.7769}\\
& RLCR  & \underline{0.2820} & \underline{0.1757} & 0.6923 & 0.6490 & 0.0943 & 0.7337 & 0.2940 & 0.2340 & 0.6816 & 0.3950 & 0.1497 & 0.7013 & 784.71 & \underline{0.1634} & 0.7022\\
\rowcolor{gray!12} \cellcolor{white}& CSR   & 0.2700 & \textbf{0.0978} & \textbf{0.9133} & 0.6668 & \textbf{0.0670} & \textbf{0.8976} & 0.3390 & \textbf{0.1571} & \textbf{0.7594} & 0.4461 & \textbf{0.1100} & \textbf{0.7670} & 663.85 & \textbf{0.1080} & \textbf{0.8343}\\
\midrule
\multirow{5}{*}{Qwen}
& Base  & 0.2364 & 0.3050 & \underline{0.8413} & 0.5016 & 0.1881 & \underline{0.8281} & \textbf{0.3217} & 0.2945 & 0.7037 & 0.2684 & 0.2487 & \underline{0.7773} & 201.67 & 0.2591 & \underline{0.7876}\\
& RLVR  & \textbf{0.2553} & 0.3007 & 0.8365 & 0.5072 & 0.1933 & 0.8241 & 0.3201 & 0.2871 & \underline{0.7094} & \underline{0.2710} & 0.2492 & 0.7594 & 194.23 & 0.2576 & 0.7824\\
& RD    & 0.2440 & \underline{0.1613} & 0.8235 & \underline{0.5200} & \underline{0.0876} & 0.8013 & 0.2970 & \textbf{0.1008} & 0.6531 & 0.2650 & \underline{0.1380} & 0.7606 & 299.52 & \underline{0.1219} & 0.7596\\
& RLCR  & 0.2100 & 0.2370 & 0.7530 & \textbf{0.5690} & 0.0894 & 0.7293 & \underline{0.3210} & 0.2031 & 0.7026 & \textbf{0.3240} & 0.2671 & 0.6255 & 641.07 & 0.1991 & 0.7026\\
\rowcolor{gray!12} \cellcolor{white}& CSR   & \underline{0.2441} & \textbf{0.1152} & \textbf{0.9260} & 0.4914 & \textbf{0.0855} & \textbf{0.9094} & 0.3135 & \underline{0.1560} & \textbf{0.8001} & 0.2657 & \textbf{0.0866} & \textbf{0.8348} & 198.54 & \textbf{0.1108} & \textbf{0.8676}\\
\midrule
\multirow{5}{*}{Mistral}
& Base  & 0.2457 & 0.2563 & 0.7745 & \textbf{0.6776} & 0.0952 & 0.7860 & \underline{0.3753} & 0.3011 & \underline{0.7087} & 0.4188 & 0.1507 & 0.7035 & 774.69 & 0.2008 & 0.7432\\
& RLVR  & \textbf{0.3065} & 0.2533 & 0.7936 & \underline{0.6751} & 0.1126 & 0.7887 & 0.3683 & 0.2939 & 0.7003 & \underline{0.4201} & 0.1635 & 0.7079 & 634.83 & 0.2058 & 0.7476\\
& RD    & 0.2460 & \underline{0.0994} & \textbf{0.9065} & 0.5490 & 0.1061 & \underline{0.8118} & 0.1760 & \textbf{0.1879} & 0.7014 & 0.1930 & \underline{0.1395} & \textbf{0.7897} & 166.58 & \underline{0.1332} & \underline{0.8024}\\
& RLCR  & \underline{0.3040} & 0.2117 & 0.6715 & 0.6310 & \textbf{0.0611} & 0.6913 & 0.3130 & 0.2669 & 0.6529 & 0.3900 & 0.1942 & 0.6856 & 839.80 & 0.1835 & 0.6753\\
\rowcolor{gray!12} \cellcolor{white}& CSR   & 0.2981 & \textbf{0.0966} & \underline{0.8710} & 0.6703 & \underline{0.0751} & \textbf{0.8530} & \textbf{0.3773} & \underline{0.2513} & \textbf{0.7351} & \textbf{0.4269} & \textbf{0.0974} & \underline{0.7606} & 731.71 & \textbf{0.1301} & \textbf{0.8049}\\
\bottomrule
\end{tabular}%
}
\end{table*}

We ask whether CSR makes the derived semantic confidence better calibrated and more discriminative than correctness-only and verbalized-confidence training, in and out of distribution (Sec.~\ref{sec:main_results}), whether the gains hold under other confidence estimators (Sec.~\ref{sec:beyond_se}) and in selective prediction (Sec.~\ref{sec:selective_main}), whether a lexical surrogate can remove the cost of the pairwise judge (Sec.~\ref{sec:f1_vs_llm_judge}), and which design choices drive the gain (Sec.~\ref{sec:ablations}). Appendix~\ref{app:visualization} shows rollout-level examples for all five methods.

\subsection{Experimental setup}\label{sec:exp_setup}

\textbf{Datasets and models.}
Following prior work~\citep{damani2025beyond,li2025conftuner,xu2024sayself},
we use HotpotQA~\citep{yang2018hotpotqa} for training and in-distribution evaluation. To assess generalization, we further evaluate on TriviaQA~\citep{joshi2017triviaqa}, MSMARCO~\citep{bajaj2016ms}, and NQ-Open~\citep{kwiatkowski2019natural,lee2019latent}, which span trivia, web-search, and open-domain questions. We fine-tune on $10{,}000$ HotpotQA training examples, with $K=8$ rollouts per question for RLVR and CSR and a sigmoid schedule for $\lambda(t)$ from $0.8$ to $1.2$ in CSR, and evaluate using $1{,}000$ examples per dataset. We use Llama-3.1-8B-Instruct~\citep{grattafiori2024llama}, Qwen2.5-7B-Instruct~\citep{qwen2025qwen25technicalreport}, and Mistral-7B-Instruct-v0.3~\citep{jiang2023mistral7b} (Llama, Qwen, and Mistral) as base models. Training details are provided in Appendix~\ref{app:exp_setup}.

\textbf{Evaluation metrics.}
We evaluate model performance and calibration quality with accuracy, expected calibration error (ECE)~\citep{guo2017calibration}, area under the ROC curve (AUROC)~\citep{fawcett2006introduction}, and token cost (Tok). ECE and AUROC measure calibration and discrimination in the sense of Sec.~\ref{sec:prelim}, and Tok is the average number of prompt and generated tokens per question. Every method, including RD and RLCR, is scored by the same GPT-4.1 nano correctness judge. At evaluation, Base, RLVR, and CSR are scored on $K=16$ sampled rollouts, with accuracy averaged over the rollouts and a question counted as correct for AUROC when at least half of its rollouts are correct. RD and RLCR are scored on a single greedy answer. Metric definitions are provided in Appendix~\ref{app:metrics}.

\textbf{Baselines.}
We compare against four baselines. \textbf{Base} is the unmodified instruction-tuned model, and \textbf{RLVR} fine-tunes with a binary correctness reward and no calibration signal. For both, uncertainty is estimated post hoc via semantic entropy, as for CSR\@. \textbf{RD}~\citep{bani2025rewarding} and \textbf{RLCR}~\citep{damani2025beyond} are verbalized-confidence methods: RD optimizes only its verbalized confidence with a clipped logarithmic scoring rule, while RLCR combines a binary correctness reward with a Brier-score calibration term. Both require an explicit confidence-reporting interface with additional format constraints. All methods are evaluated closed-book.

\subsection{Does CSR improve calibration and discrimination?}
\label{sec:main_results}

\begin{figure*}[t]
\centering
\begin{subfigure}[t]{0.24\linewidth}
\centering
\IfFileExists{figure/main/auroc_combined_qwen_hotpotqa.png}{%
  \includegraphics[width=\linewidth]{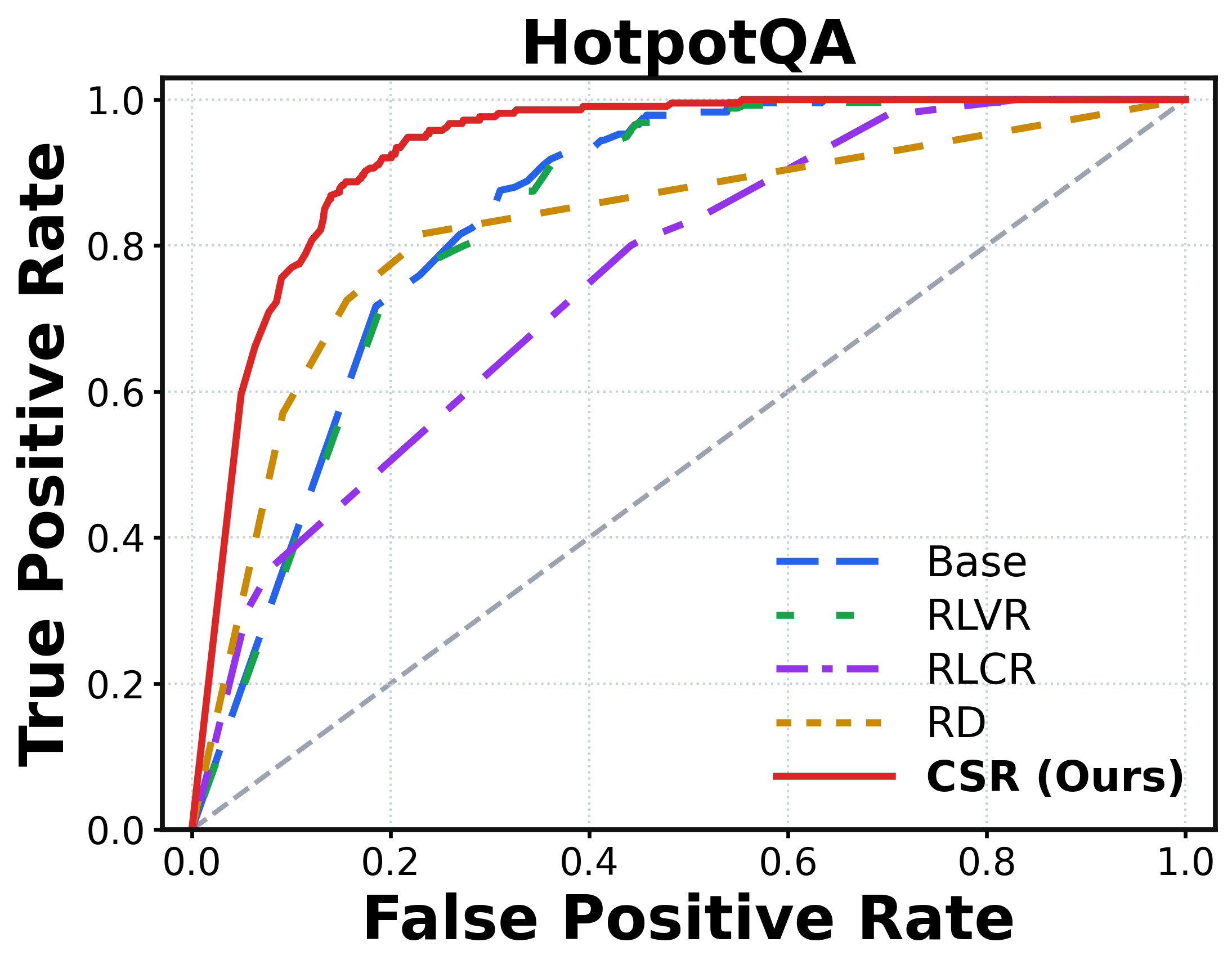}%
}{%
  \fbox{\parbox[c][1.1in][c]{\linewidth}{\centering Missing: main/auroc\_combined\_qwen\_hotpotqa.png}}%
}
\end{subfigure}
\hfill
\begin{subfigure}[t]{0.24\linewidth}
\centering
\IfFileExists{figure/main/auroc_combined_qwen_triviaqa.png}{%
  \includegraphics[width=\linewidth]{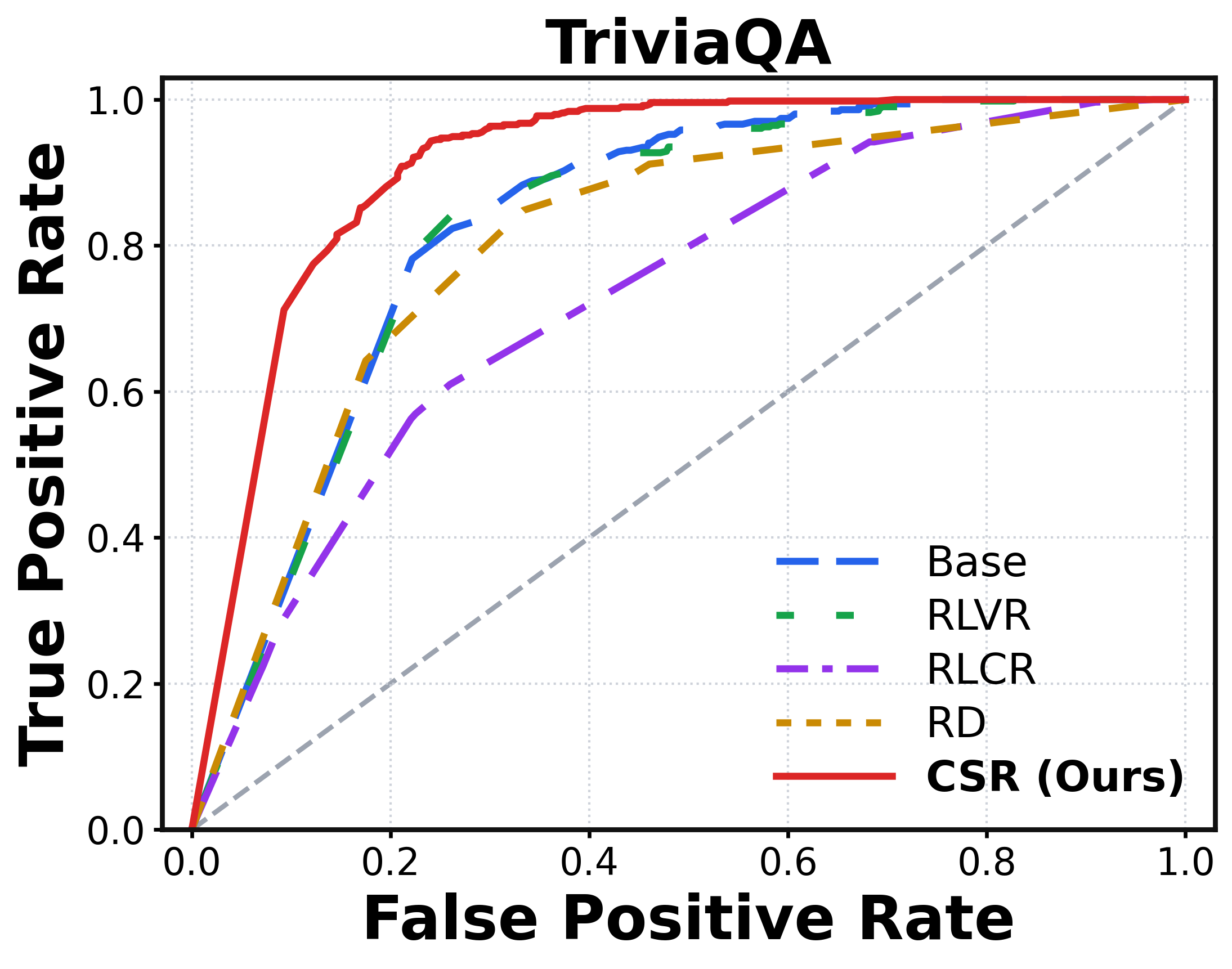}%
}{%
  \fbox{\parbox[c][1.1in][c]{\linewidth}{\centering Missing: main/auroc\_combined\_qwen\_triviaqa.png}}%
}
\end{subfigure}
\hfill
\begin{subfigure}[t]{0.24\linewidth}
\centering
\IfFileExists{figure/main/auroc_combined_qwen_msmarco.png}{%
  \includegraphics[width=\linewidth]{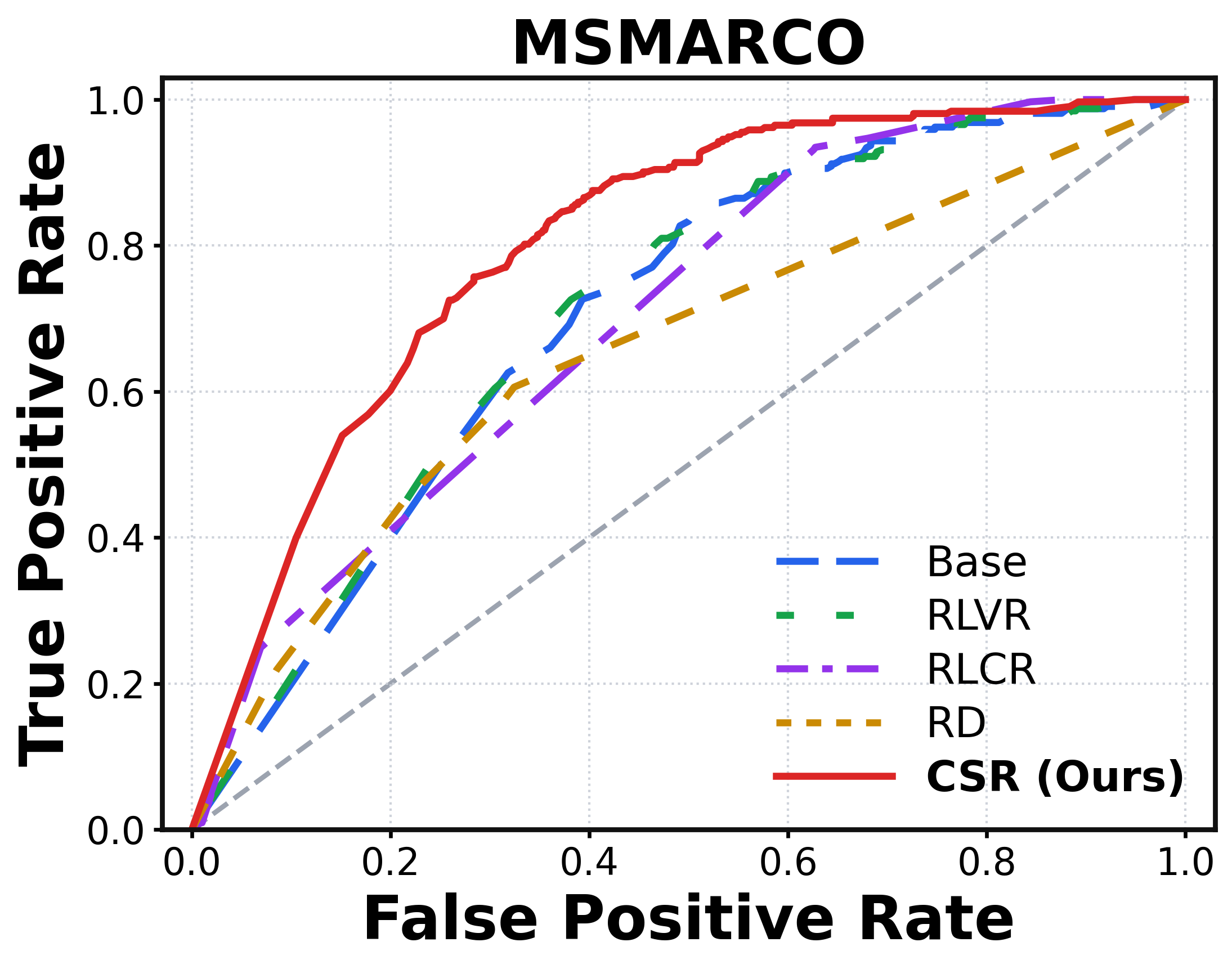}%
}{%
  \fbox{\parbox[c][1.1in][c]{\linewidth}{\centering Missing: main/auroc\_combined\_qwen\_msmarco.png}}%
}
\end{subfigure}
\hfill
\begin{subfigure}[t]{0.24\linewidth}
\centering
\IfFileExists{figure/main/auroc_combined_qwen_nq_open.png}{%
  \includegraphics[width=\linewidth]{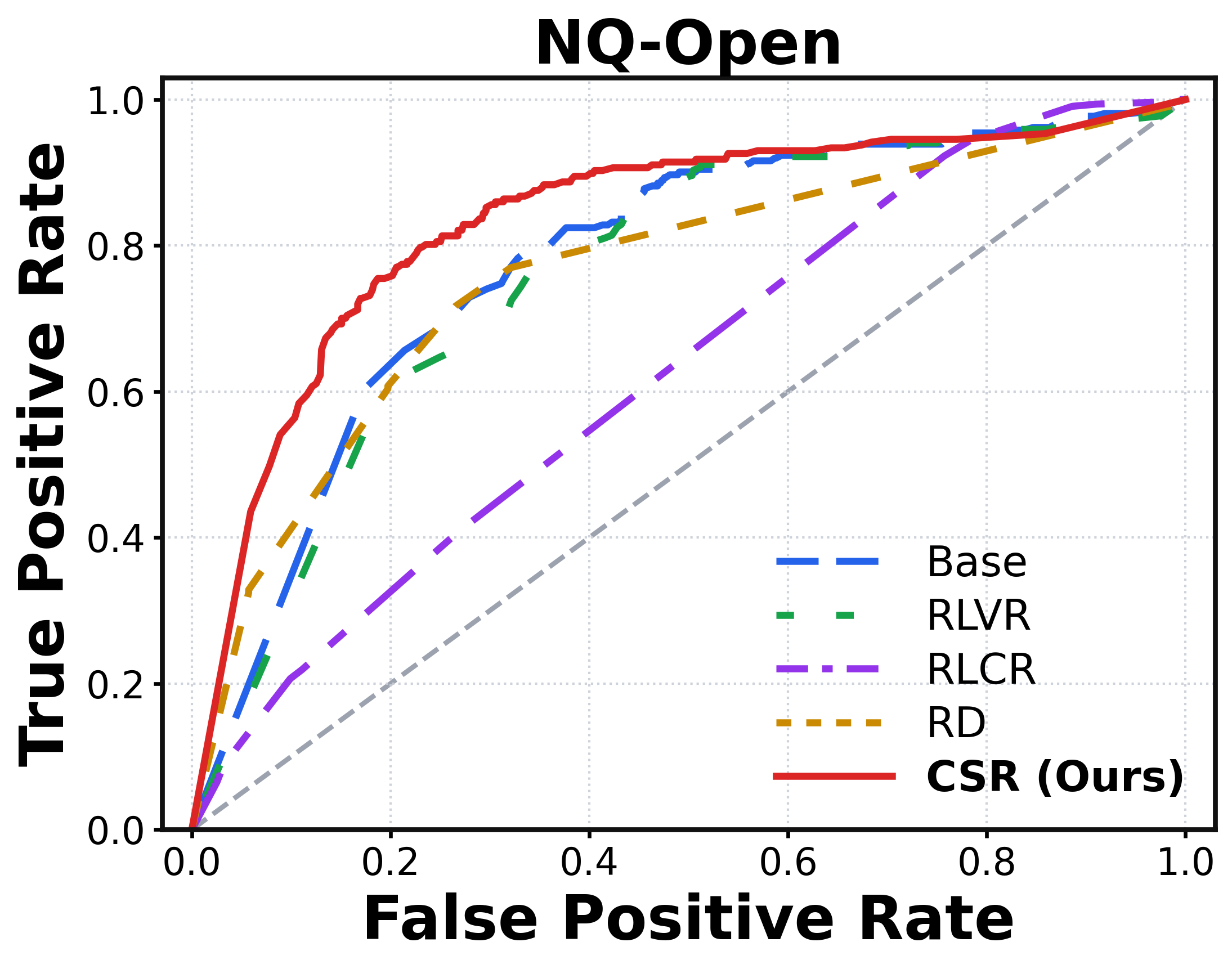}%
}{%
  \fbox{\parbox[c][1.1in][c]{\linewidth}{\centering Missing: main/auroc\_combined\_qwen\_nq\_open.png}}%
}
\end{subfigure}

\caption{ROC curves on Qwen for HotpotQA (in-distribution) and three OOD datasets. Confidence is the derived semantic confidence for Base, RLVR, and CSR and the verbalized confidence for RD and RLCR\@. CSR has the highest AUROC in all four panels.}
\label{fig:sementropy_reliability}
\end{figure*}

\begin{figure*}[t]
\centering
\begin{subfigure}{\linewidth}
\centering
\includegraphics[width=0.18\linewidth]{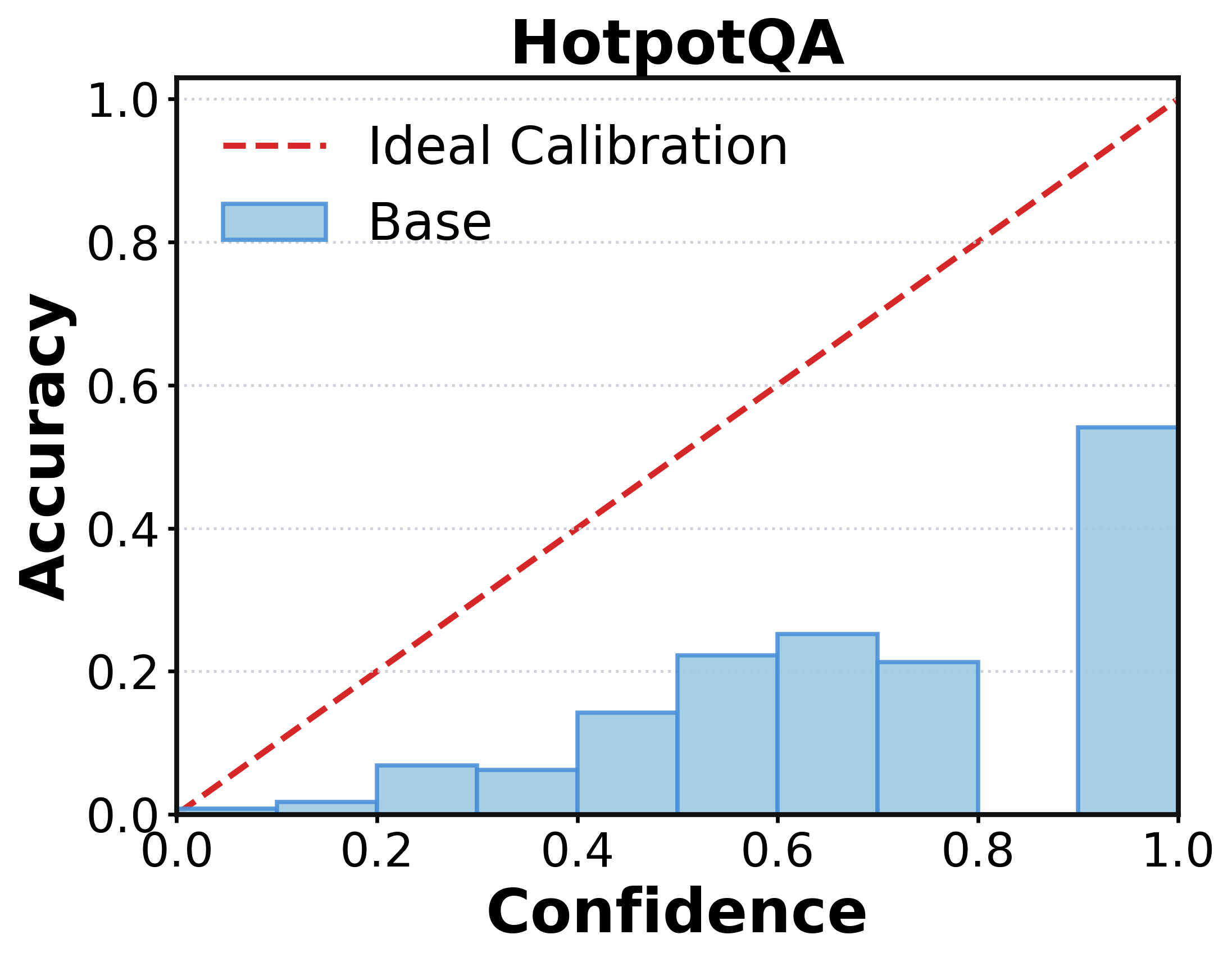}\hfill
\includegraphics[width=0.18\linewidth]{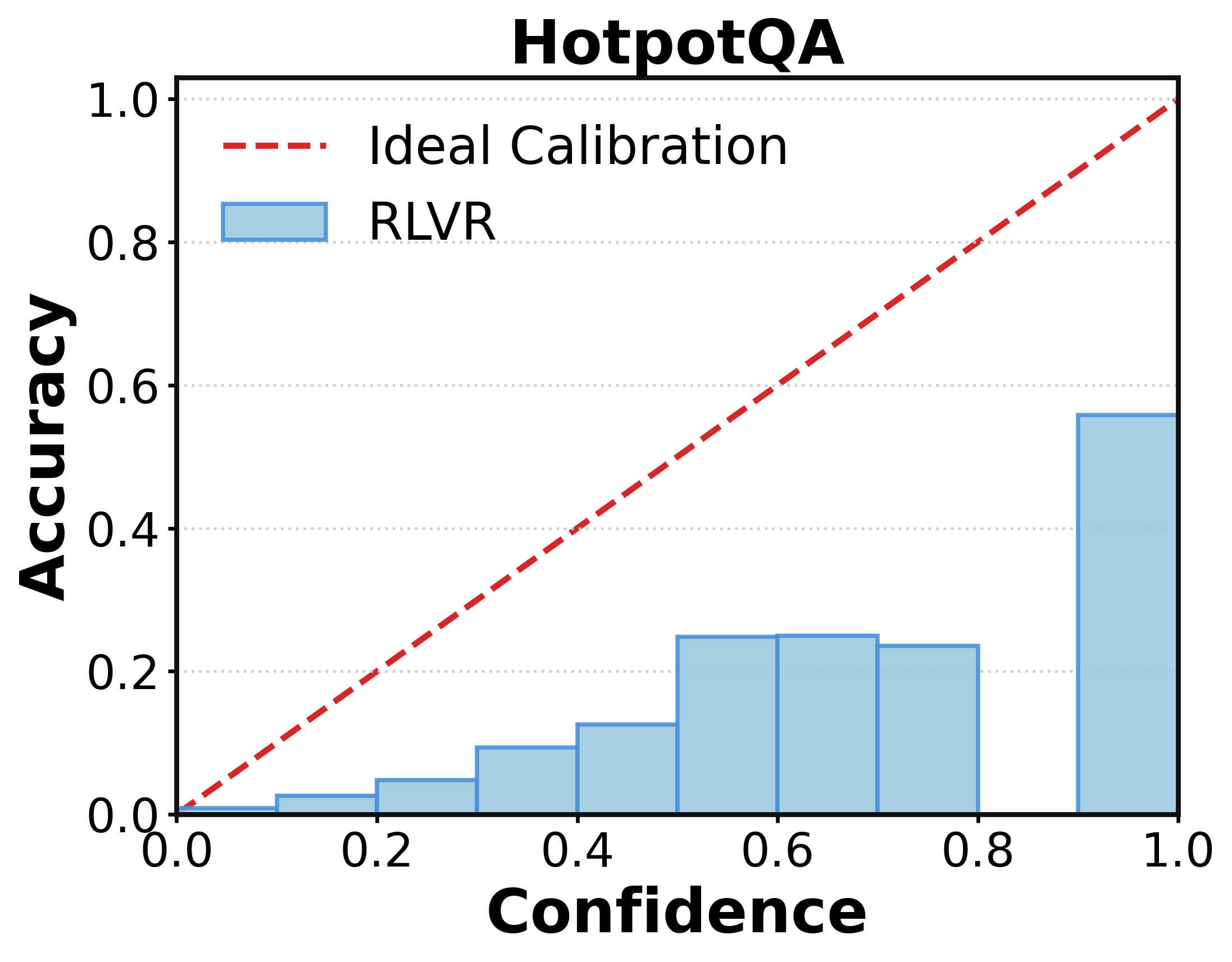}\hfill
\includegraphics[width=0.18\linewidth]{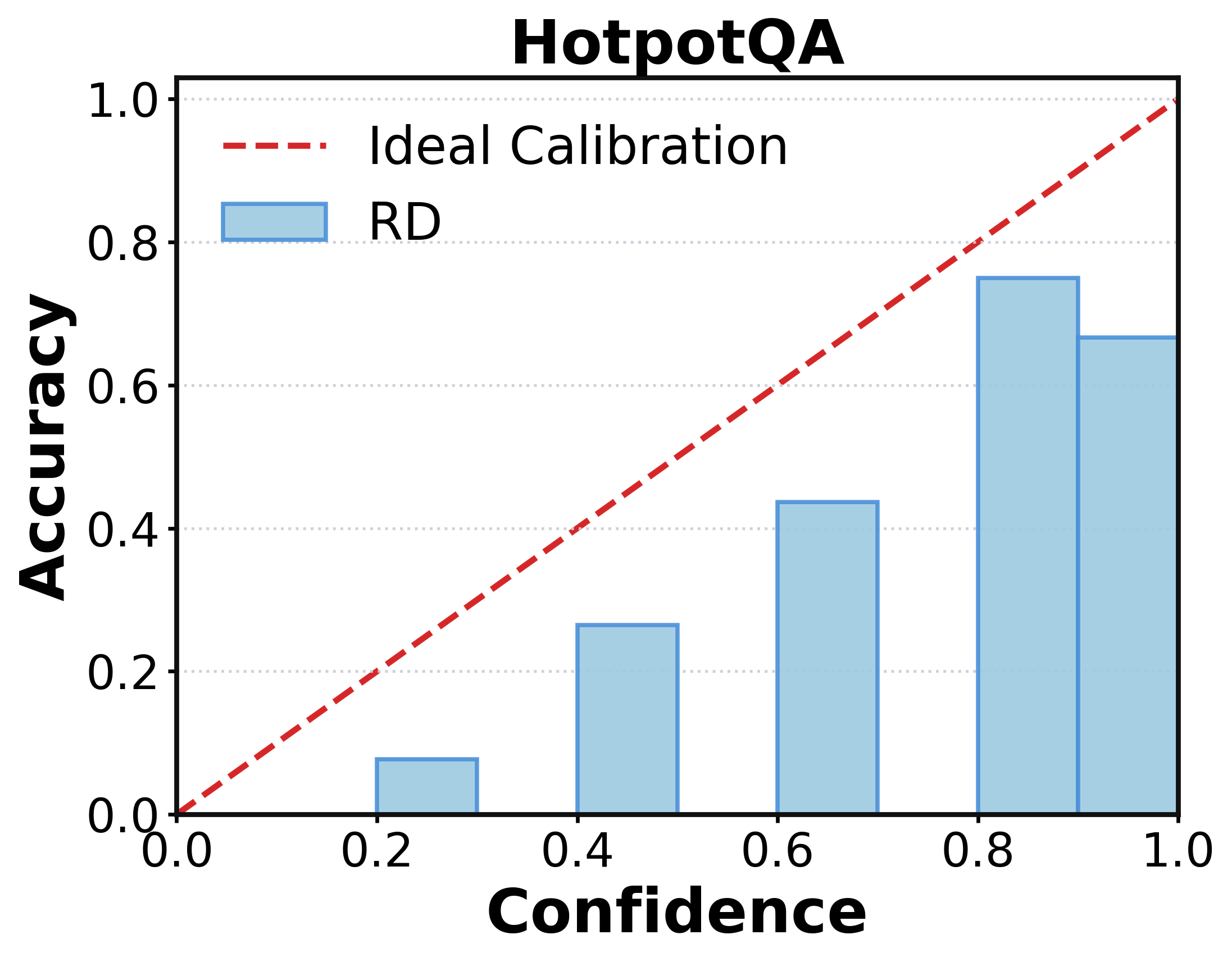}\hfill
\includegraphics[width=0.18\linewidth]{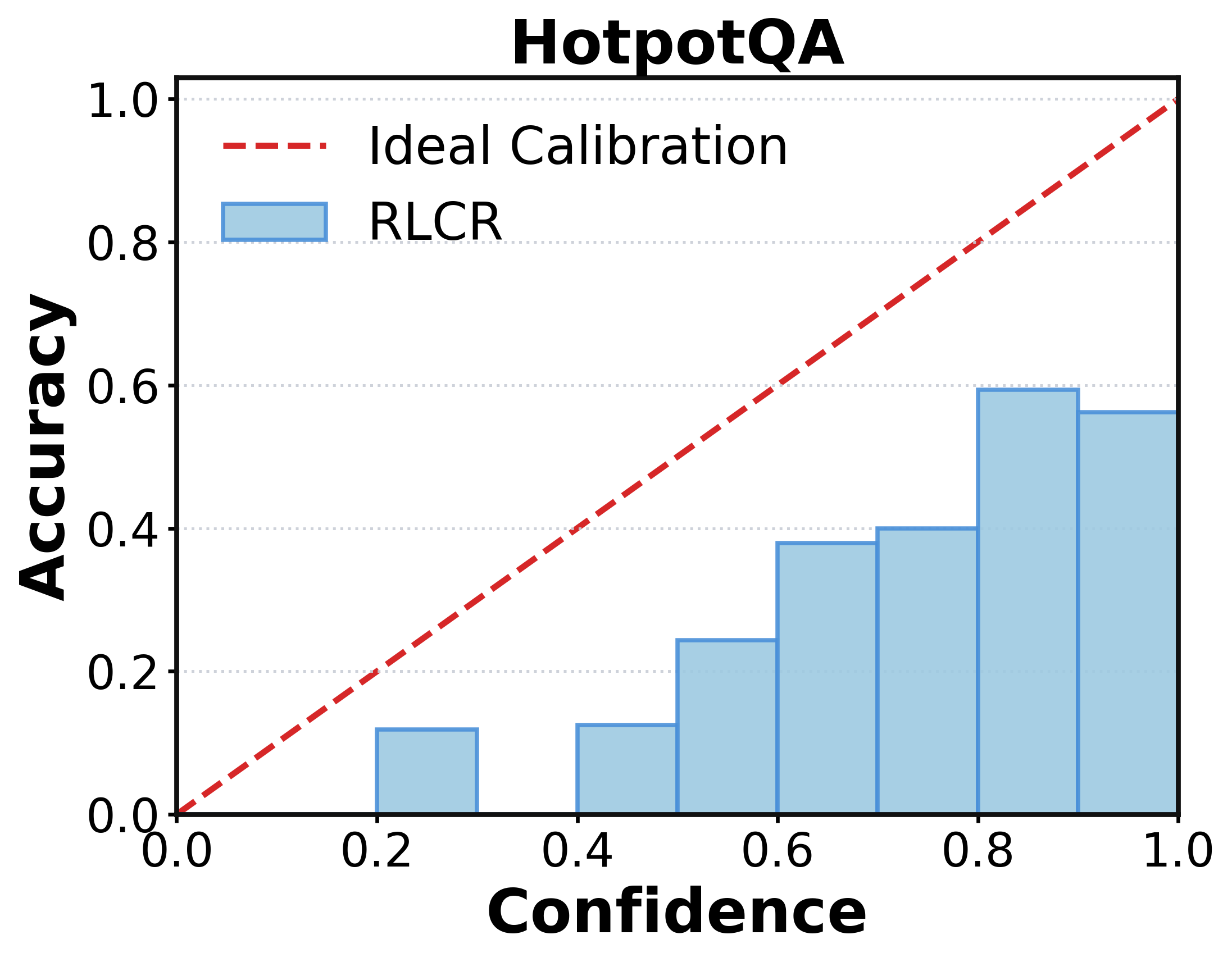}\hfill
\includegraphics[width=0.18\linewidth]{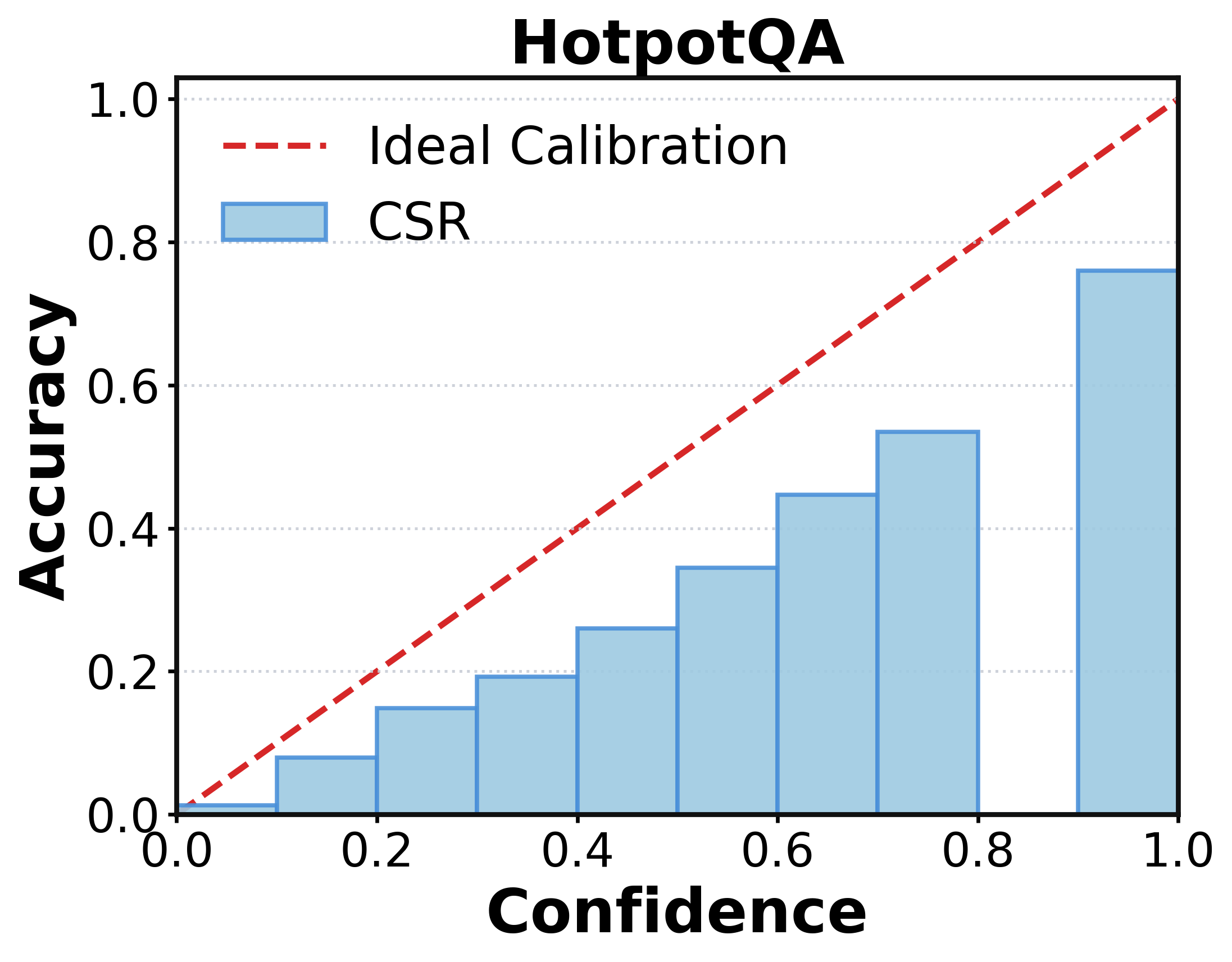}
\end{subfigure}
\vspace{-8pt}

\begin{subfigure}{\linewidth}
\centering
\includegraphics[width=0.18\linewidth]{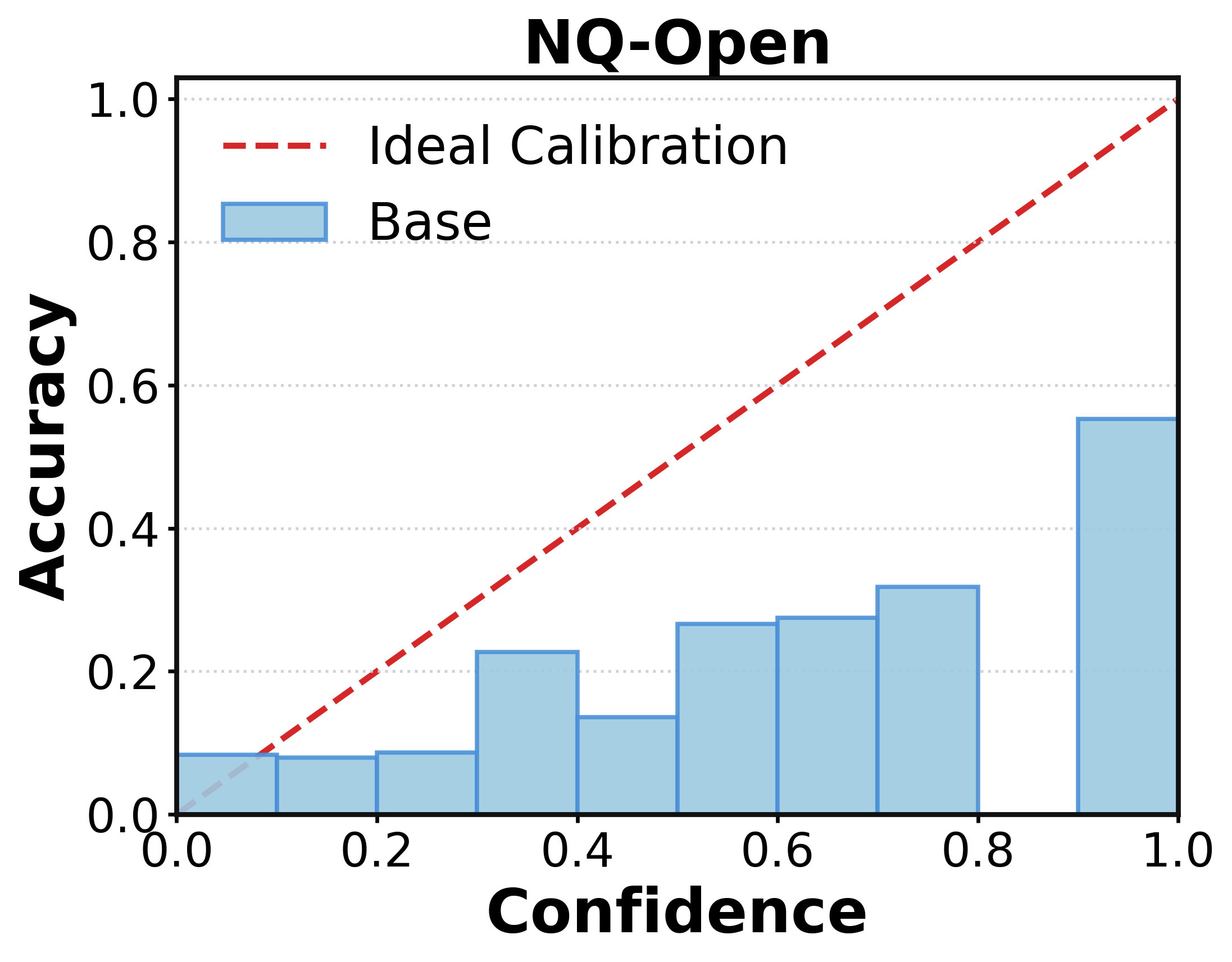}\hfill
\includegraphics[width=0.18\linewidth]{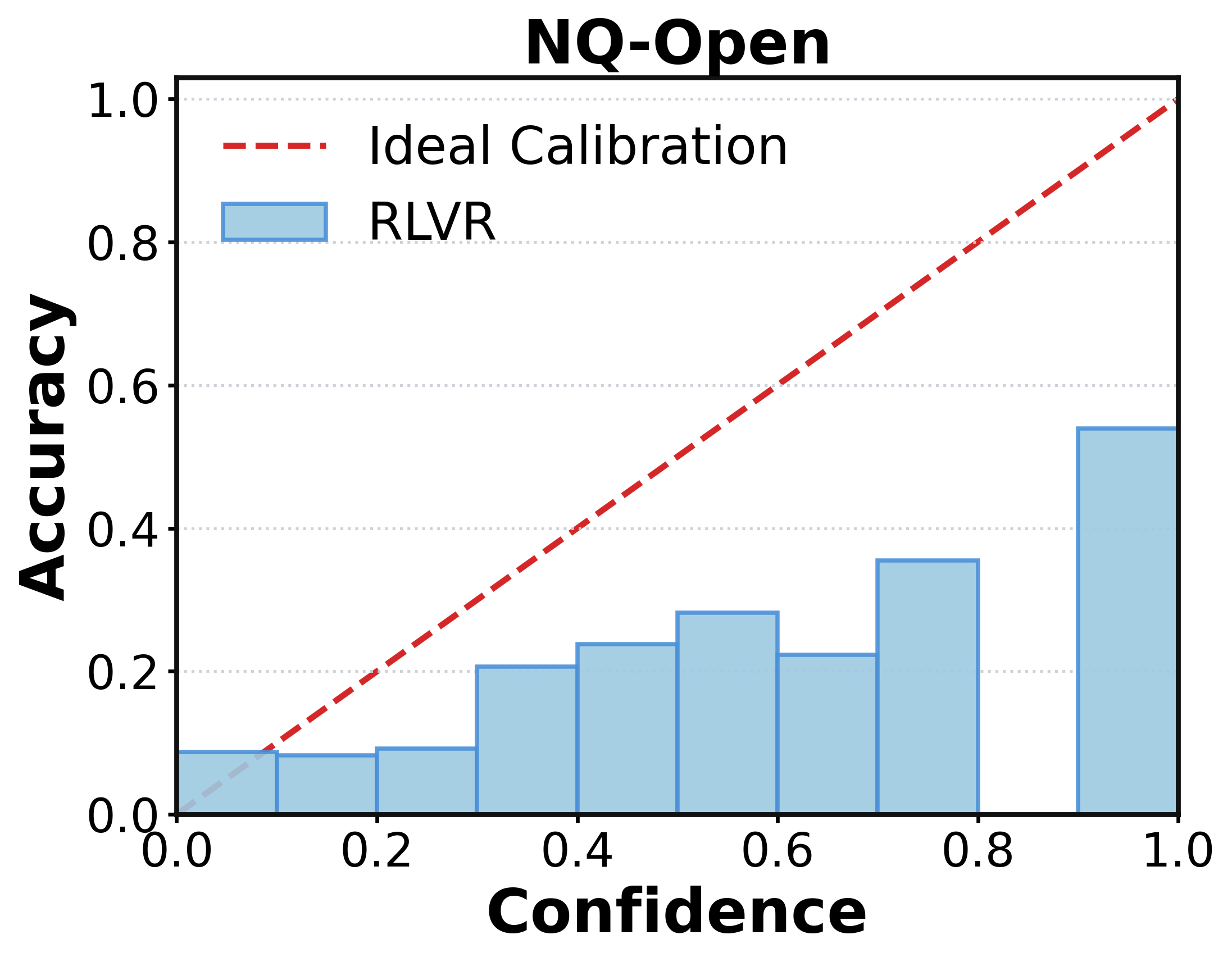}\hfill
\includegraphics[width=0.18\linewidth]{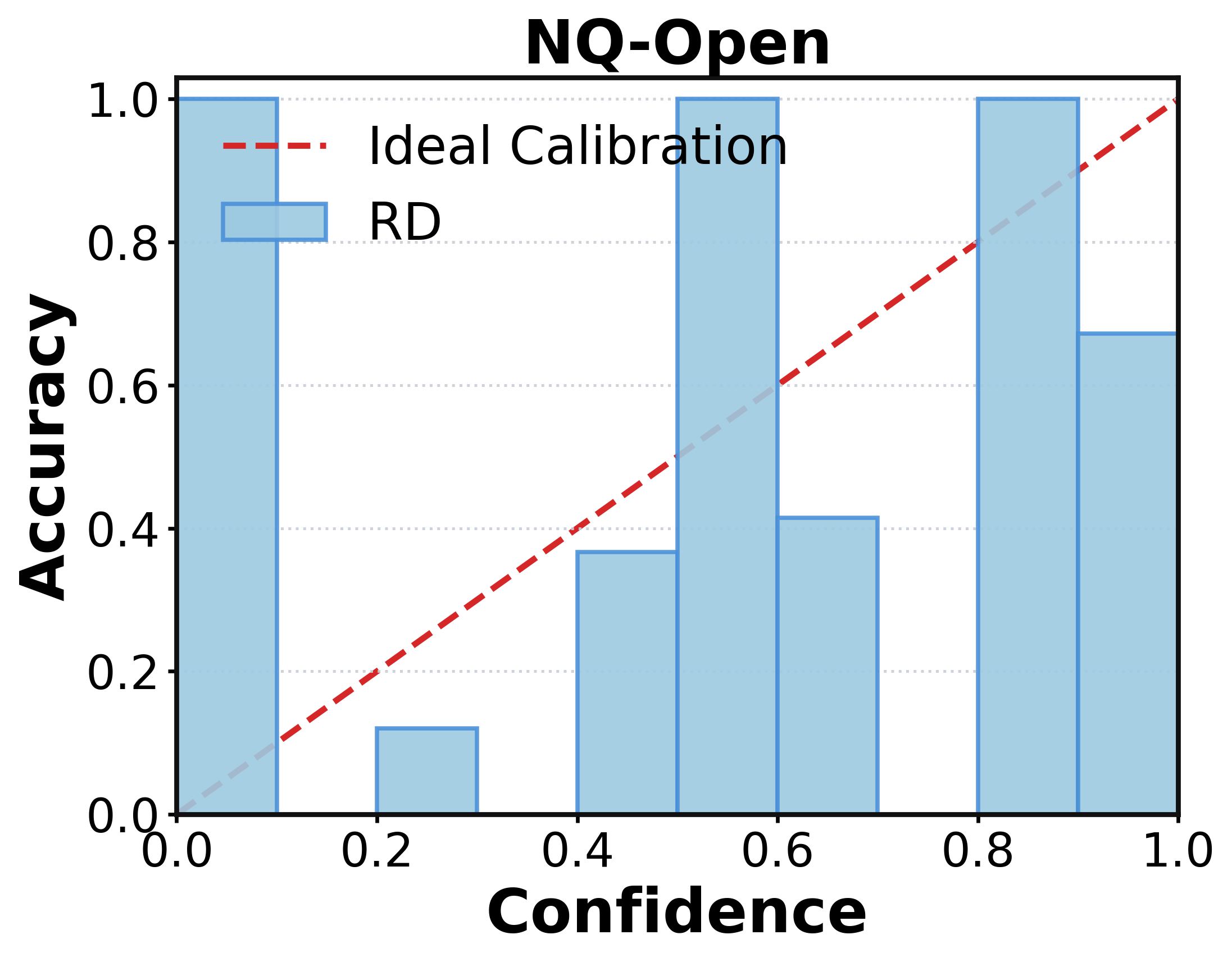}\hfill
\includegraphics[width=0.18\linewidth]{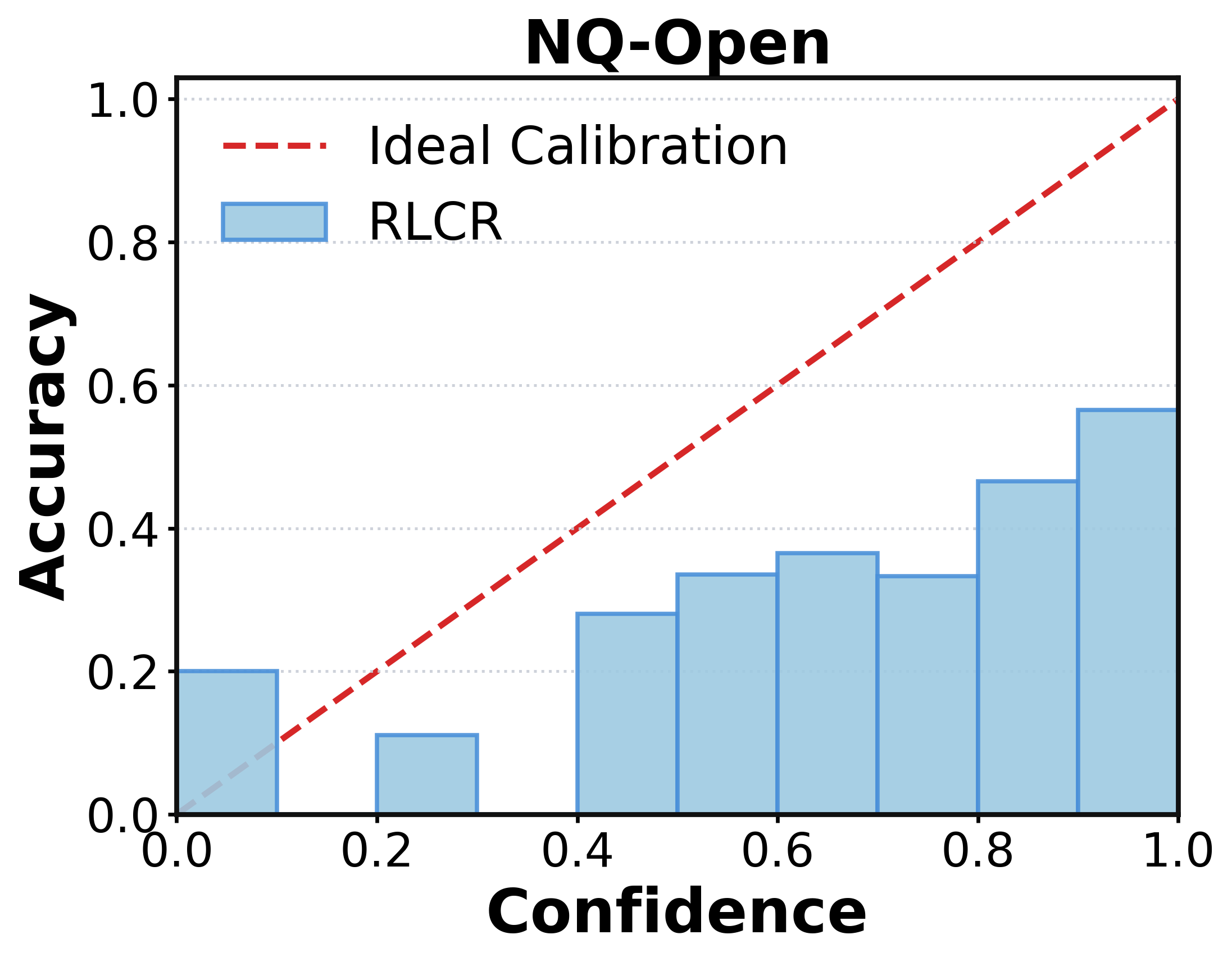}\hfill
\includegraphics[width=0.18\linewidth]{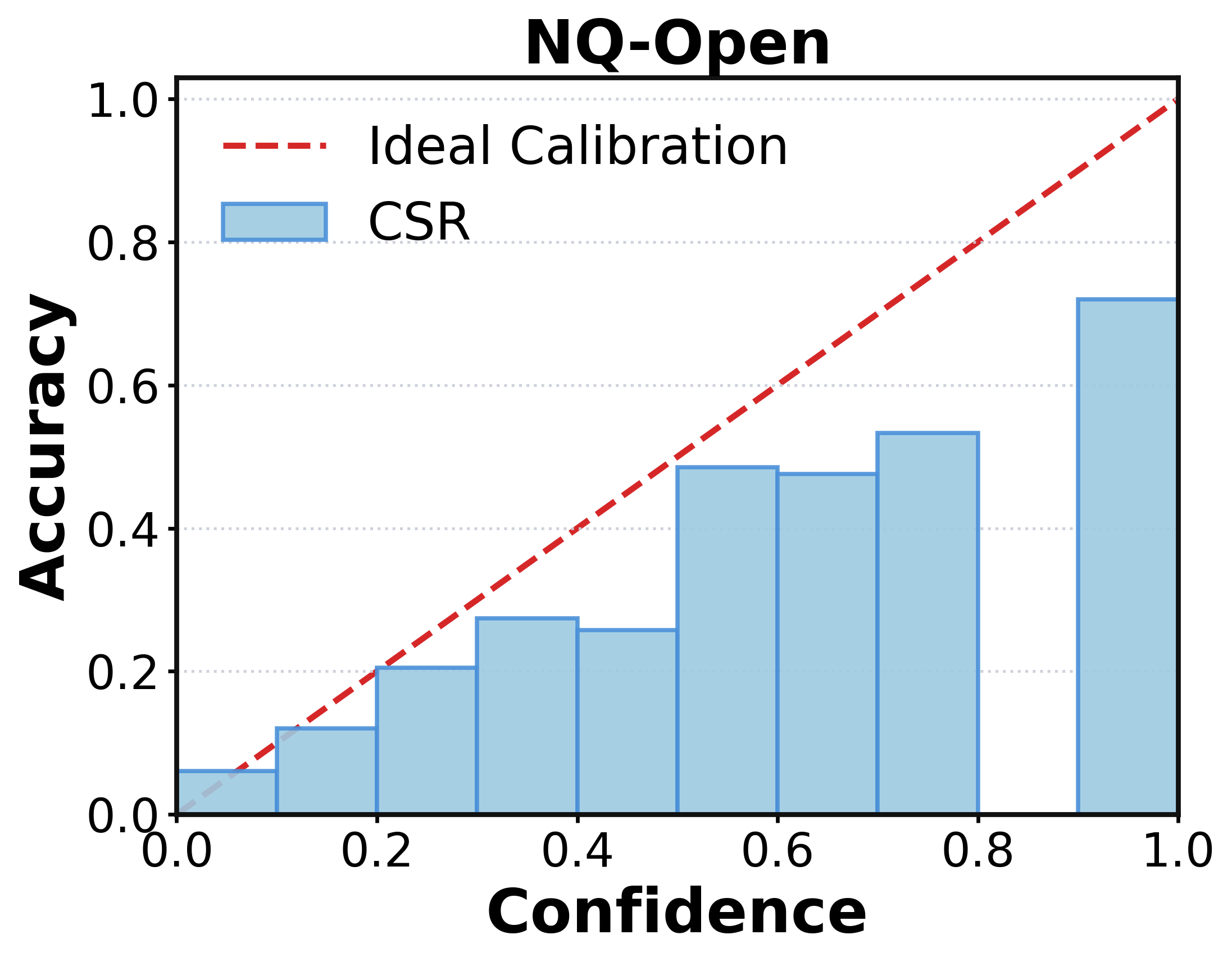}
\end{subfigure}
\caption{Reliability diagrams on Qwen. Rows are HotpotQA (in-distribution) and NQ-Open (out-of-distribution), and columns are Base, RLVR, RD, RLCR, and CSR\@. Each bar is the accuracy of the questions whose confidence falls in a bin of width $0.1$.}
\label{fig:reliability_main}
\vspace{-6pt}
\end{figure*}

Table~\ref{tab:main_results} reports accuracy, ECE, AUROC, and token cost on the in-distribution HotpotQA set and three out-of-distribution (OOD) sets for three model families. The ROC curves for Qwen (Fig.~\ref{fig:sementropy_reliability}) show that CSR's advantage is concentrated at low false-positive rates, the region that matters when only high-confidence answers are accepted: on TriviaQA, for example, CSR reaches a true-positive rate of about $0.7$ at a false-positive rate of $0.1$, while every other method stays below $0.4$. Its reliability diagrams on HotpotQA and NQ-Open also lie closest to the diagonal (Fig.~\ref{fig:reliability_main}), whereas those of Base and RLVR fall well below it.

\textbf{Calibration in and out of distribution.}
On HotpotQA, CSR has the lowest ECE of all five methods on every backbone, the highest AUROC on Llama and Qwen, and cuts ECE by 62--67\% relative to Base. On Llama, ECE falls from $0.2949$ to $0.0978$. The gains carry over to the three OOD datasets, where CSR has the highest AUROC in eight of the nine model-dataset cells and lower ECE than Base and RLVR in all nine. A verbalized baseline has lower ECE in three OOD cells, but CSR keeps the lowest average ECE on every backbone. RLVR reaches higher HotpotQA accuracy but improves ECE far less. On Llama, where its accuracy lead is largest at $0.038$, its ECE is $0.2567$, against $0.0978$ for CSR. CSR's calibration gain is thus not a by-product of accuracy, and its four-dataset accuracy is within $0.02$ of RLVR's on every backbone (Appendix~\ref{app:tradeoff}). CSR therefore learns semantic uncertainty structure that transfers beyond the training domain, which we attribute to supervising meaning rather than a reporting format. RD's low accuracy on Mistral comes in part from abstention. Its Mistral policy declines to answer $34.5\%$ of HotpotQA and $27.6\%$ of NQ-Open questions, and each abstention is scored as incorrect (Appendix~\ref{app:parsing_rate}).

\textbf{Joint calibration and discrimination.}
Averaged over datasets, CSR is the only method that both lowers ECE and raises AUROC relative to Base on every backbone. Low ECE alone is not enough, since a near-constant confidence equal to the average accuracy has low ECE but cannot separate correct from incorrect answers, and RLCR drifts toward this regime. RLCR lowers average ECE relative to Base but also lowers AUROC on every backbone, as RD does on Qwen. In two of the three OOD cells where a verbalized baseline has lower ECE than CSR, its AUROC trails CSR's by $0.15$ to $0.16$. CSR improves both because its confidence is computed from answer agreement, which training ties to correctness, rather than verbalized as a separate output.

\textbf{Inference cost.}
CSR's $K$ rollouts are answer-only, so its Tok stays comparable to Base and RLVR and is lower than RLCR's on all three backbones, by up to $3.2\times$ on Qwen. Beyond generation, CSR, like Base and RLVR, needs pairwise equivalence judgments to form semantic clusters, a cost we address in Sec.~\ref{sec:f1_vs_llm_judge}. Further token-cost analysis is provided in Appendix~\ref{app:token_efficiency}.

\subsection{Does CSR remain effective across sampling-based estimators?}
\label{sec:beyond_se}

\begin{table}[t]
\centering
\caption{Four-dataset average ECE and AUROC under four sampling-based estimators not used in training, applied to the rollouts of Table~\ref{tab:main_results}. \textbf{Bold} marks the best value in each column per estimator.}
\label{tab:beyond_se}
\small
\begin{tabular}{llcccccc}
\toprule
& & \multicolumn{2}{c}{\textbf{Llama}} & \multicolumn{2}{c}{\textbf{Qwen}} & \multicolumn{2}{c}{\textbf{Mistral}}\\
\cmidrule(lr){3-4}\cmidrule(lr){5-6}\cmidrule(lr){7-8}
\textbf{Estimator} & \textbf{Method} & ECE$\downarrow$ & AUROC$\uparrow$ & ECE$\downarrow$ & AUROC$\uparrow$ & ECE$\downarrow$ & AUROC$\uparrow$\\
\midrule
\multirow{3}{*}{SemDens} & Base & 0.1998 & 0.7438 & 0.2529 & 0.7872 & 0.2331 & 0.7415\\
 & RLVR & 0.2090 & 0.7674 & 0.2523 & 0.7820 & 0.2336 & 0.7474\\
\rowcolor{gray!12} \cellcolor{white}& CSR & \textbf{0.1251} & \textbf{0.8378} & \textbf{0.1195} & \textbf{0.8691} & \textbf{0.1517} & \textbf{0.8049}\\
\midrule
\multirow{3}{*}{Bayes-beta} & Base & 0.1950 & 0.7182 & 0.3047 & 0.7559 & 0.2214 & 0.6611\\
 & RLVR & 0.2249 & 0.7322 & 0.3145 & 0.7528 & 0.2169 & 0.6822\\
\rowcolor{gray!12} \cellcolor{white}& CSR & \textbf{0.1585} & \textbf{0.7895} & \textbf{0.1672} & \textbf{0.8383} & \textbf{0.1986} & \textbf{0.6885}\\
\midrule
\multirow{3}{*}{Arithmetic} & Base & 0.1121 & 0.7169 & 0.2851 & 0.7562 & 0.1380 & 0.6655\\
 & RLVR & 0.1516 & 0.7378 & 0.2798 & 0.7583 & 0.1358 & 0.6885\\
\rowcolor{gray!12} \cellcolor{white}& CSR & \textbf{0.0756} & \textbf{0.7985} & \textbf{0.1063} & \textbf{0.8428} & \textbf{0.1072} & \textbf{0.6918}\\
\midrule
\multirow{3}{*}{Clf-pairs+gen} & Base & \textbf{0.0287} & 0.7497 & 0.1153 & 0.6987 & 0.0388 & 0.6555\\
 & RLVR & 0.0618 & 0.7093 & 0.1294 & 0.6756 & 0.0469 & 0.6641\\
\rowcolor{gray!12} \cellcolor{white}& CSR & 0.0312 & \textbf{0.7919} & \textbf{0.0605} & \textbf{0.8160} & \textbf{0.0277} & \textbf{0.7127}\\
\bottomrule
\end{tabular}
\end{table}

CSR is evaluated with semantic entropy in Sec.~\ref{sec:main_results}, so we test whether its gains depend on that estimator. We apply four other post-hoc estimators to the same Base, RLVR, and CSR rollouts, which cover the main design choices of sampling-based estimation. Semantic Density (SemDens)~\citep{qiu2024semantic} measures agreement through semantic equivalence and weights answers by length-normalized sequence likelihood. Three SIMBA UQ aggregations~\citep{bhattacharjya2025simba} instead measure agreement by word-level Jaccard similarity, without the pairwise judge, and aggregate it with a fixed rule (Arithmetic) or a fitted model (Bayes-beta, and Clf-pairs+gen, a supervised random-forest classifier).

\textbf{Robustness to estimator design.}
CSR has lower average ECE and higher average AUROC than both Base and RLVR in 23 of the 24 estimator-backbone-metric comparisons (Table~\ref{tab:beyond_se}), and the highest average AUROC under every estimator. The only exception is ECE on Llama under Clf-pairs+gen, where CSR's $0.0312$ is above Base's $0.0287$ and all three methods are well calibrated. The gain therefore holds whether agreement is judged semantically or lexically and whether the aggregation is fixed or fitted to each policy. Because the estimators are not used in training and only the sampled answers change, the gain comes from the answer distribution CSR learns rather than from one way of reading it. Definitions and per-dataset results are provided in Appendix~\ref{app:alt_estimators}.

\subsection{Does CSR improve selective prediction?}
\label{sec:selective_main}

\begin{wraptable}{r}{0.47\textwidth}
\vspace{-10pt}
\centering
\footnotesize
\setlength{\tabcolsep}{1.5pt}
\caption{Selective prediction averaged over the four datasets. \textbf{Bold} marks the best value in each column.}
\label{tab:selective_main}
\begin{tabular}{l|ccc|ccc}
\toprule
\multicolumn{1}{l}{} & \multicolumn{3}{c}{\textbf{AURAC$\uparrow$}} & \multicolumn{3}{c}{\textbf{AUGRC$\downarrow$}}\\
\cmidrule(lr){2-4}\cmidrule(lr){5-7}
\textbf{Method} & Llama & Qwen & Mistral & Llama & Qwen & Mistral\\
\midrule
Base & 0.5689 & 0.4986 & 0.5758 & 0.2357 & 0.2755 & 0.2320\\
RLVR & 0.6220 & 0.5070 & 0.5992 & 0.2121 & 0.2708 & 0.2210\\
RD & 0.5670 & 0.5053 & 0.4890 & 0.2478 & 0.2802 & 0.2981\\
RLCR & 0.5363 & 0.4848 & 0.5212 & 0.2533 & 0.2798 & 0.2561\\
\rowcolor{gray!12} CSR & \textbf{0.6359} & \textbf{0.5546} & \textbf{0.6393} & \textbf{0.2104} & \textbf{0.2602} & \textbf{0.2087}\\
\bottomrule
\end{tabular}
\end{wraptable}
A common use of confidence is to answer only when confident and abstain otherwise, so that the model refuses the questions it is likely to get wrong. Ranking questions by each method's confidence, we report AURAC~\citep{farquhar2024detecting}, the fraction of correct questions among those retained, integrated over coverage, and AUGRC~\citep{traub2024overcoming}, the fraction of all questions that are both accepted and answered incorrectly, integrated over coverage. CSR has the best average AURAC and AUGRC on all three backbones (Table~\ref{tab:selective_main}) and the best value in 22 of the 24 per-dataset cells. Its average AURAC leads the strongest baseline by $0.014$ (Llama) to $0.048$ (Qwen). RLCR is worse than Base in both metrics on every backbone, consistent with its lower AUROC (Table~\ref{tab:main_results}).

\textbf{Gains in and out of distribution.}
Selective prediction tests whether the confidence can decide when the model should answer, which makes it a form of refusal. Here the model declines by thresholding its derived confidence and its outputs are unchanged, whereas in knowledge-aware refusal~\citep{pan2026refuse} the model itself declines in its output. AURAC and AUGRC depend only on how the confidence orders the questions, so they are unchanged by any monotone rescaling of the score and measure discrimination rather than calibration. CSR is best on HotpotQA for all three backbones under both metrics and on 16 of the 18 OOD cells. Metric definitions, per-dataset results, and a further discussion of refusal are provided in Appendices~\ref{app:metrics} and~\ref{app:selective}.

\subsection{Can token-level F1 replace the pairwise LLM judge?}
\label{sec:f1_vs_llm_judge}

Pairwise equivalence calls dominate CSR's judging cost. At evaluation with $K=16$, they account for $120$ of the $136$ judgments per question. We ask whether a lightweight lexical proxy can replace the pairwise equivalence judge during both training and post-hoc evaluation. We compare (i) an \emph{LLM judge} of pairwise semantic equivalence and (ii) \emph{token-level F1 thresholding}, which declares two answers equivalent when their F1 reaches a fixed threshold $\tau$. The threshold is chosen once per model family on training rollouts to best agree with the judge ($0.50$ for Llama and Mistral, $0.80$ for Qwen), and correctness is scored by the LLM judge for every method. Table~\ref{tab:efficient_sc_llama} reports Llama results, and Table~\ref{tab:efficient_sc} reports all three backbones.

\textbf{F1 preserves CSR's calibration gains.}
Replacing the pairwise judge with F1 barely changes CSR's calibration. Its average ECE stays within $0.01$ of CSR-LLM on every backbone, and on Qwen it even drops from $0.1108$ to $0.1021$. CSR-F1 also has higher AUROC than Base-F1 and RLVR-F1 on every dataset and backbone. On Llama its average AUROC of $0.8116$ exceeds the $0.7474$ of the LLM-judged Base. CSR's calibration gain therefore does not require a pairwise LLM judge.

Relative to CSR-LLM, CSR-F1's average AUROC is lower by $0.023$ on Llama to $0.062$ on Mistral, and the loss is concentrated where answers vary in wording. On HotpotQA, whose answers are short entities, CSR-F1 is within $0.019$ of CSR-LLM on every backbone, and on Llama it is even higher on NQ-Open. On Qwen and Mistral the drop is largest on MSMARCO, where CSR-F1 loses $0.149$ and $0.131$ AUROC. Its answers are free-form sentences whose paraphrases share few tokens, so F1 likely splits equivalent answers into separate clusters. Results for Qwen and Mistral and the judge-cost accounting are provided in Appendix~\ref{app:efficient_sc}.
\begin{table}[t]
\centering
\footnotesize
\setlength{\tabcolsep}{3.0pt}
\renewcommand{\arraystretch}{1.1}
\caption{Token-level F1 versus the LLM judge for pairwise equivalence on Llama. The suffix names the equivalence test used for clustering, and CSR-F1 also uses F1 in training. The -LLM rows, RD, and RLCR repeat Table~\ref{tab:main_results}. \textbf{Bold} marks the best and \underline{underline} the second-best value in each column.}
\label{tab:efficient_sc_llama}
\ADLactivate
\begin{tabular}{l c c|c c|c c|c c|c c}
\toprule
& \multicolumn{2}{c|}{\textbf{HotpotQA}} & \multicolumn{2}{c|}{\textbf{TriviaQA}} & \multicolumn{2}{c|}{\textbf{MSMARCO}} & \multicolumn{2}{c|}{\textbf{NQ-Open}} & \multicolumn{2}{c}{\textbf{Average}}\\
\cmidrule(lr){2-3}\cmidrule(lr){4-5}\cmidrule(lr){6-7}\cmidrule(lr){8-9}\cmidrule(lr){10-11}
\textbf{Method} & ECE$\downarrow$ & AUROC$\uparrow$ & ECE$\downarrow$ & AUROC$\uparrow$ & ECE$\downarrow$ & AUROC$\uparrow$ & ECE$\downarrow$ & AUROC$\uparrow$ & ECE$\downarrow$ & AUROC$\uparrow$\\
\midrule
RD & 0.1764 & 0.8316 & \underline{0.0746} & 0.7996 & 0.3302 & 0.7298 & 0.2081 & 0.7464 & 0.1973 & 0.7769 \\
RLCR & 0.1757 & 0.6923 & 0.0943 & 0.7337 & 0.2340 & 0.6816 & 0.1497 & 0.7013 & 0.1634 & 0.7022 \\
\noalign{\vskip 0.2ex}\cdashline{1-11}\noalign{\vskip 0.2ex}
Base-LLM & 0.2949 & 0.7050 & 0.0772 & 0.8423 & 0.1996 & 0.7062 & 0.1186 & 0.7361 & 0.1726 & 0.7474 \\
RLVR-LLM & 0.2567 & 0.8089 & 0.0901 & 0.8268 & 0.2280 & \underline{0.7371} & 0.1686 & 0.6957 & 0.1859 & 0.7671 \\
\rowcolor{gray!12} CSR-LLM & \textbf{0.0978} & \textbf{0.9133} & \textbf{0.0670} & \textbf{0.8976} & \underline{0.1571} & \textbf{0.7594} & \textbf{0.1100} & \underline{0.7670} & \textbf{0.1080} & \textbf{0.8343} \\
\noalign{\vskip 0.2ex}\cdashline{1-11}\noalign{\vskip 0.2ex}
Base-F1 & 0.2713 & 0.6681 & 0.1212 & 0.7929 & 0.1612 & 0.6094 & 0.1223 & 0.6992 & 0.1690 & 0.6924 \\
RLVR-F1 & 0.2727 & 0.7770 & 0.0899 & 0.7819 & 0.2296 & 0.6670 & 0.1798 & 0.6771 & 0.1930 & 0.7258 \\
\rowcolor{gray!12} CSR-F1 & \underline{0.1240} & \underline{0.9008} & 0.1005 & \underline{0.8498} & \textbf{0.1042} & 0.7223 & \underline{0.1109} & \textbf{0.7735} & \underline{0.1099} & \underline{0.8116} \\
\bottomrule
\end{tabular}
\end{table}

\subsection{Ablation studies}
\label{sec:ablations}

\begin{figure*}[t]
\centering
\begin{subfigure}[t]{0.32\textwidth}
\centering
\includegraphics[width=0.98\linewidth]{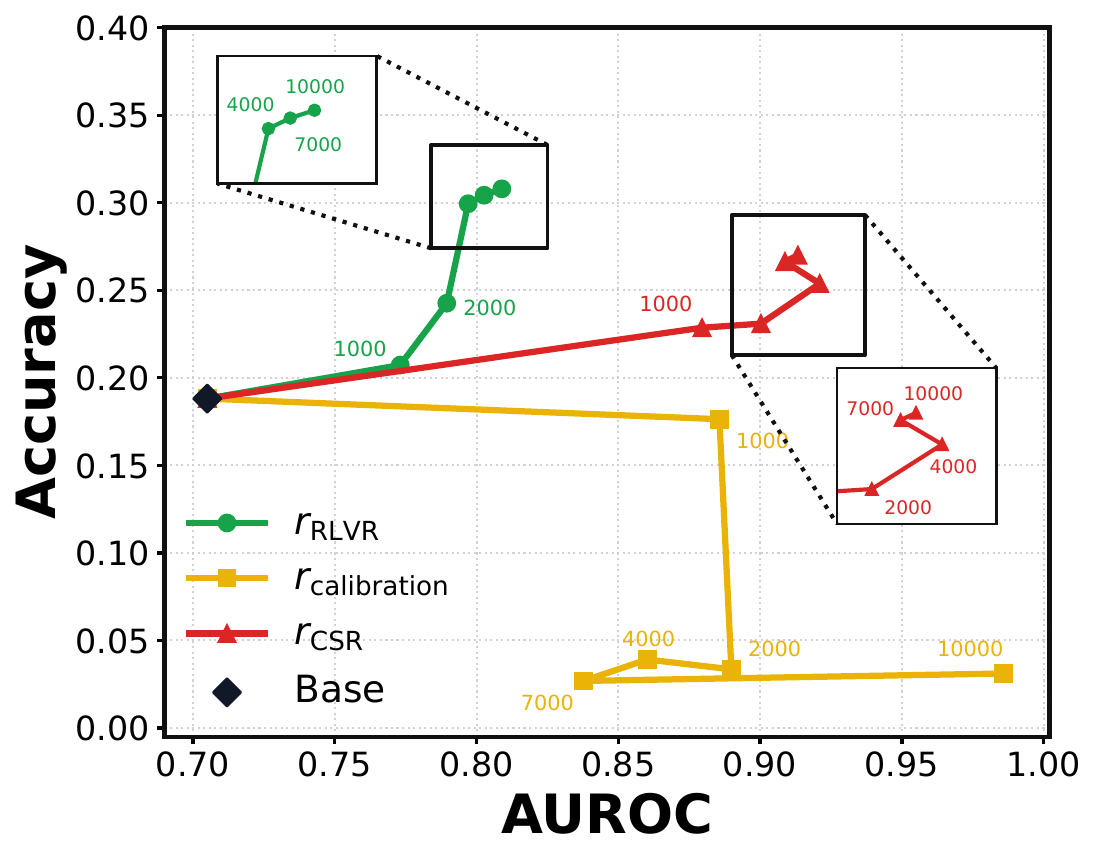}
\caption{Reward component ablation}
\label{fig:ablation_reward_components}
\end{subfigure}
\hfill
\begin{subfigure}[t]{0.32\textwidth}
\centering
\includegraphics[width=0.98\linewidth]{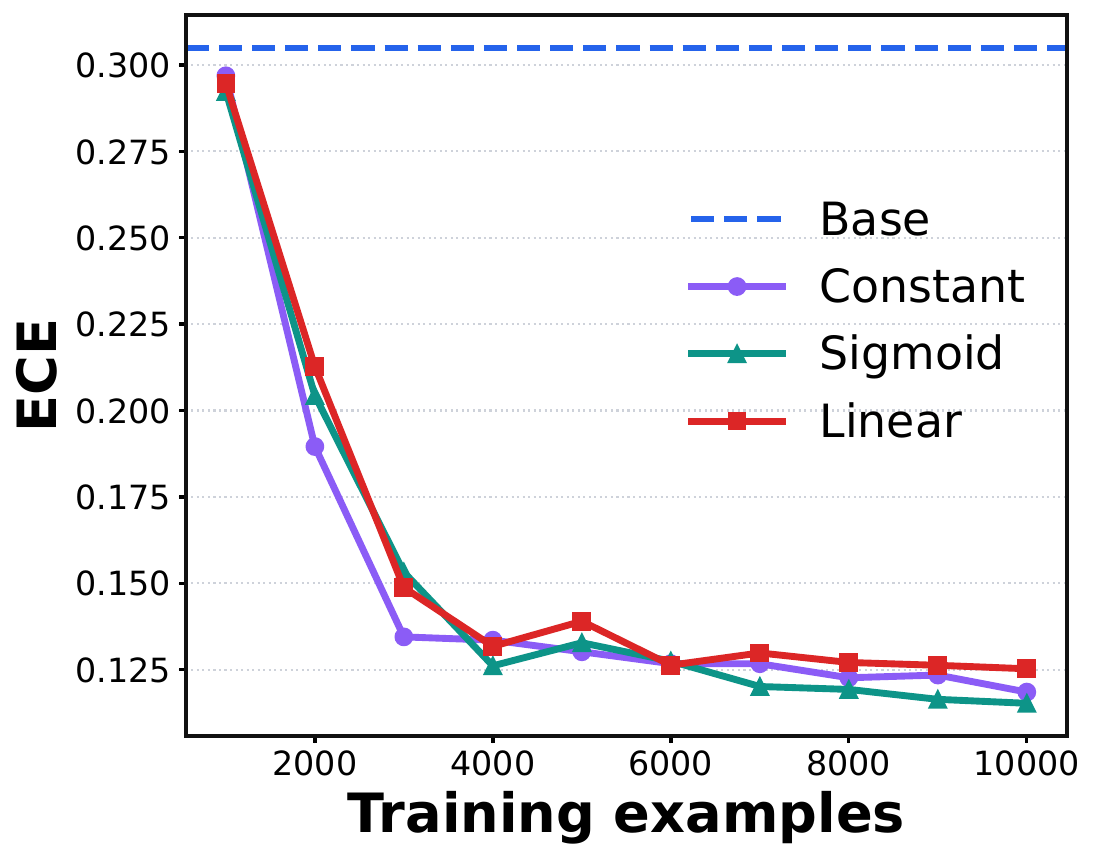}
\caption{Schedule comparison}
\label{fig:ablation_schedule_comparison}
\end{subfigure}
\hfill
\begin{subfigure}[t]{0.32\textwidth}
\centering
\includegraphics[width=0.98\linewidth]{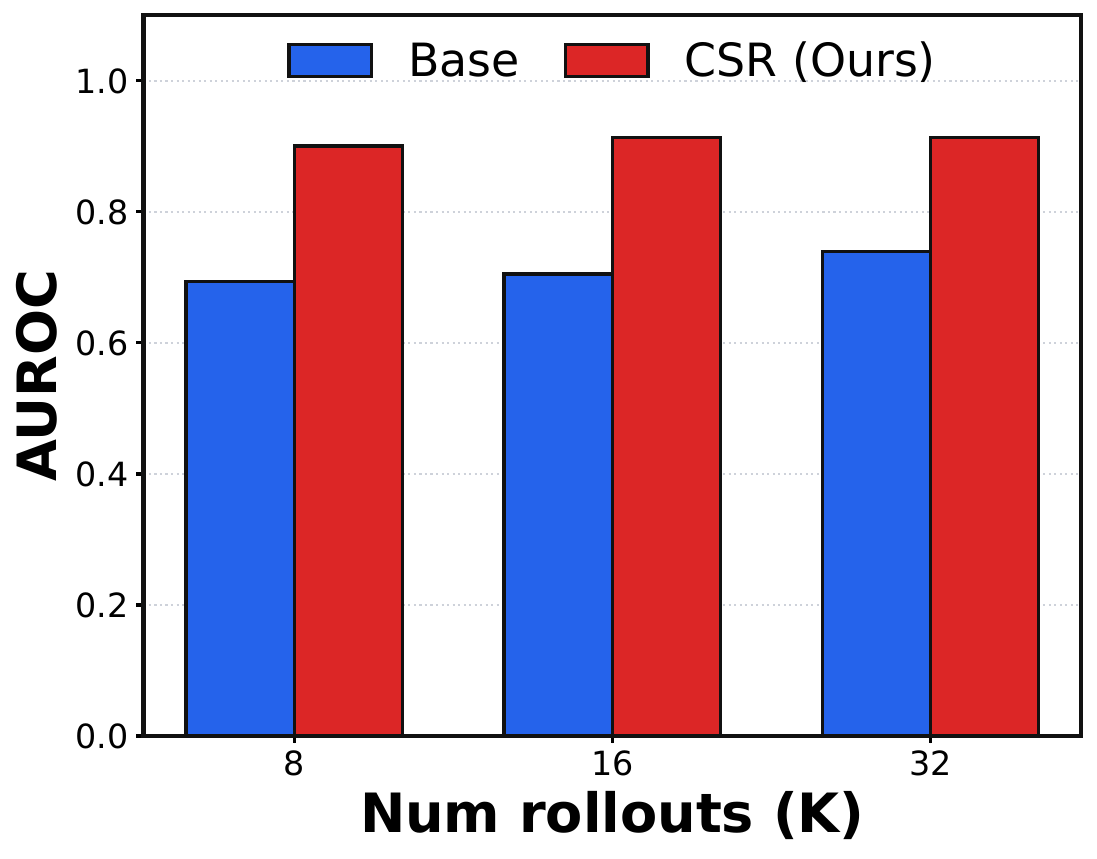}
\caption{Evaluation rollout budget}
\label{fig:ablation_sample_count}
\end{subfigure}
\caption{Ablations on HotpotQA. (a)~Training trajectories of RLVR, the calibration reward alone, and CSR in the AUROC-accuracy plane on Llama, starting from Base, with point labels giving the number of training examples. (b)~ECE over training for three $\lambda(t)$ schedules on Qwen, with Base as the dashed line. (c)~AUROC of Base and CSR against the evaluation budget $K$ on Llama.}
\label{fig:ablation_row}
\end{figure*}

All ablations are run on HotpotQA, with Llama-3.1-8B-Instruct in Figs.~\ref{fig:ablation_row}(a) and~\ref{fig:ablation_row}(c) and Qwen2.5-7B-Instruct in Fig.~\ref{fig:ablation_row}(b). Ablation settings are provided in Appendix~\ref{app:exp_setup}, and analyses of larger backbones, context-grounded training, and smaller budgets are provided in Appendix~\ref{app:supp_analyses}.

\textbf{Reward component ablation.}
To test whether both rewards are needed, we train with RLVR alone, the calibration reward alone, and CSR, and track accuracy and AUROC over training (Fig.~\ref{fig:ablation_row}(a)). RLVR raises accuracy from $0.19$ to $0.31$, but its AUROC reaches only $0.81$. RLVR is indifferent to whether incorrect rollouts agree, so a wrong answer produced consistently looks as certain as a right one. The calibration reward alone fails in the opposite way. Its accuracy falls to about $0.03$ within 2k examples, the low-accuracy regime of Proposition~\ref{prop:meanfield}, and its AUROC then swings between $0.84$ and $0.99$ because it is computed on very few correct questions. CSR's AUROC stays above RLVR's at every checkpoint. At 10k examples it reaches $0.91$ against $0.81$ for RLVR, while its accuracy rises to $0.27$, about two-thirds of RLVR's gain. In CSR, the correctness term selects the answer and the calibration term shapes agreement around it.

\textbf{Schedule comparison.}
To test whether the gain depends on how $\lambda(t)$ is ramped, we compare the sigmoid schedule of the main experiments with a constant weight and a linear schedule (Fig.~\ref{fig:ablation_row}(b)). All three cut ECE from about $0.30$ to $0.13$ within the first 4k examples and then improve slowly. The constant weight, which is largest early in training, lowers ECE fastest and reaches $0.19$ at 2k examples, while sigmoid and linear reach $0.20$ and $0.21$. The sigmoid schedule ends lowest at $0.1152$, against $0.118$ for constant and $0.125$ for linear. The final values differ by about $0.01$ while all three stay far below Base and RLVR, so the gain comes from the calibration reward itself, and the schedule mainly trades early speed against the final value.

\textbf{Evaluation rollout budget.}
To test how discrimination depends on the number of rollouts sampled at test time, we vary the evaluation budget $K\in\{8,16,32\}$ (Fig.~\ref{fig:ablation_row}(c)). Base's AUROC rises with $K$, from $0.69$ at $K=8$ to $0.74$ at $K=32$, while CSR's stays at $0.90$--$0.91$. The AUROC gap between CSR and Base narrows from $0.21$ at $K=8$ to $0.17$ at $K=32$, and CSR with $K=8$ still outperforms Base with $K=32$. A likely reason is that training concentrates CSR's correct rollouts in one cluster, which a few samples already identify, whereas Base spreads its mass over more meanings and needs more samples for a stable entropy estimate. CSR's AUROC still rises slightly from $K=8$ to $K=16$, as a larger budget gives a more accurate estimate.

\section{Conclusion}\label{sec:conclusion}
We propose Calibration with Semantic Reward (CSR), a framework that calibrates the confidence derived from a language model's semantic answer distribution without any explicit confidence interface. CSR is a training-time policy intervention rather than a confidence-elicitation method or a post-hoc calibration layer: coupling a verifiable correctness reward with a semantic calibration reward shapes the answer distribution so that semantic agreement across rollouts becomes an informative proxy for correctness under a fixed post-hoc estimator. Across three model families and four open-ended QA datasets, CSR achieves the lowest four-dataset average ECE and the highest average AUROC on every backbone, while trading at most $0.02$ of four-dataset average accuracy against training on correctness alone. The gains also hold under four other sampling-based estimators and in selective prediction.

\textbf{Limitations and future work.} The semantic equivalence judge requires additional LLM evaluations among rollouts, introducing computational overhead. Although token-level F1 thresholding provides an efficient alternative, a performance gap remains compared with the LLM judge. Future work could further address this by designing more fine-grained, rule-based consistency metrics for rollout comparison. Our experiments also cover only short-answer QA, and long-form generation or code would need an equivalence judge for longer outputs. Despite these limitations, CSR represents a promising direction toward reliable uncertainty estimation without explicit confidence interfaces.

\newpage
\begin{ack}
This work is supported in part by the U.S. Army Research Office under Army-ECASE award W911NF-07-R-0003-03, the U.S. Department of Energy, Office of Science, ARPA-H-SOL-24-101 program, IARPA HAYSTAC Program, DARPA YFA, NSF Grants \#2205093, \#2146343, \#2134274, \#2441832. This work is partially supported by the NSF award CCF-2112665 (TILOS), also supported in part by the CDC-RFA-FT-23-0069 from the CDC's Center for Forecasting and Outbreak Analytics. The authors declare no competing interests.
\end{ack}

\bibliographystyle{plainnat}
\bibliography{references}

\newpage
\appendix

\section{Comparison of training-time calibration methods}
\label{app:related_work_full}

Training-time calibration methods differ in what the training signal is computed from and in what the deployed model must emit for a confidence score to exist, as summarized in Table~\ref{tab:rw_comparison}.

\begin{table}[H]
\centering
\small
\setlength{\tabcolsep}{5pt}
\renewcommand{\arraystretch}{1.15}
\caption{Training-time methods for calibrated confidence, by the source of the training signal and the output the deployed model must emit for a confidence score to exist.}
\label{tab:rw_comparison}
\begin{tabular}{@{}>{\raggedright\arraybackslash}p{0.27\textwidth}>{\raggedright\arraybackslash}p{0.43\textwidth}>{\raggedright\arraybackslash}p{0.24\textwidth}@{}}
\toprule
\textbf{Method} & \textbf{Training signal computed from} & \textbf{Emitted at inference}\\
\midrule
SaySelf~\citep{xu2024sayself} & Sampled-answer consistency (SFT targets), then stated confidence (RL) & Rationale $+$ confidence\\
\addlinespace[3pt]
RD~\citep{bani2025rewarding} & Stated confidence (log scoring rule) & Confidence token\\
\addlinespace[3pt]
RLCR~\citep{damani2025beyond} & Stated confidence (Brier term) & Confidence token\\
\addlinespace[3pt]
ConfTuner~\citep{li2025conftuner} & Confidence-token distribution & Confidence token\\
\addlinespace[3pt]
LACIE~\citep{stengel2024lacie} & Listener's inferred confidence & Listener-judged utterance\\
\addlinespace[3pt]
LoVeC~\citep{lovec2025} & Stated confidence per statement & Per-statement confidence\\
\addlinespace[3pt]
\citet{kapoor2024large} & Correctness labels of the model's own answers & Confidence head or token\\
\addlinespace[3pt]
FUT~\citep{eikema2025faithfully} & Sampled answers & Verbal hedge\\
\addlinespace[3pt]
RLMF~\citep{liu2026metacognitive} & Agreement of stated confidence with sampled-answer consistency & Numeric confidence $\rightarrow$ verbal hedge\\
\midrule
\textbf{CSR (Ours)} & \textbf{Semantic agreement of sampled rollouts, rewarded to match correctness} & \textbf{Answer only ($K$ samples)}\\
\bottomrule
\end{tabular}
\end{table}

The table shows two differences from CSR\@. First, every prior method requires the deployed model to produce something beyond the answer, such as a verbalized confidence, a hedged utterance, or the output of a trained confidence head, so its confidence score depends on that additional channel. Second, the methods that share our source signal, SaySelf's supervised stage, FUT, and RLMF, still use that signal as a target to be reported rather than as a reward that reshapes the distribution the signal is computed from. RLMF, the closest of these, rewards a stated confidence for matching the consistency of sampled answers, whereas CSR scores the rollouts themselves: it rewards the agreement structure, which reshapes the answer distribution, so its gains persist under estimators not used in training (Appendix~\ref{app:alt_estimators}).

\section{Experimental details}
\label{app:exp_details}

\subsection{Training and evaluation settings}
\label{app:exp_setup}

\textbf{Hardware and software.} All experiments run on a server with Ubuntu 22.04.4 LTS and NVIDIA A100 80GB PCIe GPUs, using CUDA 12.8, Python 3.11.14, and PyTorch 2.9.0. Each individual fine-tuning or evaluation run uses a single A100 80GB GPU.

\textbf{Fine-tuning setup (CSR and RLVR).} We implement CSR and RLVR with Unsloth~\citep{unsloth}, TRL~\citep{vonwerra2020trl}, and vLLM~\citep{kwon2023efficient} and use one recipe across the three model families of Table~\ref{tab:main_results}. For training, we shuffle HotpotQA~\citep{yang2018hotpotqa} with seed 42 and use the first 10{,}000 examples. Models are loaded in 4-bit precision and fine-tuned with LoRA adapters using rank 32 and alpha 32. We sample $K=8$ rollouts per example with temperature $0.7$ and top-$p$ $0.95$, as at evaluation, use one prompt group per microbatch with gradient accumulation of 4 (effective batch size of 4 prompts and 32 rollouts per optimizer step), and train for one epoch of $2{,}500$ optimizer steps. We set the KL coefficient $\beta=0.01$, the agreement clipping constant of Eq.~\eqref{eq:rlsr} to $\epsilon=10^{-6}$, the maximum prompt length to 256 tokens, and the maximum completion length to 768 tokens, and use a learning rate of $5\times 10^{-6}$, an 8-bit AdamW optimizer ($\beta_1=0.9$, $\beta_2=0.99$), weight decay 0.1, a warmup ratio of 0.1, and a cosine learning-rate schedule.

\textbf{Fine-tuning setup (RD and RLCR).} RD and RLCR follow the original settings and released code of \citet{bani2025rewarding} and \citet{damani2025beyond}, and we list only the settings that differ. Both are trained on the same $10{,}000$ HotpotQA examples and backbones as CSR, with the same rank-32 LoRA adapters. RD is trained for one epoch of $1{,}250$ PPO steps with 8 prompts per step. RLCR uses the same training rollout count ($K=8$) and batch size (4 prompts and 32 rollouts per optimizer step) as CSR\@. It keeps its original context-grounded HotpotQA prompt with stochastic paragraph removal. The original release evaluates HotpotQA with these paragraphs and the other datasets without them, whereas we evaluate every dataset closed-book, as for all methods. We do not retrain RLCR closed-book because the prompt and context format are part of the method.

\textbf{CSR reward settings.} For CSR, the calibration reward enters in its normalized form (Sec.~\ref{sec:csr_obj}), which does not depend on $\epsilon$. We judge each unordered rollout pair once with the semantic-equivalence judge prompt (Appendix~\ref{app:system_prompt}) and share its verdict between both endpoints, giving $\binom{8}{2}=28$ pairwise judgments per group. Every main-results run uses the sigmoid calibration-weight schedule (iii) defined in the ablation setup below, with $\lambda_{\min}=0.8$, $\lambda_{\max}=1.2$, and $\kappa=10$ on every backbone except Qwen3-14B, which uses $\lambda_{\min}=0.5$ and $\lambda_{\max}=0.7$ (Appendix~\ref{app:scale_14b}), and the other schedules are used only in the schedule comparison of Fig.~\ref{fig:ablation_row}(b).

\textbf{Evaluation setup.} For evaluation on the four datasets (HotpotQA, TriviaQA, MSMARCO, NQ-Open), we use the validation split of each dataset, which does not intersect with the HotpotQA training data used for fine-tuning. We shuffle each split with seed 42 and use the first 1{,}000 examples. Base, RLVR, and CSR sample $K=16$ rollouts per question with temperature $0.7$ and top-$p$ $0.95$, while RD and RLCR use greedy decoding. Generation is capped at 256 new tokens for Base, RLVR, and CSR, and for RD, whose answer and confidence come in one short output, and at 1536 for RLCR, whose structured output includes a reasoning segment.

\textbf{Motivation experiment setup.}
The motivation experiments use the RLCR fine-tuned Llama-3.1-8B-Instruct model described above. For Fig.~\ref{fig:fig1_mismatch_separation}(a), we greedy-decode the response to a single HotpotQA question and plot the probabilities the model assigns to each confidence value at the token position where the confidence is emitted. For Figs.~\ref{fig:fig1_mismatch_separation}(b) and~\ref{fig:fig1_mismatch_separation}(c), we sample $8$ answers per question with temperature $0.7$ and top-$p$ $0.95$. The example in Fig.~\ref{fig:fig1_mismatch_separation}(b) shows three out of eight rollouts that all produce the same correct answer, ``Take It Easy'', but report different confidence values. For Fig.~\ref{fig:fig1_mismatch_separation}(c), we measure confidence variance within semantically equivalent groups over $1{,}000$ HotpotQA examples. For each question, we use the LLM judge to cluster the $8$ sampled answers into different semantic groups. For each group with at least two semantically equivalent samples, we compute the standard deviation of the reported verbalized confidence values within that group. We then aggregate these group-level standard deviations by group size and plot their distributions, measuring whether verbalized confidence remains stable among rollouts that express the same answer.

\textbf{Ablation setup.} We keep the fine-tuning details of the main results and run all ablations on HotpotQA\@. Fig.~\ref{fig:ablation_row}(a) uses Llama-3.1-8B-Instruct and shows the effect of each reward component using checkpoints of each method's 10{,}000-example run after $N\in\{1{,}000,\,2{,}000,\,4{,}000,\,7{,}000,\,10{,}000\}$ examples, each evaluated with $K=16$ rollouts per question as in Table~\ref{tab:main_results}.

Fig.~\ref{fig:ablation_row}(b) uses Qwen2.5-7B-Instruct and compares the sigmoid schedule used in the main experiments against alternative schedules over training step $t$ and total training steps $T$, with $\lambda_{\min}=0.8$ and $\lambda_{\max}=1.2$. We consider (i) a constant weight $\lambda(t)=1.0$, (ii) linear growth:
\[
\lambda(t)=\lambda_{\min}+(\lambda_{\max}-\lambda_{\min})\frac{t}{T},
\]
and (iii) sigmoid-shaped growth:
\[
\lambda(t)=\lambda_{\min}+(\lambda_{\max}-\lambda_{\min})\,\frac{\sigma\!\big(\kappa(t/T-0.5)\big)-\sigma(-\kappa/2)}{\sigma(\kappa/2)-\sigma(-\kappa/2)},
\]
where $\sigma(\cdot)$ denotes the logistic function, $\kappa=10$ controls the slope, and $T=2{,}500$ optimizer steps (10k examples at 4 prompts per step). The normalization makes the sigmoid schedule, like the linear one, start exactly at $\lambda_{\min}$ and end exactly at $\lambda_{\max}$.

Fig.~\ref{fig:ablation_row}(c) evaluates the Llama-3.1-8B-Instruct main-results checkpoint on the HotpotQA evaluation set of Table~\ref{tab:main_results} with $K\in\{8,16,32\}$ rollouts per question. Each budget is a separate evaluation with $K$ rollouts per question.

\subsection{Formal definitions}
\label{app:definitions}

We use standard terminology~\citep{guo2017calibration,geng2024survey,lin2024generating}. \emph{Uncertainty} is a functional of the predictive distribution $\pi_\theta(\cdot\mid x)$~\citep{lin2024generating}, here the semantic entropy of $\pi_\theta(\cdot\mid x)$, which $\mathrm{SE}(x)$ estimates from $K$ rollouts, and is thus a property of the question alone. \emph{Confidence} estimates the probability that a \emph{particular} answer is correct, so it depends on both the input and the evaluated output. Write $\mathbf{Y}=(Y^{(1)},\ldots,Y^{(K)})\sim\pi_\theta(\cdot\mid x)^{K}$ for a rollout group and $\mathbf{Y}^{(-j)}$ for the siblings of $Y^{(j)}$. A response-level estimate $\psi_\phi(x,Y^{(j)};\mathbf{Y}^{(-j)})\in[0,1]$ (the subscript $\phi$ denotes the estimator and any fitted parameters) may also read those siblings. It may be \emph{verbalized} by the model or \emph{derived} post hoc from samples, as ours is. Aggregating the response-level estimates, or reading the group directly as semantic entropy does, gives a \emph{question-level UQ score} $c(x)\in[0,1]$ (Appendix~\ref{app:metrics}). Because it is oriented so that larger values indicate greater expected correctness, it is a question-level \emph{confidence score}, and its finite-sample value depends on the sampled group $\mathbf{Y}$ as well as on $x$. We keep $\mathrm{SE}$ uncertainty-oriented (larger means less certain) and use $\hat s$ as $c(x)$ for Base, RLVR, and CSR.

Let $X$ be a random evaluation question, let $Q=J(Y,Y^{*})\in\{0,1\}$ be the judged correctness of a fresh rollout $Y\sim\pi_\theta(\cdot\mid X)$ against the gold answer $Y^{*}$, and let $\eta(x)=\mathbb{E}_{Y\sim\pi_\theta(\cdot\mid x)}[Q]$ be the policy's correctness probability for the question. A \emph{fixed} post-hoc score is \emph{calibrated} when
\begin{equation}
\mathbb{E}\!\left[\eta(X)\,\middle|\,c(X)=v\right]=v
\qquad\text{for every } v\in[0,1],
\label{eq:calibration_condition}
\end{equation}
and ECE estimates the deviation from this condition, with $\eta(x)$ replaced by the accuracy $\mathrm{Acc}(x)$ of the scored answers (the $K$ rollouts, or the single greedy answer for RD and RLCR)\@. \emph{Discrimination}, ranking correct questions above incorrect ones, is measured by AUROC.

\subsection{Evaluation metrics}
\label{app:metrics}

We evaluate model performance using accuracy, expected calibration error, the area under the receiver operating characteristic curve, two selective-prediction summaries (AURAC and AUGRC), and token cost. All metrics are computed from question-level quantities: for Base, RLVR, and CSR we aggregate the $K$ sampled rollouts of each question, and for RD and RLCR we use the single greedy answer.

\textbf{Accuracy (Acc).} For each generated answer, we use GPT-4.1 nano as an LLM judge to determine whether the generated answer is semantically correct with respect to the gold answer(s). The judge prompts are given in Appendix~\ref{app:system_prompt}. We define question-level accuracy as the average judged correctness $Q_j$ (Sec.~\ref{sec:method}) of the $K$ sampled rollouts:
\begin{equation}
\mathrm{Acc}(x)=\frac{1}{K}\sum_{j=1}^{K} Q_j.
\end{equation}
We then report overall accuracy by averaging this quantity over the evaluation questions. For RD and RLCR, which decode greedily, $K=1$ and $\mathrm{Acc}(x)$ is the $0/1$ correctness of the single answer.

\textbf{Confidence score.}
We write $c(x)\in[0,1]$ for the confidence score of question $x$: the semantic confidence $\hat s(x)$ of Sec.~\ref{sec:prelim} for Base, RLVR, and CSR, computed from the discrete semantic entropy of Eq.~\eqref{eq:se}~\citep{farquhar2024detecting} over semantic-equivalence classes formed from one LLM-judge verdict per unordered rollout pair, and the verbalized confidence for RD and RLCR.

\textbf{Expected calibration error (ECE).} With $c(x)$ and $\mathrm{Acc}(x)$ as above, we partition the $n$ evaluation questions into $B=10$ equal-width bins according to their confidence values, following the binning procedure of \citet{guo2017calibration}. Let $\mathcal{B}_b$ denote the set of questions whose confidence falls into bin $b$, with empty bins contributing zero. The ECE is computed as
\begin{equation}
\mathrm{ECE}
=
\sum_{b=1}^{B}
\frac{|\mathcal{B}_b|}{n}
\left|
\frac{1}{|\mathcal{B}_b|}\sum_{x\in \mathcal{B}_b} \mathrm{Acc}(x)
-
\frac{1}{|\mathcal{B}_b|}\sum_{x\in \mathcal{B}_b} c(x)
\right|.
\end{equation}
Lower ECE indicates better agreement between confidence and observed accuracy.

\textbf{AUROC.} AUROC evaluates how well confidence distinguishes correct from incorrect predictions. Since our question-level accuracy $\mathrm{Acc}(x)$ is a continuous value in $[0,1]$ rather than a binary label, we binarize it using a threshold of $0.5$:
\begin{equation}
\tilde a(x)=\mathbf{1}\!\left[\mathrm{Acc}(x)\ge 0.5\right].
\end{equation}
We then compute AUROC by treating $c(x)$ as the ranking score and $\tilde a(x)\in\{0,1\}$ as the binary correctness label. Formally, AUROC is the probability that a randomly chosen positive example receives a higher confidence score than a randomly chosen negative example, with ties counted as one half:
\begin{equation}
\mathrm{AUROC}
=
\Pr\big(c(x^+) > c(x^-)\big)+\tfrac{1}{2}\Pr\big(c(x^+) = c(x^-)\big),
\end{equation}
where $x^+$ and $x^-$ denote questions with $\tilde a(x^+)=1$ and $\tilde a(x^-)=0$, respectively. A higher AUROC indicates that the confidence score more effectively ranks correct questions above incorrect ones.

\textbf{Selective prediction.} Two risk-coverage summaries are computed from the same confidence score $c(x)$ (Sec.~\ref{sec:selective_main}, Appendix~\ref{app:selective}). Ordering the $n$ evaluation questions by decreasing $c(x)$, the retained set $\mathcal{R}(\rho)$ at coverage $\rho\in(0,1]$ consists of the $\lceil\rho n\rceil$ most confident questions. Because $c(x)$ has many ties (RD, for example, emits only a few distinct values), a coverage level can fall inside a group of tied questions whose order the score does not determine, so we evaluate both integrals below in exact expectation over uniformly random orderings within each tied group.

\par\noindent\textbf{AURAC.} The area under the rejection-accuracy curve summarizes the fraction of retained questions with $\tilde a(x)=1$ (at least half of the scored answers correct) as increasingly uncertain questions are rejected~\citep{farquhar2024detecting}. It is defined as
\begin{equation}
\mathrm{AURAC}
=\int_0^1 \frac{1}{|\mathcal{R}(\rho)|}
  \sum_{x\in\mathcal{R}(\rho)} \tilde a(x)\; d\rho ,
\end{equation}
where higher values are better.

\par\noindent\textbf{AUGRC.} The area under the generalized risk-coverage curve instead integrates the joint probability that a question is both accepted and answered incorrectly~\citep{traub2024overcoming}. It is defined as
\begin{equation}
\mathrm{AUGRC}
=\int_0^1 \frac{1}{n}
  \sum_{x\in\mathcal{R}(\rho)} \bigl(1-\tilde a(x)\bigr)\; d\rho .
\end{equation}
Lower values indicate better selective performance. We report AUGRC on $[0,1]$, without the $\times1000$ display scaling used by \citet{traub2024overcoming}.

\textbf{Token cost (Tok).} We also report the token cost per question, defined as the total number of prompt tokens and generated output tokens:
\begin{equation}
\mathrm{Tok}(x)=T_{\mathrm{prompt}}(x)+T_{\mathrm{output}}(x),
\end{equation}
where $T_{\mathrm{prompt}}(x)$ is the number of prompt tokens for question $x$, and $T_{\mathrm{output}}(x)$ is the number of generated output tokens, summed over all $K$ rollouts ($K=1$ for RD and RLCR), with the prompt counted once. We report Tok by averaging this quantity over evaluation questions.

\subsection{Prompts}
\label{app:system_prompt}
Base and RLVR use the same prompt as CSR.

\definecolor{CSRRed}{HTML}{DC2626}
\begin{tcolorbox}[breakable,
  title=CSR prompt,
  colback=white,
  colframe=CSRRed,
  colbacktitle=CSRRed,
  coltitle=white,
  fonttitle=\bfseries,
  boxrule=1pt,
  arc=2mm,
  left=8pt,
  right=8pt,
  top=4pt,
  bottom=4pt,
]
Provide a concise and direct answer to the question. Avoid unnecessary explanations or additional text.
\end{tcolorbox}

\definecolor{RLCRPurple}{HTML}{9333EA}
\begin{tcolorbox}[breakable,
  title=RLCR prompt,
  colback=white,
  colframe=RLCRPurple,
  colbacktitle=RLCRPurple,
  coltitle=white,
  fonttitle=\bfseries,
  boxrule=1pt,
  arc=2mm,
  left=8pt,
  right=8pt,
  top=4pt,
  bottom=4pt,
]
\raggedright
A conversation between User and Assistant. The user asks a question, and the Assistant solves it. The assistant first thinks about the reasoning process in the mind, provides the user with the final answer, then analyzes its confidence about the solution and then provides the user with its confidence level.

The confidence level is a number between 0 and 1 (inclusive) enclosed within \texttt{\textless confidence\textgreater\textless/confidence\textgreater} tags. The final answer is enclosed between \texttt{\textless answer\textgreater\textless/answer\textgreater} tags. The analysis about confidence and uncertainty is enclosed within \texttt{\textless analysis\textgreater\textless/analysis\textgreater} tags. The assistant should reason about its confidence in the solution and its uncertainty in the solution within these tags.

Here are some guidelines for the analysis:

1. Your task is to point out things where the model could be wrong in its thinking, or things where there might be ambiguity in the solution steps, or in the reasoning process itself.

2. You should not suggest ways of fixing the response; your job is only to reason about uncertainties.

3. For some questions, the response might be correct. In these cases, it is also okay to have only a small number of uncertainties and then explicitly say that you are unable to spot more uncertainties.

4. Uncertainties might be different from errors. For example, uncertainties may arise from ambiguities in the question, or from the application of a particular lemma/proof.

5. If there are alternate potential approaches that may lead to different answers, you should mention them.

6. List out plausible uncertainties; do not make generic statements; be as specific about uncertainties as possible.

7. Enclose this uncertainty analysis within \texttt{\textless analysis\textgreater\textless/analysis\textgreater} tags.

The final format that must be followed is:
\texttt{\textless think\textgreater reasoning process here \textless/think\textgreater\textless answer\textgreater final answer here \textless/answer\textgreater\textless analysis\textgreater analysis about confidence and uncertainty here \textless/analysis\textgreater\textless confidence\textgreater confidence level here (number between 0 and 1) \textless/confidence\textgreater}
\end{tcolorbox}

\definecolor{RDGold}{HTML}{CA8A04}
\begin{tcolorbox}[breakable,
  title=Rewarding Doubt (RD) prompt,
  colback=white,
  colframe=RDGold,
  colbacktitle=RDGold,
  coltitle=white,
  fonttitle=\bfseries,
  boxrule=1pt,
  arc=2mm,
  left=8pt,
  right=8pt,
  top=4pt,
  bottom=4pt,
]
You will get questions. Answer with the correct answer.
Additionally provide a confidence between 0, 1, 2, 3, 4, 5, 6, 7, 8, 9, 10, of how sure you are the answer is correct.
A value close to 0 means you think there is a high probability that the answer is wrong. The closer the value is to 10, the higher you think is the probability that the answer is correct.
The output should have the format \texttt{Answer: \textless answer\textgreater, Confidence: \textless confidence\textgreater} and nothing else.
\end{tcolorbox}

\definecolor{JudgeGray}{HTML}{4B5563}
\begin{tcolorbox}[breakable,
  height fixed for=first and middle,
  title=Judge prompt (semantic equivalence),
  colback=white,
  colframe=JudgeGray,
  colbacktitle=JudgeGray,
  coltitle=white,
  fonttitle=\bfseries,
  boxrule=1pt,
  arc=2mm,
  left=8pt,
  right=8pt,
  top=4pt,
  bottom=4pt,
]
Given a question and two potential answers, you need to analyze whether the two answers convey the same meaning. Respond with ``yes'' or ``no''.\\
Your response should be completely based on the similarity between the two answers.\\
DO NOT use your own knowledge base when do the comparison.\\[4pt]
Question:\\
\{q\}\\
Response 1:\\
\{a\}\\
Response 2:\\
\{b\}\\[4pt]
Your response should use the following format:\\
\texttt{Correct: \textless yes or no\textgreater}
\end{tcolorbox}

\textbf{Judge prompts.} Every call to the LLM judge uses GPT-4.1 nano with the system message ``You are a helpful assistant that helps people find information.'' The semantic-equivalence prompt above produces the pairwise agreement $s_{ji}$ and, with the gold answer in place of \emph{Response 2}, the correctness label $Q_j$ on every dataset.

\section{Analysis of the calibration reward}
\label{app:analysis}

This appendix proves Proposition~\ref{prop:meanfield} and contrasts the calibration reward with an explicit regularizer.

\subsection{Proof of Proposition~\ref{prop:meanfield}}
\label{app:meanfield_proof}

Fix a question $x$ and a rollout group $Y^{(1)},\ldots,Y^{(K)}$ drawn i.i.d.\ from $\pi_\theta(\cdot\mid x)$, and recall from Sec.~\ref{sec:method} and Proposition~\ref{prop:meanfield}
\[
\begin{gathered}
s_{ji}=J\big(Y^{(j)},Y^{(i)}\big),\qquad Q_j=J\big(Y^{(j)},Y^{*}\big),\qquad \alpha=\frac{1}{K}\sum_{j=1}^{K}Q_j,\qquad c_\epsilon=\log\frac{1-\epsilon}{\epsilon},\\
\hat p_j=\frac{1}{K-1}\sum_{i\neq j}s_{ji},\qquad p_j=\mathbb P_{Y'\sim\pi_\theta(\cdot\mid x)}\big(J(Y^{(j)},Y')=1\mid Y^{(j)}\big)
\end{gathered}
\]
that is, the pairwise agreement, the correctness of rollout $j$, the empirical accuracy of the group, the cross-entropy scale, the empirical agreement rate of rollout $j$, and its policy-level counterpart.

\textbf{Step 1 (one pair).} Because $s_{ji},Q_j\in\{0,1\}$, the $\epsilon$-softened summand of Eq.~\eqref{eq:rlsr} takes only two values:
\[
-\mathrm{CE}(s_{ji},Q_j)=
\begin{cases}
\log(1-\epsilon), & s_{ji}=Q_j,\\
\log\epsilon, & s_{ji}\neq Q_j.
\end{cases}
\]
Hence the summand is affine in the indicator of agreement with correctness:
\[
-\mathrm{CE}(s_{ji},Q_j)=\log\epsilon+c_\epsilon\,\mathbf 1[s_{ji}=Q_j].
\]

\textbf{Step 2 (one rollout).} Each unordered pair is judged once and its verdict is shared by both rollouts, so $s_{ji}=s_{ij}$. We assume accordingly that $J$ is symmetric, $J(y,y')=J(y',y)$, so that $s_{ji}=J\big(Y^{(j)},Y^{(i)}\big)$ holds for both orderings of the pair and $p_j$ above is defined with this symmetric judge. For a fixed $j$, the fraction of siblings whose agreement with $Y^{(j)}$ matches $Q_j$ is
\[
\frac{1}{K-1}\sum_{i\neq j}\mathbf 1[s_{ji}=Q_j]=Q_j\hat p_j+(1-Q_j)(1-\hat p_j).
\]
Averaging Step 1 over $i\neq j$ therefore gives
\[
r^{(j)}_{\mathrm{calibration}}=\log\epsilon+c_\epsilon\big[Q_j\hat p_j+(1-Q_j)(1-\hat p_j)\big].
\]
Conditional on $Y^{(j)}$, the other rollouts are i.i.d.\ draws from $\pi_\theta(\cdot\mid x)$, so each $s_{ji}$ with $i\neq j$ has conditional mean $p_j$, and
\[
\mathbb E\big[\hat p_j\mid Y^{(j)}\big]=p_j.
\]

\textbf{Step 3 (the group).} Split the rollouts by correctness into $\mathcal C=\{j:Q_j=1\}$ and $\mathcal I=\{j:Q_j=0\}$, so that $|\mathcal C|=K\alpha$ and $|\mathcal I|=K(1-\alpha)$, and define the mean agreement rates
\[
\begin{aligned}
\bar p_{\mathcal C}&=\frac{1}{|\mathcal C|}\sum_{j\in\mathcal C}\hat p_j,\\
\bar p_{\mathcal I}&=\frac{1}{|\mathcal I|}\sum_{j\in\mathcal I}\hat p_j .
\end{aligned}
\]
Averaging Step 2 over $j$ and splitting the sum accordingly gives
\[
\bar r_{\mathrm{calibration}}
=\log\epsilon+c_\epsilon\Big[\alpha\,\bar p_{\mathcal C}+(1-\alpha)\big(1-\bar p_{\mathcal I}\big)\Big],
\]
where a term whose set is empty is omitted.

\textbf{Consequences.} When $\alpha\to0$, the second term dominates and is maximized by $p_j\to0$. When $\alpha\to1$, the first term dominates and is maximized by $p_j\to1$. Neither regime directly rewards a change from an incorrect to a correct semantic mode. \hfill$\square$

\textbf{Remark (failure of the calibration reward alone).} Consider a group in which rollout $1$ is correct and agrees with no sibling, and the other $K-1$ rollouts are incorrect and mutually disagreeing. Then $\hat p_j=0$ for every $j$, so the normalized reward is
\[
Q_j\hat p_j+(1-Q_j)(1-\hat p_j)=
\begin{cases}
0, & j=1 \text{ (the correct rollout)},\\
1, & j\neq 1 \text{ (every incorrect rollout)}.
\end{cases}
\]
Rollout $1$ thus has the lowest reward in its group and a negative group-relative advantage: the calibration reward alone pushes probability away from the only correct answer.

Coupling the calibration reward with $r_{\mathrm{RLVR}}$ (Eq.~\eqref{eq:rlvr}) adds a unit reward to correct rollouts only. In this group the correct rollout then receives $1$ and each incorrect rollout $\lambda(t)$, so it outranks them whenever $\lambda(t)<1$. The curriculum therefore starts with $\lambda(t)$ below $1$, when accuracy is low and such isolated correct rollouts are common.

\subsection{Calibration reward versus explicit regularization}
\label{app:regularization_view}
The semantic calibration reward is an \emph{auxiliary reward with a regularizing effect} on the policy's semantic answer distribution, but it is not, in general, equivalent to adding an independent regularization term to the loss.

\textbf{Reward-space formulation.}
For one prompt $x$, let $Y^{(1)},\ldots,Y^{(K)}\sim\pi_\theta(\cdot\mid x)$ be its rollout group. For brevity write $r_{\mathrm{R}}^{(j)}=r_{\mathrm{RLVR}}^{(j)}$ and let $r_{\mathrm{C}}^{(j)}$ be the normalized calibration reward used in training (Sec.~\ref{sec:csr_obj} and Eq.~\eqref{eq:calibration_reward_group} below), and define their group-centered versions together with the combined group scale by
\begin{equation}
\begin{aligned}
\widetilde r_{u}^{(j)}
&=r_{u}^{(j)}-\frac{1}{K}\sum_{\ell=1}^{K}r_{u}^{(\ell)},
\qquad u\in\{\mathrm{R},\mathrm{C}\},\\
\sigma_{\mathrm{CSR}}
&=\left[
\frac{1}{K}\sum_{\ell=1}^{K}
\left(\widetilde r_{\mathrm{R}}^{(\ell)}
+\lambda(t)\widetilde r_{\mathrm{C}}^{(\ell)}\right)^{2}
\right]^{1/2}+\delta.
\end{aligned}
\label{eq:csr_centered_rewards}
\end{equation}
CSR combines the two rewards \emph{before} group normalization, yielding
\begin{equation}
A_{\mathrm{CSR}}^{(j)}
=
\frac{\widetilde r_{\mathrm{R}}^{(j)}
+\lambda(t)\widetilde r_{\mathrm{C}}^{(j)}}
{\sigma_{\mathrm{CSR}}}.
\label{eq:csr_advantage}
\end{equation}
Here $\delta>0$ is the usual group-normalization stabilizer, distinct from the cross-entropy softening constant $\epsilon$ in Eq.~\eqref{eq:rlsr}. Equation~\eqref{eq:csr_advantage} shows that the calibration term changes rollout-level credit assignment, rather than adding a batch-level penalty after the RLVR advantage is fixed.

\textbf{Effect within a rollout group.}
Under pure RLVR, two rollouts with the same correctness receive the same reward. To expose the additional ordering induced by CSR, recall that, with $Q_j$ and $\hat p_j$ as in Proposition~\ref{prop:meanfield}, this normalized reward is
\begin{equation}
r_{\mathrm{C}}^{(j)}
=
Q_j\hat p_j+(1-Q_j)(1-\hat p_j).
\label{eq:calibration_reward_group}
\end{equation}
For two rollouts $j$ and $j'$ with equal correctness, the correctness-reward terms cancel exactly. Substituting Eq.~\eqref{eq:calibration_reward_group} into Eq.~\eqref{eq:csr_advantage} gives
\begin{equation}
\begin{aligned}
A_{\mathrm{CSR}}^{(j)}-A_{\mathrm{CSR}}^{(j')}
&=
\frac{\lambda(t)}{\sigma_{\mathrm{CSR}}}
\left(r_{\mathrm{C}}^{(j)}-r_{\mathrm{C}}^{(j')}\right)\\
&=
\frac{\lambda(t)}{\sigma_{\mathrm{CSR}}}
\begin{cases}
\hat p_j-\hat p_{j'},
& Q_j=Q_{j'}=1,\\[6pt]
\hat p_{j'}-\hat p_j,
& Q_j=Q_{j'}=0.
\end{cases}
\end{aligned}
\label{eq:csr_pairwise_credit}
\end{equation}
Thus, among correct rollouts, greater semantic agreement receives greater advantage. Among incorrect rollouts, greater agreement receives \emph{lower} advantage because $1-\hat p$ decreases with $\hat p$. CSR thus concentrates credit on consistent correct answers while suppressing concentrated incorrect modes, which is the sense in which the auxiliary reward has a regularizing effect.

\textbf{Comparison with an explicit regularizer.}
A separate regularization construction would first normalize the correctness reward alone and then add an independent calibration term:
\begin{equation}
\begin{aligned}
A_{\mathrm{R}}^{(j)}
&=\frac{\widetilde r_{\mathrm{R}}^{(j)}}{\sigma_{\mathrm{R}}},
\qquad
\sigma_{\mathrm{R}}
=\left[
\frac{1}{K}\sum_{\ell=1}^{K}
\left(\widetilde r_{\mathrm{R}}^{(\ell)}\right)^2
\right]^{1/2}+\delta,\\
\mathcal{L}_{\mathrm{sep}}
&=
\mathcal{L}_{\mathrm{GRPO}}\!\left(A_{\mathrm{R}}\right)
+\gamma(t)\,\Omega_{\mathrm{C}}
+\beta\,\mathcal{L}_{\mathrm{KL}}.
\end{aligned}
\label{eq:separate_regularizer}
\end{equation}
Here $\gamma(t)\geq 0$ weights a calibration term $\Omega_{\mathrm{C}}$. Because $r_{\mathrm{C}}^{(j)}$ is computed from discrete rollouts and an external judge, $\Omega_{\mathrm{C}}$ must use a score-function surrogate to transmit a policy gradient, $\Omega_{\mathrm{C}}=-\frac{1}{K}\sum_{j}\operatorname{sg}\!\big(\widetilde r_{\mathrm{C}}^{(j)}\big)\log\pi_\theta\!\left(Y^{(j)}\mid x\right)$ with stop-gradient $\operatorname{sg}$, so that $-\nabla_\theta\Omega_{\mathrm{C}}=g_{\mathrm{C}}$, with $g_{\mathrm{C}}$ defined in Eq.~\eqref{eq:reward_gradient_components} below. To compare the two constructions, assume that this surrogate uses the same samples and the same centered calibration rewards as CSR, and define the per-group score-function components
\begin{equation}
g_u
=
\frac{1}{K}\sum_{j=1}^{K}
\widetilde r_u^{(j)}
\nabla_\theta\log\pi_\theta\!\left(Y^{(j)}\mid x\right),
\qquad u\in\{\mathrm{R},\mathrm{C}\}.
\label{eq:reward_gradient_components}
\end{equation}
Ignoring importance-ratio clipping and treating the sampled rewards, judge outputs, and normalization statistics as stop-gradient quantities, the relevant per-group policy-gradient directions, up to a shared positive constant, are
\begin{equation}
\widehat g_{\mathrm{CSR}}
=
\frac{g_{\mathrm{R}}+\lambda(t)g_{\mathrm{C}}}
{\sigma_{\mathrm{CSR}}},
\qquad
\widehat g_{\mathrm{sep}}
=
\frac{g_{\mathrm{R}}}{\sigma_{\mathrm{R}}}
+\gamma(t)g_{\mathrm{C}}.
\label{eq:csr_regularizer_gradients}
\end{equation}
For a single rollout group with nonzero variance, and generically non-collinear $g_{\mathrm{R}}$ and $g_{\mathrm{C}}$, these directions are proportional only if
\begin{equation}
\gamma(t)=\frac{\lambda(t)}{\sigma_{\mathrm{R}}},
\qquad\text{in which case}\qquad
\widehat g_{\mathrm{sep}}
=
\frac{\sigma_{\mathrm{CSR}}}{\sigma_{\mathrm{R}}}
\widehat g_{\mathrm{CSR}}.
\label{eq:regularizer_equivalence_condition}
\end{equation}
This condition is restrictive: $\sigma_{\mathrm{R}}$ varies across prompt groups, whereas $\gamma(t)$ is ordinarily a single global coefficient, and the factor $\sigma_{\mathrm{CSR}}/\sigma_{\mathrm{R}}$ also changes the relative weighting of groups when their updates are aggregated. Importance-ratio clipping introduces an additional mismatch. Consequently, a fixed additive regularizer does not generally reproduce CSR over a training batch.

\textbf{Choice in CSR.}
CSR uses the reward-space construction in Eq.~\eqref{eq:csr_advantage}, so the semantic structure of grouped rollouts enters before group normalization and directly changes rollout-level credit assignment. The KL term $\beta\,\mathcal{L}_{\mathrm{KL}}$ remains the explicit policy regularizer.

\section{Supplementary empirical analyses}
\label{app:supp_analyses}

\newcommand{\boldanswer}[1]{\mbox{\rmfamily\textbf{#1}}}
\newcommand{\metrics}[2]{%
  \par\nopagebreak\vspace{1.5pt}\nopagebreak\normalfont\noindent\rule{\linewidth}{0.4pt}\par\nopagebreak\vspace{1.5pt}\nopagebreak%
  \noindent\textbf{Accuracy:} #1 \quad \textbf{Confidence:} #2%
}
\definecolor{CSRRed}{HTML}{DC2626}
\definecolor{BaseBlue}{HTML}{2563EB}
\definecolor{RLVRGreen}{HTML}{16A34A}
\definecolor{RDGold}{HTML}{CA8A04}
\definecolor{RLCRPurple}{HTML}{9333EA}

\subsection{Transfer to other confidence estimators}
\label{app:alt_estimators}
\begin{table}[!t]
\centering
\small
\setlength{\tabcolsep}{3.0pt}
\renewcommand{\arraystretch}{1.1}
\caption{Three aggregations from SIMBA UQ~\citep{bhattacharjya2025simba} on the rollouts of Table~\ref{tab:main_results}: Bayesian aggregation (Bayes-beta), unsupervised arithmetic aggregation (Arithmetic), and the supervised Clf-pairs+gen classifier. \textbf{Bold} marks the best and \underline{underline} the second-best value in each column within each model block.}
\label{tab:alt_estimators_full}
\resizebox{\textwidth}{!}{%
\begin{tabular}{@{}c@{}}
\begin{tabular}{l|l c c|c c|c c|c c|c c}
\toprule
\multicolumn{12}{c}{\textbf{Bayesian aggregation (Bayes-beta, supervised)}}\\
\midrule
\multicolumn{1}{l}{} & & \multicolumn{2}{c}{\textbf{In-distribution}} & \multicolumn{2}{c}{} & \multicolumn{2}{c}{\textbf{Out-of-distribution}} & \multicolumn{2}{c}{} & \multicolumn{2}{c}{}\\
\cmidrule(lr){3-4}\cmidrule(lr){5-10}
\multicolumn{1}{l}{} & & \multicolumn{2}{c}{\textbf{HotpotQA}} & \multicolumn{2}{c}{\textbf{TriviaQA}} & \multicolumn{2}{c}{\textbf{MSMARCO}} & \multicolumn{2}{c}{\textbf{NQ-Open}} & \multicolumn{2}{c}{\textbf{Average}}\\
\cmidrule(lr){3-4}\cmidrule(lr){5-6}\cmidrule(lr){7-8}\cmidrule(lr){9-10}\cmidrule(lr){11-12}
\textbf{Model} & \textbf{Method} & ECE$\downarrow$ & AUROC$\uparrow$ & ECE$\downarrow$ & AUROC$\uparrow$ & ECE$\downarrow$ & AUROC$\uparrow$ & ECE$\downarrow$ & AUROC$\uparrow$ & ECE$\downarrow$ & AUROC$\uparrow$\\
\midrule
\multirow{3}{*}{Llama}
& Base  & \underline{0.2034} & 0.7575 & \underline{0.1492} & \underline{0.8283} & \underline{0.2391} & 0.5841 & \underline{0.1881} & \underline{0.7028} & \underline{0.1950} & 0.7182\\
& RLVR  & 0.2084 & \underline{0.8152} & 0.2066 & 0.7946 & 0.2502 & \underline{0.6392} & 0.2343 & 0.6797 & 0.2249 & \underline{0.7322}\\
\rowcolor{gray!12} \cellcolor{white}& CSR   & \textbf{0.1440} & \textbf{0.8889} & \textbf{0.1278} & \textbf{0.8699} & \textbf{0.1882} & \textbf{0.6531} & \textbf{0.1741} & \textbf{0.7460} & \textbf{0.1585} & \textbf{0.7895}\\
\midrule
\multirow{3}{*}{Qwen}
& Base  & \underline{0.2330} & \underline{0.8233} & \underline{0.4971} & 0.8152 & \underline{0.2546} & 0.6256 & 0.2340 & \underline{0.7595} & \underline{0.3047} & \underline{0.7559}\\
& RLVR  & 0.2550 & 0.8190 & 0.4996 & \underline{0.8163} & 0.2701 & \underline{0.6351} & \underline{0.2333} & 0.7406 & 0.3145 & 0.7528\\
\rowcolor{gray!12} \cellcolor{white}& CSR   & \textbf{0.1260} & \textbf{0.9191} & \textbf{0.1529} & \textbf{0.9079} & \textbf{0.2138} & \textbf{0.6759} & \textbf{0.1760} & \textbf{0.8501} & \textbf{0.1672} & \textbf{0.8383}\\
\midrule
\multirow{3}{*}{Mistral}
& Base  & 0.2199 & 0.7015 & 0.2318 & 0.7026 & \textbf{0.1853} & \textbf{0.5943} & 0.2487 & \underline{0.6460} & 0.2214 & 0.6611\\
& RLVR  & \underline{0.2158} & \underline{0.7380} & \underline{0.1929} & \underline{0.7491} & 0.2215 & 0.5779 & \underline{0.2373} & \textbf{0.6638} & \underline{0.2169} & \underline{0.6822}\\
\rowcolor{gray!12} \cellcolor{white}& CSR   & \textbf{0.1576} & \textbf{0.7974} & \textbf{0.1892} & \textbf{0.7518} & \underline{0.2187} & \underline{0.5781} & \textbf{0.2288} & 0.6266 & \textbf{0.1986} & \textbf{0.6885}\\
\bottomrule
\end{tabular}\\[1.2ex]
\begin{tabular}{l|l c c|c c|c c|c c|c c}
\toprule
\multicolumn{12}{c}{\textbf{Arithmetic aggregation (Arithmetic, unsupervised)}}\\
\midrule
\multicolumn{1}{l}{} & & \multicolumn{2}{c}{\textbf{In-distribution}} & \multicolumn{2}{c}{} & \multicolumn{2}{c}{\textbf{Out-of-distribution}} & \multicolumn{2}{c}{} & \multicolumn{2}{c}{}\\
\cmidrule(lr){3-4}\cmidrule(lr){5-10}
\multicolumn{1}{l}{} & & \multicolumn{2}{c}{\textbf{HotpotQA}} & \multicolumn{2}{c}{\textbf{TriviaQA}} & \multicolumn{2}{c}{\textbf{MSMARCO}} & \multicolumn{2}{c}{\textbf{NQ-Open}} & \multicolumn{2}{c}{\textbf{Average}}\\
\cmidrule(lr){3-4}\cmidrule(lr){5-6}\cmidrule(lr){7-8}\cmidrule(lr){9-10}\cmidrule(lr){11-12}
\textbf{Model} & \textbf{Method} & ECE$\downarrow$ & AUROC$\uparrow$ & ECE$\downarrow$ & AUROC$\uparrow$ & ECE$\downarrow$ & AUROC$\uparrow$ & ECE$\downarrow$ & AUROC$\uparrow$ & ECE$\downarrow$ & AUROC$\uparrow$\\
\midrule
\multirow{3}{*}{Llama}
& Base  & \underline{0.2525} & 0.7165 & \underline{0.0810} & \underline{0.8214} & \underline{0.0692} & 0.6025 & \underline{0.0458} & \underline{0.7274} & \underline{0.1121} & 0.7169\\
& RLVR  & 0.2725 & \underline{0.8011} & \textbf{0.0629} & 0.7991 & 0.1545 & \underline{0.6628} & 0.1166 & 0.6881 & 0.1516 & \underline{0.7378}\\
\rowcolor{gray!12} \cellcolor{white}& CSR   & \textbf{0.1323} & \textbf{0.8789} & 0.0903 & \textbf{0.8759} & \textbf{0.0444} & \textbf{0.6829} & \textbf{0.0355} & \textbf{0.7562} & \textbf{0.0756} & \textbf{0.7985}\\
\midrule
\multirow{3}{*}{Qwen}
& Base  & \underline{0.3430} & \underline{0.8208} & 0.2043 & \underline{0.8182} & \underline{0.2861} & 0.6330 & 0.3072 & \underline{0.7530} & 0.2851 & 0.7562\\
& RLVR  & 0.3436 & 0.8201 & \underline{0.2015} & 0.8178 & 0.2863 & \underline{0.6476} & \underline{0.2878} & 0.7476 & \underline{0.2798} & \underline{0.7583}\\
\rowcolor{gray!12} \cellcolor{white}& CSR   & \textbf{0.1339} & \textbf{0.9227} & \textbf{0.0799} & \textbf{0.9111} & \textbf{0.1223} & \textbf{0.6890} & \textbf{0.0891} & \textbf{0.8483} & \textbf{0.1063} & \textbf{0.8428}\\
\midrule
\multirow{3}{*}{Mistral}
& Base  & 0.3059 & 0.7037 & \underline{0.1058} & 0.6866 & \underline{0.0546} & \underline{0.6024} & \underline{0.0856} & \textbf{0.6691} & 0.1380 & 0.6655\\
& RLVR  & \underline{0.2873} & \underline{0.7462} & \textbf{0.0786} & \underline{0.7398} & 0.0706 & 0.5990 & 0.1067 & \underline{0.6690} & \underline{0.1358} & \underline{0.6885}\\
\rowcolor{gray!12} \cellcolor{white}& CSR   & \textbf{0.1891} & \textbf{0.7694} & 0.1699 & \textbf{0.7450} & \textbf{0.0342} & \textbf{0.6112} & \textbf{0.0358} & 0.6414 & \textbf{0.1072} & \textbf{0.6918}\\
\bottomrule
\end{tabular}\\[1.2ex]
\begin{tabular}{l|l c c|c c|c c|c c|c c}
\toprule
\multicolumn{12}{c}{\textbf{Random-forest classifier (Clf-pairs+gen, supervised)}}\\
\midrule
\multicolumn{1}{l}{} & & \multicolumn{2}{c}{\textbf{In-distribution}} & \multicolumn{2}{c}{} & \multicolumn{2}{c}{\textbf{Out-of-distribution}} & \multicolumn{2}{c}{} & \multicolumn{2}{c}{}\\
\cmidrule(lr){3-4}\cmidrule(lr){5-10}
\multicolumn{1}{l}{} & & \multicolumn{2}{c}{\textbf{HotpotQA}} & \multicolumn{2}{c}{\textbf{TriviaQA}} & \multicolumn{2}{c}{\textbf{MSMARCO}} & \multicolumn{2}{c}{\textbf{NQ-Open}} & \multicolumn{2}{c}{\textbf{Average}}\\
\cmidrule(lr){3-4}\cmidrule(lr){5-6}\cmidrule(lr){7-8}\cmidrule(lr){9-10}\cmidrule(lr){11-12}
\textbf{Model} & \textbf{Method} & ECE$\downarrow$ & AUROC$\uparrow$ & ECE$\downarrow$ & AUROC$\uparrow$ & ECE$\downarrow$ & AUROC$\uparrow$ & ECE$\downarrow$ & AUROC$\uparrow$ & ECE$\downarrow$ & AUROC$\uparrow$\\
\midrule
\multirow{3}{*}{Llama}
& Base  & \textbf{0.0200} & \underline{0.8062} & \underline{0.0302} & \underline{0.8413} & \textbf{0.0245} & 0.6127 & \underline{0.0401} & \underline{0.7387} & \textbf{0.0287} & \underline{0.7497}\\
& RLVR  & 0.0808 & 0.7624 & 0.0694 & 0.7847 & 0.0464 & \underline{0.6270} & 0.0505 & 0.6632 & 0.0618 & 0.7093\\
\rowcolor{gray!12} \cellcolor{white}& CSR   & \underline{0.0465} & \textbf{0.8622} & \textbf{0.0205} & \textbf{0.8686} & \underline{0.0329} & \textbf{0.6940} & \textbf{0.0248} & \textbf{0.7428} & \underline{0.0312} & \textbf{0.7919}\\
\midrule
\multirow{3}{*}{Qwen}
& Base  & \underline{0.1113} & \underline{0.7735} & \underline{0.1454} & \underline{0.7657} & \underline{0.0937} & \underline{0.5763} & 0.1108 & \underline{0.6795} & \underline{0.1153} & \underline{0.6987}\\
& RLVR  & 0.1530 & 0.7115 & 0.1499 & 0.7507 & 0.1054 & \underline{0.5763} & \underline{0.1095} & 0.6639 & 0.1294 & 0.6756\\
\rowcolor{gray!12} \cellcolor{white}& CSR   & \textbf{0.0848} & \textbf{0.8688} & \textbf{0.0795} & \textbf{0.8833} & \textbf{0.0304} & \textbf{0.6941} & \textbf{0.0474} & \textbf{0.8179} & \textbf{0.0605} & \textbf{0.8160}\\
\midrule
\multirow{3}{*}{Mistral}
& Base  & \underline{0.0390} & 0.6834 & \underline{0.0574} & 0.6860 & 0.0225 & \underline{0.5955} & 0.0364 & 0.6571 & \underline{0.0388} & 0.6555\\
& RLVR  & 0.0690 & \underline{0.6839} & 0.0726 & \underline{0.7134} & \underline{0.0155} & 0.5887 & \underline{0.0305} & \textbf{0.6702} & 0.0469 & \underline{0.6641}\\
\rowcolor{gray!12} \cellcolor{white}& CSR   & \textbf{0.0357} & \textbf{0.7991} & \textbf{0.0459} & \textbf{0.7791} & \textbf{0.0120} & \textbf{0.6029} & \textbf{0.0173} & \underline{0.6696} & \textbf{0.0277} & \textbf{0.7127}\\
\bottomrule
\end{tabular}
\end{tabular}%
}
\end{table}
This section defines the four post-hoc estimators of Sec.~\ref{sec:beyond_se} and reports their per-dataset results on the rollouts of Table~\ref{tab:main_results} (Tables~\ref{tab:alt_estimators_full} and~\ref{tab:alt_semdens}). Each estimator is a response-level estimate $\psi_j=\psi_\phi(x,Y^{(j)};\mathbf{Y}^{(-j)})\in[0,1]$ in the sense of Appendix~\ref{app:definitions}, and the question-level confidence of Appendix~\ref{app:metrics} is its average over the $K$ rollouts, $c(x)=K^{-1}\sum_{j=1}^{K}\psi_j$. We write $\mathrm{Jac}_{ji}$ for the word-level Jaccard similarity between rollouts $Y^{(j)}$ and $Y^{(i)}$, $\bar\pi_j=\exp\!\big(|Y^{(j)}|^{-1}\log\pi_\theta(Y^{(j)}\mid x)\big)$ for the length-normalized sequence probability of $Y^{(j)}$, and $\mathcal M(Y)$ for the semantic-equivalence class containing $Y$ (Sec.~\ref{sec:prelim}).

\textbf{Semantic Density} (SemDens)~\citep{qiu2024semantic} scores a rollout by its probability-weighted density among the set $\mathcal U$ of distinct answers in the group (the rollout itself included, so $|\mathcal U|\le K$),
\begin{equation}
\psi_j=\frac{\sum_{i\in\mathcal U}\bar\pi_i\,\mathbf 1\big[Y^{(i)}\in\mathcal M(Y^{(j)})\big]}{\sum_{i\in\mathcal U}\bar\pi_i},
\end{equation}
computed as in the official implementation, except that the references are the temperature-sampled rollouts and the kernel is the same-class indicator of our semantic-equivalence judge in place of the original NLI-based similarity. The outer average in $c(x)$ runs over all $K$ rollouts, duplicates included, whereas the inner normalization runs over $\mathcal U$.

The other three are aggregations from SIMBA UQ~\citep{bhattacharjya2025simba}. The \textbf{Bayesian aggregation} (Bayes-beta) is the posterior probability that a rollout is correct given its similarities to the others, under Beta likelihoods $f_1$ and $f_0$ for correct and incorrect rollouts and a prior $p_0$,
\begin{equation}
\psi_j=\frac{p_0\prod_{i\neq j}f_1(\mathrm{Jac}_{ji})}{p_0\prod_{i\neq j}f_1(\mathrm{Jac}_{ji})+(1-p_0)\prod_{i\neq j}f_0(\mathrm{Jac}_{ji})}.
\end{equation}
The unsupervised \textbf{arithmetic aggregation} (Arithmetic) averages the similarities to the $K-1$ siblings, $\psi_j=(K-1)^{-1}\sum_{i\neq j}\mathrm{Jac}_{ji}$, so that $c(x)$ is the group's mean pairwise Jaccard similarity. The supervised \textbf{Clf-pairs+gen} classifier is a random forest $h$ over the similarities and the generative score $\bar\pi_j$,
\begin{equation}
\psi_j=h\big((\mathrm{Jac}_{ji})_{i\neq j},\,\bar\pi_j\big).
\end{equation}
The five parameters of Bayes-beta and the random forest are fitted \emph{per policy} with question-grouped five-fold cross-fitting, so every reported confidence is an out-of-fold score.

\begin{table}[!t]
\centering
\small
\setlength{\tabcolsep}{3.0pt}
\renewcommand{\arraystretch}{1.1}
\caption{Semantic Density (SemDens)~\citep{qiu2024semantic} on the rollouts of Table~\ref{tab:main_results}. \textbf{Bold} marks the best and \underline{underline} the second-best value in each column within each model block.}
\label{tab:alt_semdens}
\resizebox{\textwidth}{!}{%
\begin{tabular}{l|l c c|c c|c c|c c|c c}
\toprule
\multicolumn{1}{l}{} & & \multicolumn{2}{c}{\textbf{In-distribution}} & \multicolumn{2}{c}{} & \multicolumn{2}{c}{\textbf{Out-of-distribution}} & \multicolumn{2}{c}{} & \multicolumn{2}{c}{}\\
\cmidrule(lr){3-4}\cmidrule(lr){5-10}
\multicolumn{1}{l}{} & & \multicolumn{2}{c}{\textbf{HotpotQA}} & \multicolumn{2}{c}{\textbf{TriviaQA}} & \multicolumn{2}{c}{\textbf{MSMARCO}} & \multicolumn{2}{c}{\textbf{NQ-Open}} & \multicolumn{2}{c}{\textbf{Average}}\\
\cmidrule(lr){3-4}\cmidrule(lr){5-6}\cmidrule(lr){7-8}\cmidrule(lr){9-10}\cmidrule(lr){11-12}
\textbf{Model} & \textbf{Method} & ECE$\downarrow$ & AUROC$\uparrow$ & ECE$\downarrow$ & AUROC$\uparrow$ & ECE$\downarrow$ & AUROC$\uparrow$ & ECE$\downarrow$ & AUROC$\uparrow$ & ECE$\downarrow$ & AUROC$\uparrow$\\
\midrule
\multirow{3}{*}{Llama}
& Base  & 0.3418 & 0.6928 & \underline{0.0656} & \underline{0.8420} & \underline{0.2542} & 0.7061 & \underline{0.1378} & \underline{0.7342} & \underline{0.1998} & 0.7438\\
& RLVR  & \underline{0.2832} & \underline{0.8073} & 0.0902 & 0.8295 & 0.2707 & \underline{0.7332} & 0.1919 & 0.6997 & 0.2090 & \underline{0.7674}\\
\rowcolor{gray!12} \cellcolor{white}& CSR   & \textbf{0.1213} & \textbf{0.9173} & \textbf{0.0509} & \textbf{0.9039} & \textbf{0.2140} & \textbf{0.7574} & \textbf{0.1143} & \textbf{0.7727} & \textbf{0.1251} & \textbf{0.8378}\\
\midrule
\multirow{3}{*}{Qwen}
& Base  & 0.2901 & \underline{0.8407} & \underline{0.1776} & \underline{0.8288} & 0.3038 & 0.7030 & \underline{0.2400} & \underline{0.7765} & 0.2529 & \underline{0.7872}\\
& RLVR  & \underline{0.2857} & 0.8360 & 0.1835 & 0.8252 & \underline{0.2994} & \underline{0.7067} & 0.2407 & 0.7601 & \underline{0.2523} & 0.7820\\
\rowcolor{gray!12} \cellcolor{white}& CSR   & \textbf{0.1170} & \textbf{0.9309} & \textbf{0.0810} & \textbf{0.9096} & \textbf{0.1843} & \textbf{0.8009} & \textbf{0.0957} & \textbf{0.8348} & \textbf{0.1195} & \textbf{0.8691}\\
\midrule
\multirow{3}{*}{Mistral}
& Base  & 0.2885 & 0.7715 & \underline{0.1066} & 0.7854 & 0.3502 & \underline{0.7075} & \underline{0.1873} & 0.7016 & \underline{0.2331} & 0.7415\\
& RLVR  & \underline{0.2753} & \underline{0.7932} & 0.1222 & \underline{0.7907} & \underline{0.3425} & 0.6979 & 0.1944 & \underline{0.7078} & 0.2336 & \underline{0.7474}\\
\rowcolor{gray!12} \cellcolor{white}& CSR   & \textbf{0.1239} & \textbf{0.8647} & \textbf{0.0703} & \textbf{0.8561} & \textbf{0.3019} & \textbf{0.7364} & \textbf{0.1108} & \textbf{0.7624} & \textbf{0.1517} & \textbf{0.8049}\\
\bottomrule
\end{tabular}%
}
\end{table}

\textbf{Results.} The per-dataset values are consistent with the averages of Table~\ref{tab:beyond_se}. CSR is not best in 9 of the 96 per-dataset cells (3 on Llama, 6 on Mistral), none on Qwen. RLVR, by contrast, is often no better than Base, so correctness training alone does not produce the effect.

\subsection{Quantifying the accuracy-calibration trade-off}
\label{app:tradeoff}

The calibration reward alone does not reward correctness (Proposition~\ref{prop:meanfield}, Fig.~\ref{fig:ablation_row}(a)), so the combined objective may cost accuracy. From Table~\ref{tab:main_results} we compute, for each method and each metric, the relative change against Base, averaged over the three model families, and split into in-distribution (HotpotQA) and out-of-distribution (computed on the metric averaged over the other three datasets). We use $(\text{method}-\text{Base})/\text{Base}$ for accuracy and AUROC and the relative ECE reduction $(\text{Base}-\text{method})/\text{Base}$, so that a positive value always indicates improvement (Table~\ref{tab:tradeoff}).

\begin{table}[!t]
\centering
\small
\setlength{\tabcolsep}{3.0pt}
\renewcommand{\arraystretch}{1.1}
\caption{Relative change (\%) against Base, computed from Table~\ref{tab:main_results} and averaged over the three model families. ECE red.\ is the relative ECE reduction, so positive values indicate improvement. \textbf{Bold} marks the best and \underline{underline} the second-best value in each column.}
\label{tab:tradeoff}
\begin{tabular}{l|c c c|c c c}
\toprule
\multicolumn{1}{l}{} & \multicolumn{3}{c}{\textbf{In-distribution} (HotpotQA)} & \multicolumn{3}{c}{\textbf{Out-of-distribution} (avg.\ of 3)}\\
\cmidrule(lr){2-4}\cmidrule(lr){5-7}
\textbf{Method} & Acc$\uparrow$ & ECE red.$\uparrow$ & AUROC$\uparrow$ & Acc$\uparrow$ & ECE red.$\uparrow$ & AUROC$\uparrow$\\
\midrule
RLVR   & \textbf{+32.1} & +5.2 & +5.5 & \textbf{+0.3} & \textminus9.0 & \textminus0.6\\
RD     & +13.3 & \underline{+49.5} & \underline{+11.0} & \textminus16.9 & \underline{+7.0} & \underline{+0.1}\\
RLCR   & +20.8 & +26.7 & \textminus8.5 & \textminus2.5 & +2.4 & \textminus8.6\\
\rowcolor{gray!12} CSR    & \underline{+22.7} & \textbf{+63.8} & \textbf{+17.4} & \underline{\textminus1.1} & \textbf{+31.1} & \textbf{+7.7}\\
\bottomrule
\end{tabular}
\end{table}

\textbf{Results.} CSR improves ECE and AUROC over Base by the largest margin in both settings, reducing ECE by $63.8\%$ in distribution and $31.1\%$ out of distribution. CSR costs some accuracy relative to RLVR, and the cost is modest. In distribution, CSR raises accuracy over Base by $22.7\%$ and RLVR by $32.1\%$: measured as relative gains over Base averaged over the three backbones, CSR keeps about 70\% of RLVR's accuracy gain. Out of distribution, neither method changes accuracy much: CSR lowers it by $1.1\%$ and RLVR raises it by $0.3\%$.

\subsection{Selective prediction}
\label{app:selective}
Per-dataset AURAC and AUGRC (Appendix~\ref{app:metrics}) are reported in Table~\ref{tab:selective}, together with the four-dataset averages of Table~\ref{tab:selective_main}.

\begin{table}[!t]
\centering
\small
\setlength{\tabcolsep}{3.0pt}
\renewcommand{\arraystretch}{1.1}
\caption{Per-dataset selective prediction with the four-dataset average. Confidence is verbalized for RD and RLCR and the semantic confidence for the other methods. \textbf{Bold} marks the best and \underline{underline} the second-best value in each column within each model block.}
\label{tab:selective}
\resizebox{\textwidth}{!}{%
\begin{tabular}{l|l c c|c c|c c|c c|c c}
\toprule
\multicolumn{1}{l}{} & & \multicolumn{2}{c}{\textbf{In-distribution}} & \multicolumn{2}{c}{} & \multicolumn{2}{c}{\textbf{Out-of-distribution}} & \multicolumn{2}{c}{} & \multicolumn{2}{c}{}\\
\cmidrule(lr){3-4}\cmidrule(lr){5-10}
\multicolumn{1}{l}{} & & \multicolumn{2}{c}{\textbf{HotpotQA}} & \multicolumn{2}{c}{\textbf{TriviaQA}} & \multicolumn{2}{c}{\textbf{MSMARCO}} & \multicolumn{2}{c}{\textbf{NQ-Open}} & \multicolumn{2}{c}{\textbf{Average}}\\
\cmidrule(lr){3-4}\cmidrule(lr){5-6}\cmidrule(lr){7-8}\cmidrule(lr){9-10}\cmidrule(lr){11-12}
\textbf{Model} & \textbf{Method} & AURAC$\uparrow$ & AUGRC$\downarrow$ & AURAC$\uparrow$ & AUGRC$\downarrow$ & AURAC$\uparrow$ & AUGRC$\downarrow$ & AURAC$\uparrow$ & AUGRC$\downarrow$ & AURAC$\uparrow$ & AUGRC$\downarrow$\\
\midrule
\multirow{5}{*}{Llama}
& Base & 0.2716 & 0.3783 & \underline{0.8621} & 0.0853 & 0.5053 & \textbf{0.2672} & \underline{0.6366} & \underline{0.2121} & 0.5689 & 0.2357\\
& RLVR & \underline{0.5023} & \underline{0.2811} & 0.8578 & \underline{0.0838} & \textbf{0.5150} & \underline{0.2674} & 0.6128 & 0.2160 & \underline{0.6220} & \underline{0.2121}\\
& RD & 0.4668 & 0.3085 & 0.8280 & 0.1029 & 0.4237 & 0.3128 & 0.5496 & 0.2670 & 0.5670 & 0.2478\\
& RLCR & 0.4103 & 0.3204 & 0.7866 & 0.1224 & 0.4026 & 0.3157 & 0.5458 & 0.2547 & 0.5363 & 0.2533\\
\rowcolor{gray!12} \cellcolor{white}& CSR & \textbf{0.5060} & \textbf{0.2779} & \textbf{0.8840} & \textbf{0.0791} & \underline{0.5083} & 0.2745 & \textbf{0.6453} & \textbf{0.2099} & \textbf{0.6359} & \textbf{0.2104}\\
\midrule
\multirow{5}{*}{Qwen}
& Base & 0.4170 & 0.3229 & 0.7137 & 0.1662 & 0.4348 & 0.2972 & 0.4287 & 0.3158 & 0.4986 & 0.2755\\
& RLVR & 0.4445 & \underline{0.3089} & 0.7150 & 0.1652 & 0.4435 & \underline{0.2942} & 0.4248 & 0.3149 & \underline{0.5070} & \underline{0.2708}\\
& RD & \underline{0.4520} & 0.3187 & 0.7200 & 0.1650 & 0.4049 & 0.3199 & \underline{0.4441} & 0.3171 & 0.5053 & 0.2802\\
& RLCR & 0.3545 & 0.3534 & \underline{0.7236} & \underline{0.1595} & \underline{0.4501} & 0.2957 & 0.4109 & \underline{0.3108} & 0.4848 & 0.2798\\
\rowcolor{gray!12} \cellcolor{white}& CSR & \textbf{0.4686} & \textbf{0.3020} & \textbf{0.7680} & \textbf{0.1514} & \textbf{0.5039} & \textbf{0.2793} & \textbf{0.4780} & \textbf{0.3079} & \textbf{0.5546} & \textbf{0.2602}\\
\midrule
\multirow{5}{*}{Mistral}
& Base & 0.3991 & 0.3278 & \underline{0.8288} & \underline{0.0967} & 0.5009 & 0.2672 & 0.5745 & \underline{0.2365} & 0.5758 & 0.2320\\
& RLVR & 0.4893 & \underline{0.2838} & 0.8277 & 0.0973 & \underline{0.5026} & \underline{0.2664} & \underline{0.5771} & \underline{0.2365} & \underline{0.5992} & \underline{0.2210}\\
& RD & \underline{0.4978} & 0.3020 & 0.7932 & 0.1485 & 0.2933 & 0.3832 & 0.3717 & 0.3588 & 0.4890 & 0.2981\\
& RLCR & 0.4130 & 0.3121 & 0.7555 & 0.1402 & 0.4008 & 0.3110 & 0.5157 & 0.2611 & 0.5212 & 0.2561\\
\rowcolor{gray!12} \cellcolor{white}& CSR & \textbf{0.5309} & \textbf{0.2763} & \textbf{0.8710} & \textbf{0.0815} & \textbf{0.5279} & \textbf{0.2594} & \textbf{0.6273} & \textbf{0.2177} & \textbf{0.6393} & \textbf{0.2087}\\
\bottomrule
\end{tabular}%
}

\end{table}

\textbf{Results.} CSR is best in 22 of the 24 per-dataset cells and has the best four-dataset average of both metrics on every backbone. Its lead in average AURAC over the strongest baseline, RLVR, ranges from $0.014$ on Llama to $0.048$ on Qwen. The verbalized baselines rank questions less reliably: RLCR is worse than Base on both metrics on every backbone, and RD has the lowest average AURAC on Mistral, where it often abstains.

\textbf{Relation to output-based calibration and refusal.} Output-based signals, such as verbalized confidence or knowledge-aware refusal~\citep{pan2026refuse}, are read from what the model emits for a single generation, so methods that train them act on that reporting behavior. CSR reads confidence from the distribution of $K$ sampled answers, so training acts on the answer distribution itself. Both support abstention or deferral when the model is likely to be wrong. The output-based signal needs only one generation but can vary across semantically equivalent outputs (Fig.~\ref{fig:fig1_mismatch_separation}), whereas the semantic confidence needs $K$ samples and does not depend on a reported confidence. The results above show better confidence-thresholded selective prediction, and we do not claim that CSR improves spontaneous refusal or verbalized confidence.

\subsection{Context-grounded multi-hop QA}
\label{app:csr_reasoning_tasks}
\textbf{Setup.} Table~\ref{tab:main_results} is closed-book, so the model must recall each answer. Here the supporting passages are supplied in training and evaluation, so the answer can be composed from evidence, which is also the setting RLCR is trained in. Every method sees the full paragraph set at evaluation. Base, RLVR, and CSR use the system prompt ``Provide a concise and direct answer to the question. Use the provided context and reason step-by-step. End with \texttt{Final answer: \textless answer\textgreater}.'', RD uses \texttt{Context: \textless passages\textgreater, Question: \textless question\textgreater} as its user turn, and RLCR keeps its own format. Training uses HotpotQA only. MuSiQue~\citep{trivedi2022musique} and the compositional questions of 2WikiMultiHopQA~\citep{ho2020constructing} serve as out-of-distribution test sets, presented in the same way with their own supporting and distractor paragraphs. For each, we shuffle the validation split with seed 42 and use the first $1{,}000$ questions, for 2WikiMultiHopQA after keeping only the compositional questions. CSR and RLVR are retrained on HotpotQA with the paragraphs in the prompt, a maximum prompt length of 3072 tokens, and a completion budget of 1024 tokens. RD is retrained on the same data with no separate prompt cap, only a maximum sequence length of 4096 tokens. RLCR is the checkpoint of Table~\ref{tab:main_results}, which is already trained with paragraphs, so the two tables differ only in whether it is evaluated with them. Evaluation uses a maximum sequence length of 4096 tokens on HotpotQA, 6144 on MuSiQue, and 5120 on 2WikiMultiHopQA, whose passages are longer, and the generation budgets of Appendix~\ref{app:exp_setup}.

\begin{table}[!t]
\centering
\small
\setlength{\tabcolsep}{1.8pt}
\renewcommand{\arraystretch}{1.1}
\caption{Context-grounded QA with Qwen2.5-7B-Instruct, trained on HotpotQA and evaluated on HotpotQA (in-distribution) and on MuSiQue and 2WikiMultiHopQA (out-of-distribution). Average is the macro-average over the three datasets. \textbf{Bold} marks the best and \underline{underline} the second-best value in each column.}
\label{tab:csr_reasoning_tasks}
\begin{tabular}{l|c c c|c c c|c c c|c c}
\toprule
\multicolumn{1}{l}{} & \multicolumn{3}{c}{\textbf{In-distribution}} & \multicolumn{6}{c}{\textbf{Out-of-distribution}} & \multicolumn{2}{c}{}\\
\cmidrule(lr){2-4}\cmidrule(lr){5-10}
\multicolumn{1}{l}{} & \multicolumn{3}{c}{\textbf{HotpotQA}} & \multicolumn{3}{c}{\textbf{MuSiQue}} & \multicolumn{3}{c}{\textbf{2WikiMultiHopQA}} & \multicolumn{2}{c}{\textbf{Average}}\\
\cmidrule(lr){2-4}\cmidrule(lr){5-7}\cmidrule(lr){8-10}\cmidrule(lr){11-12}
\textbf{Method} & Acc$\uparrow$ & ECE$\downarrow$ & AUROC$\uparrow$ & Acc$\uparrow$ & ECE$\downarrow$ & AUROC$\uparrow$ & Acc$\uparrow$ & ECE$\downarrow$ & AUROC$\uparrow$ & ECE$\downarrow$ & AUROC$\uparrow$\\
\midrule
Base & 0.6725 & 0.0751 & 0.7588 & 0.3308 & 0.1573 & 0.7038 & 0.4873 & 0.1908 & 0.6768 & 0.1411 & 0.7131 \\
RLVR & \textbf{0.7774} & \underline{0.0711} & \underline{0.7667} & \textbf{0.4482} & \underline{0.0608} & \underline{0.8220} & \underline{0.5863} & 0.1492 & 0.7504 & \underline{0.0937} & \underline{0.7797} \\
RD   & 0.6860 & 0.0781 & 0.6989 & 0.3390 & 0.1529 & 0.7084 & 0.4380 & \textbf{0.0976} & \textbf{0.7666} & 0.1095 & 0.7246 \\
RLCR & 0.6970 & 0.1242 & 0.6650 & 0.3450 & 0.0735 & 0.7208 & 0.5370 & \underline{0.1065} & 0.7438 & 0.1014 & 0.7099 \\
\rowcolor{gray!12} CSR  & \underline{0.7692} & \textbf{0.0587} & \textbf{0.7807} & \underline{0.4198} & \textbf{0.0564} & \textbf{0.8268} & \textbf{0.5923} & 0.1404 & \underline{0.7601} & \textbf{0.0852} & \textbf{0.7892} \\
\bottomrule
\end{tabular}
\end{table}

\textbf{Results.} Averaged over the three datasets, CSR has the lowest ECE, $0.0852$, and the highest AUROC, $0.7892$, and RLVR is second on both (Table~\ref{tab:csr_reasoning_tasks}). In distribution, CSR has the lowest ECE and the highest AUROC and raises accuracy over Base by $0.097$, but RLVR is slightly more accurate and CSR leads it by at most $0.014$ on the other two metrics. With the passages in context, Base already reaches an ECE of $0.0751$ on HotpotQA, against $0.3050$ closed-book, so little calibration error is left for the calibration term to remove. Base is less calibrated on 2WikiMultiHopQA, with an ECE of $0.1908$, and there CSR lowers ECE from $0.1492$ for RLVR to $0.1404$ and raises AUROC from $0.7504$ to $0.7601$. RD has the best ECE and AUROC on this dataset, while CSR has the highest accuracy. On MuSiQue, CSR has the lowest ECE and the highest AUROC.

\subsection{Reliability diagrams}
\label{app:reliability}
\begin{figure}[!t]
\centering
\begin{subfigure}{\linewidth}
\centering
\includegraphics[width=0.19\linewidth]{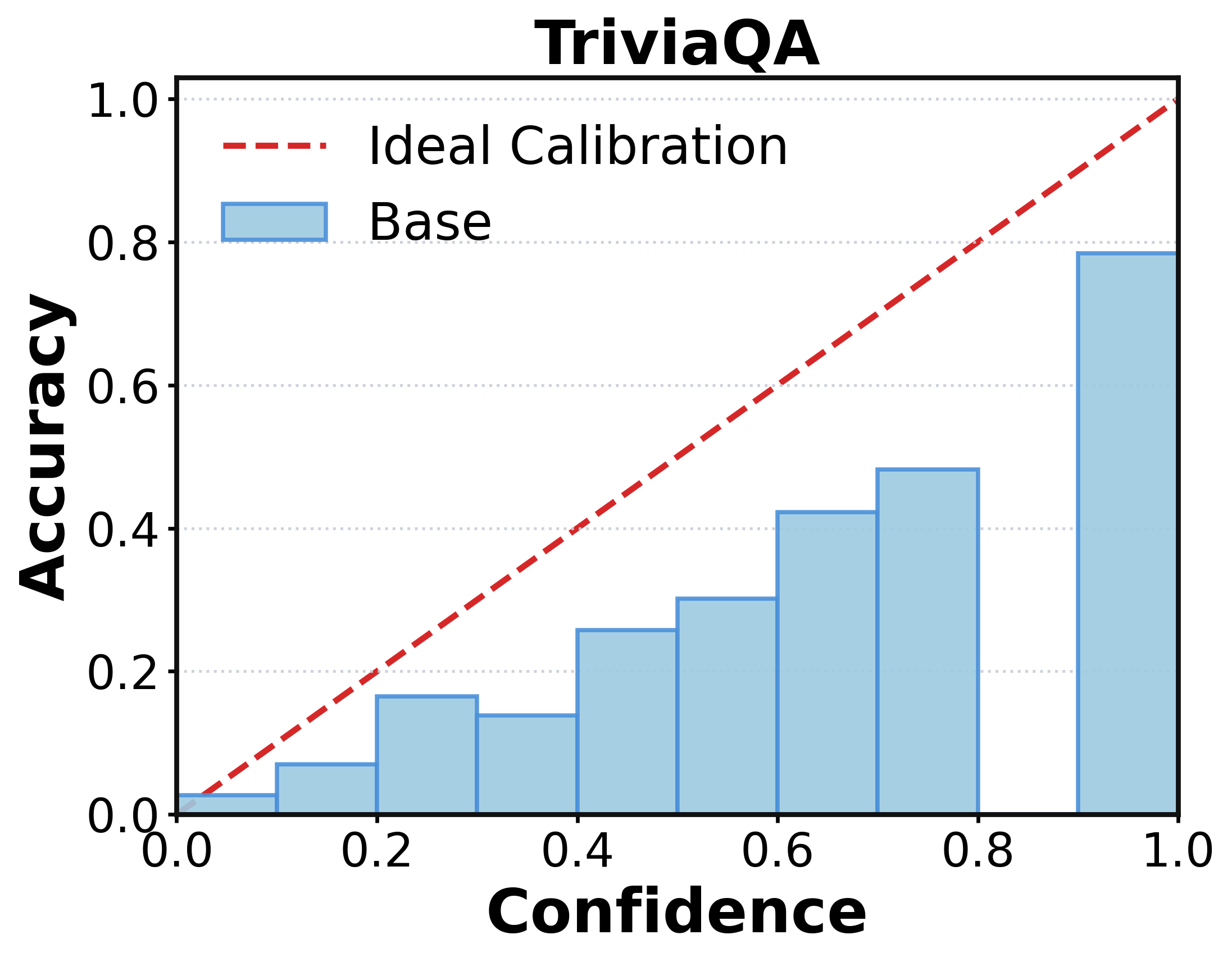}\hfill
\includegraphics[width=0.19\linewidth]{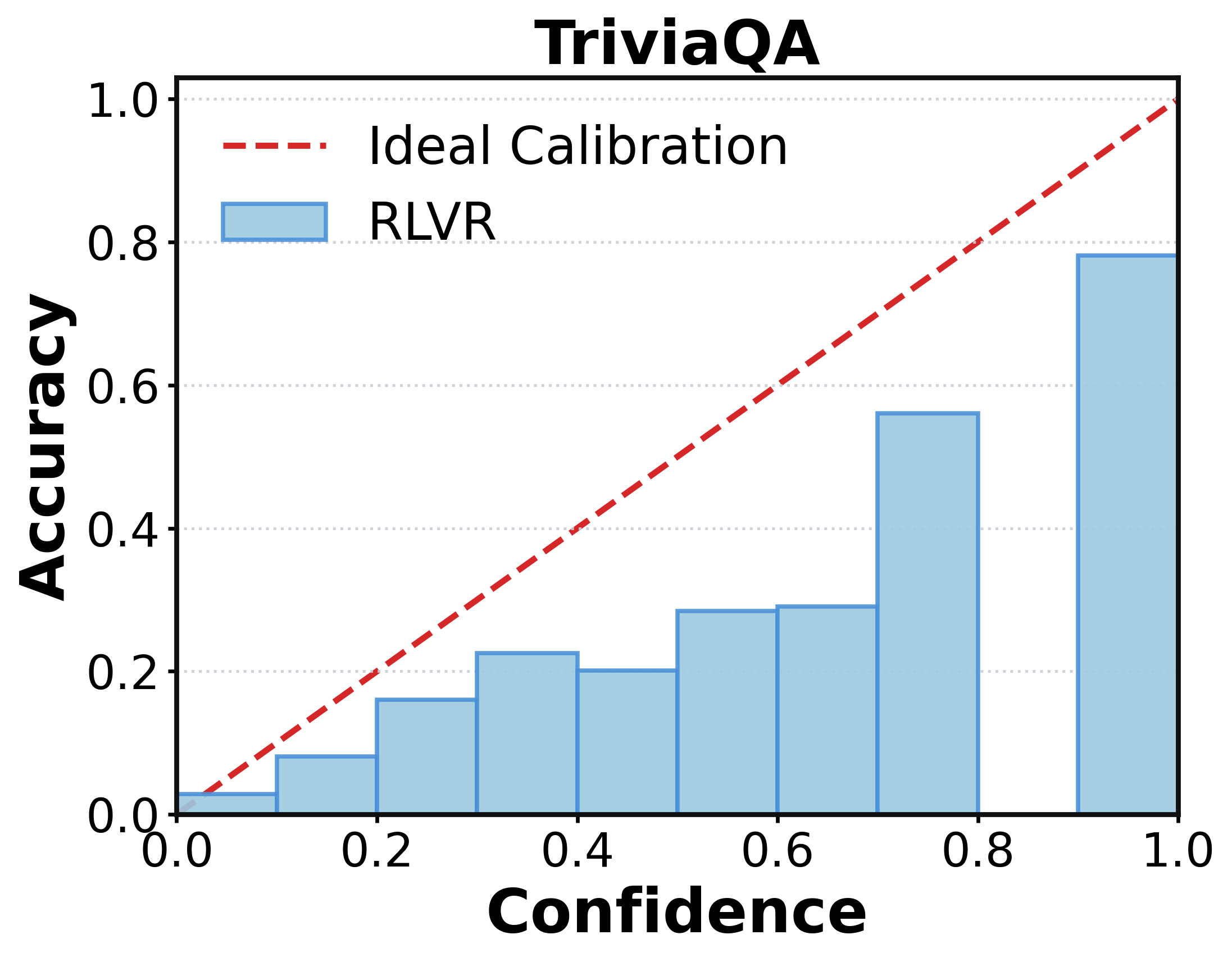}\hfill
\includegraphics[width=0.19\linewidth]{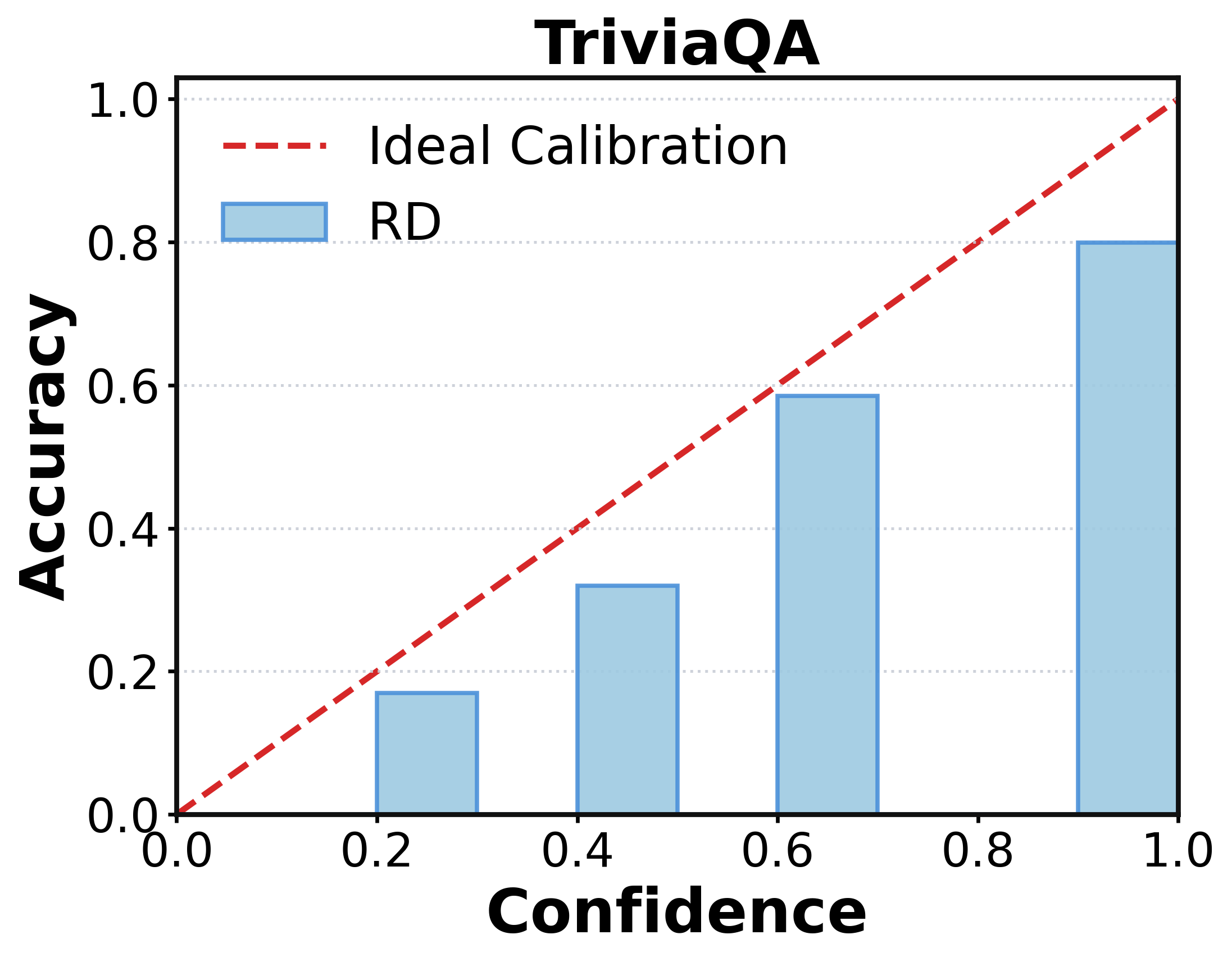}\hfill
\includegraphics[width=0.19\linewidth]{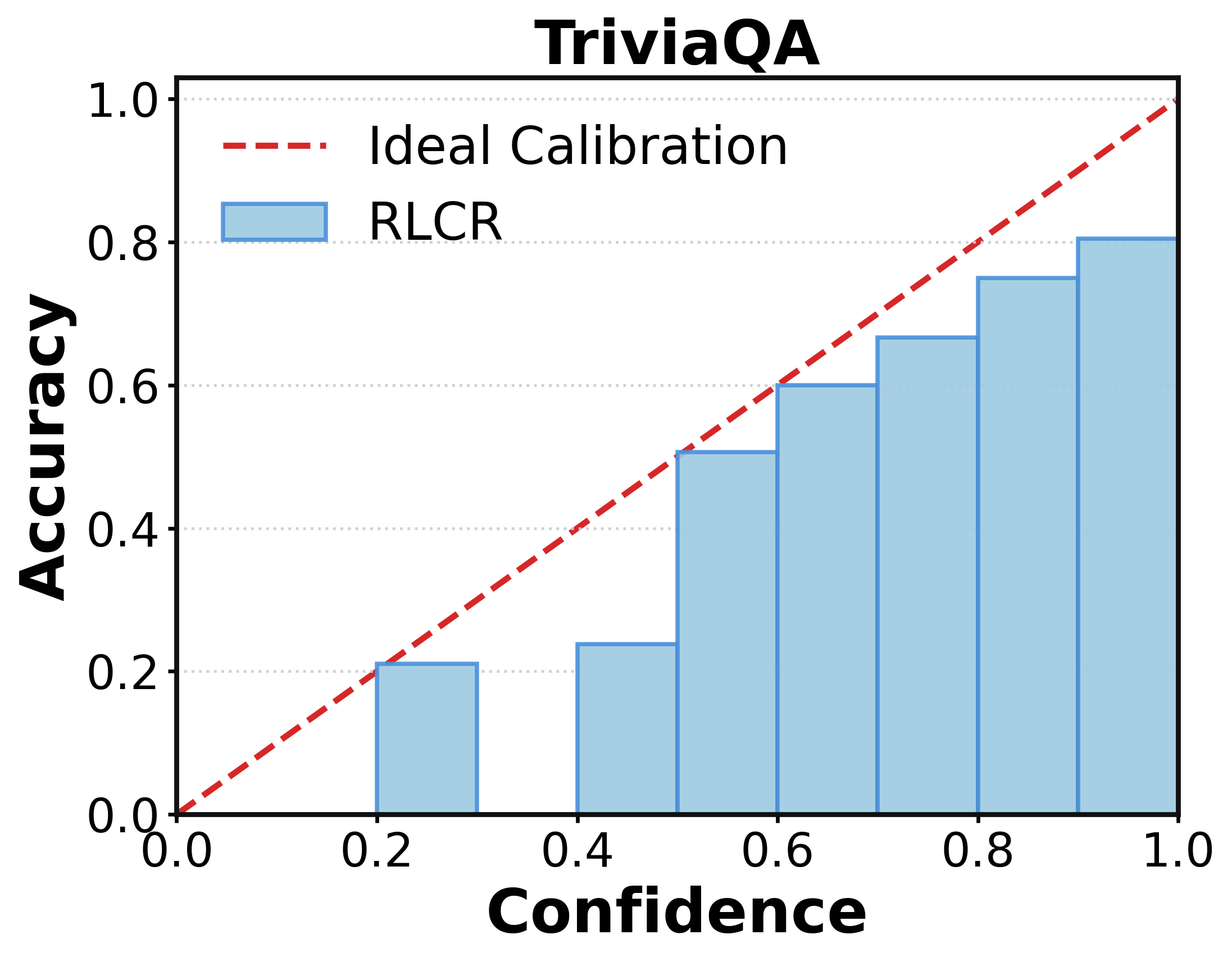}\hfill
\includegraphics[width=0.19\linewidth]{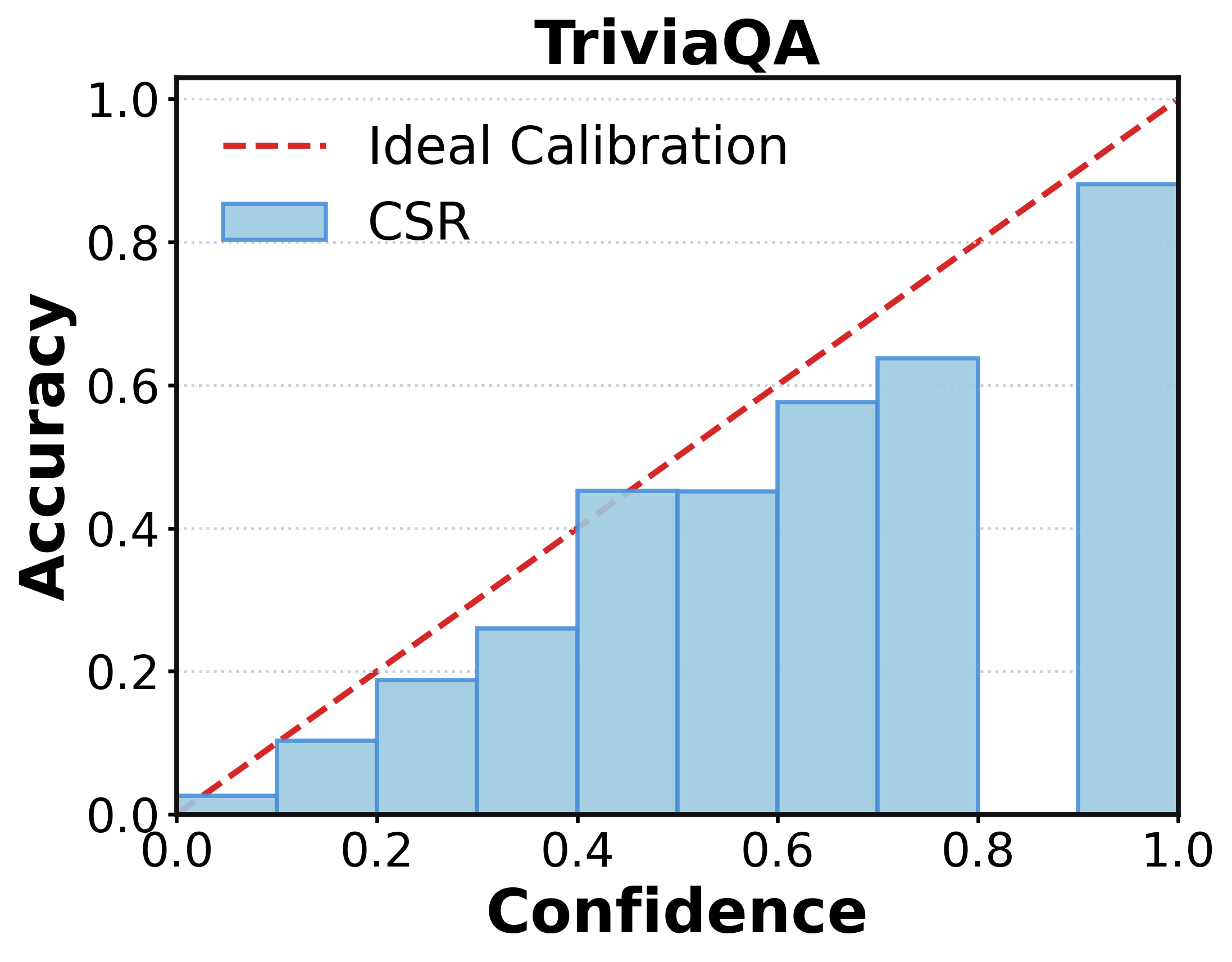}
\end{subfigure}
\vspace{2pt}

\begin{subfigure}{\linewidth}
\centering
\includegraphics[width=0.19\linewidth]{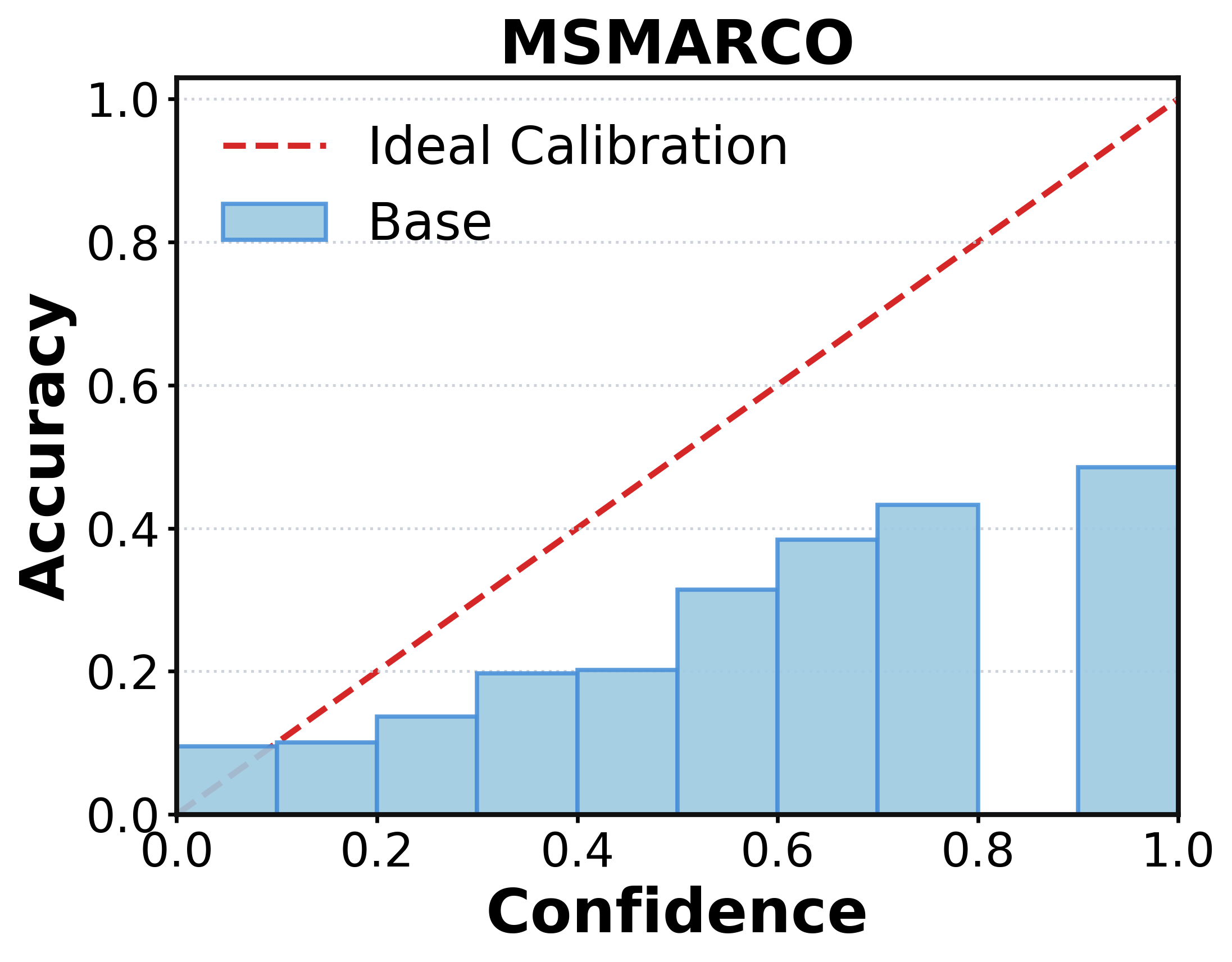}\hfill
\includegraphics[width=0.19\linewidth]{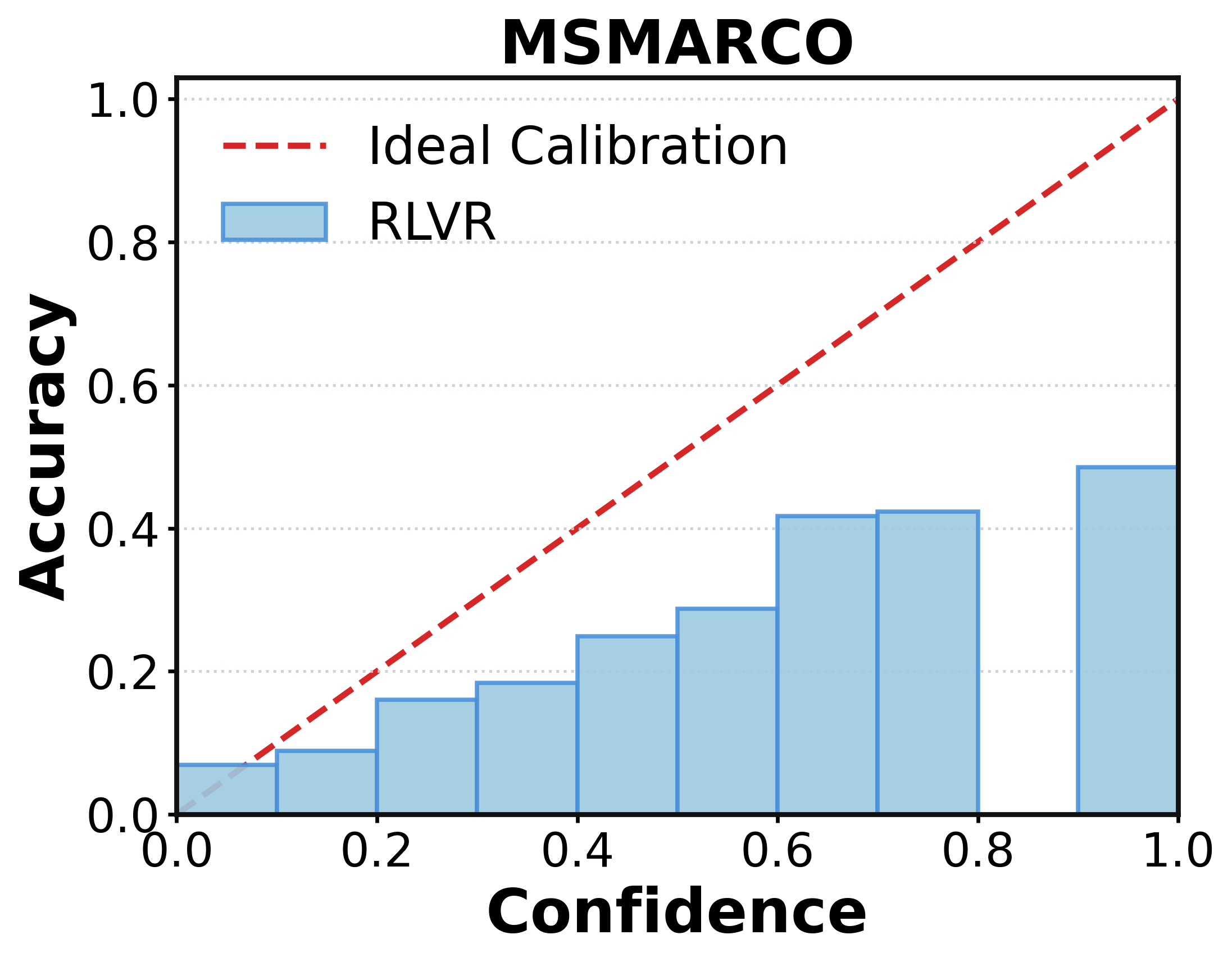}\hfill
\includegraphics[width=0.19\linewidth]{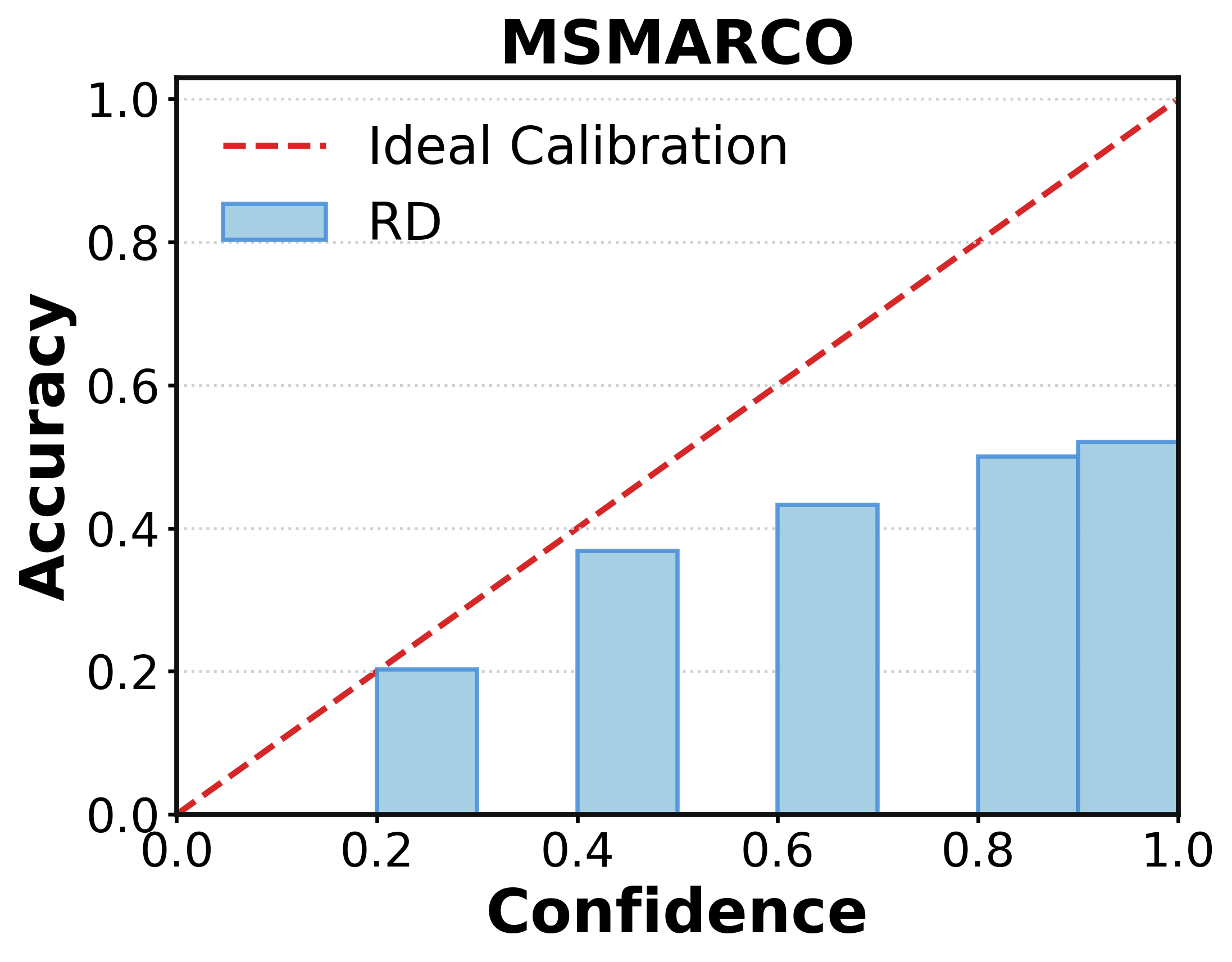}\hfill
\includegraphics[width=0.19\linewidth]{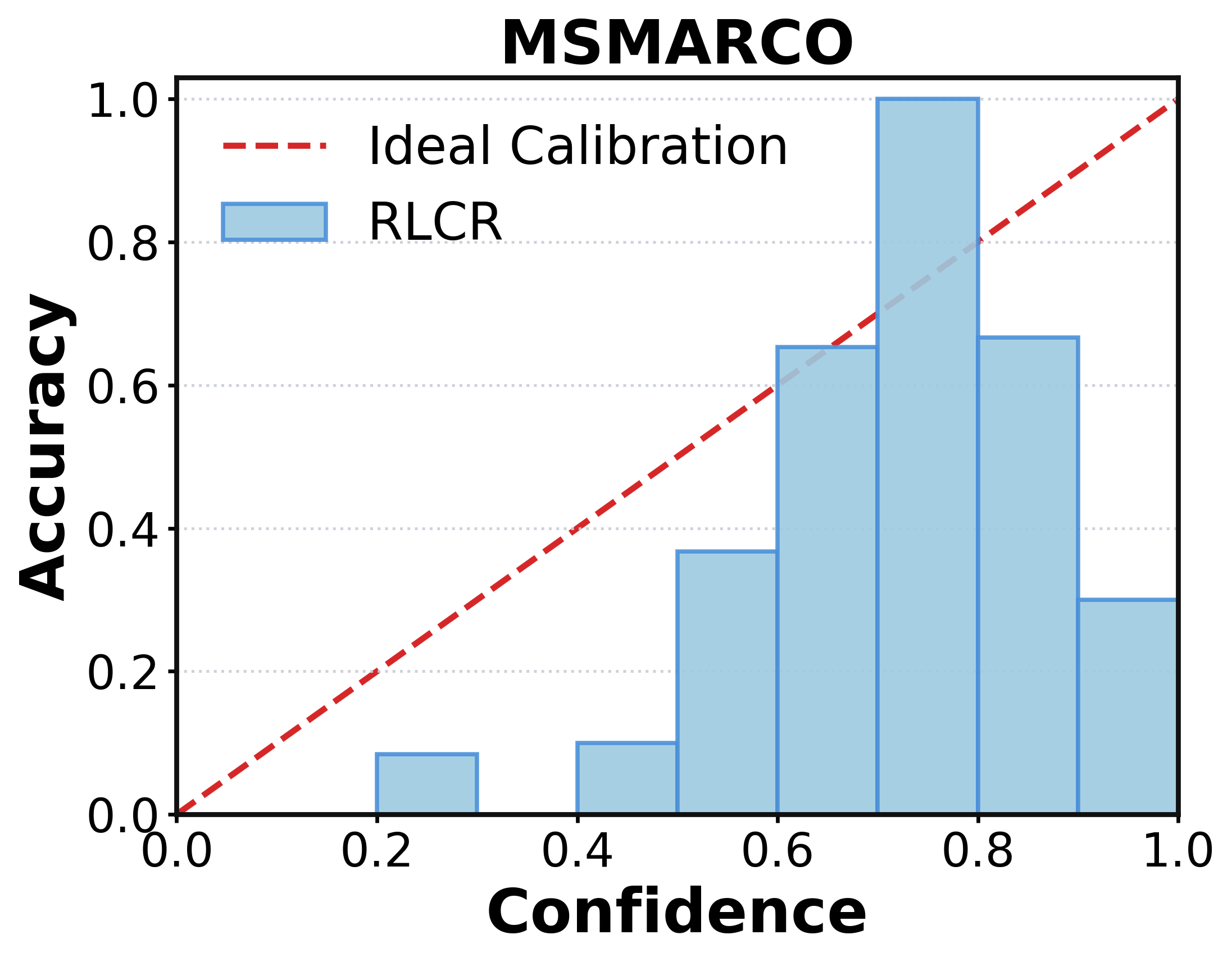}\hfill
\includegraphics[width=0.19\linewidth]{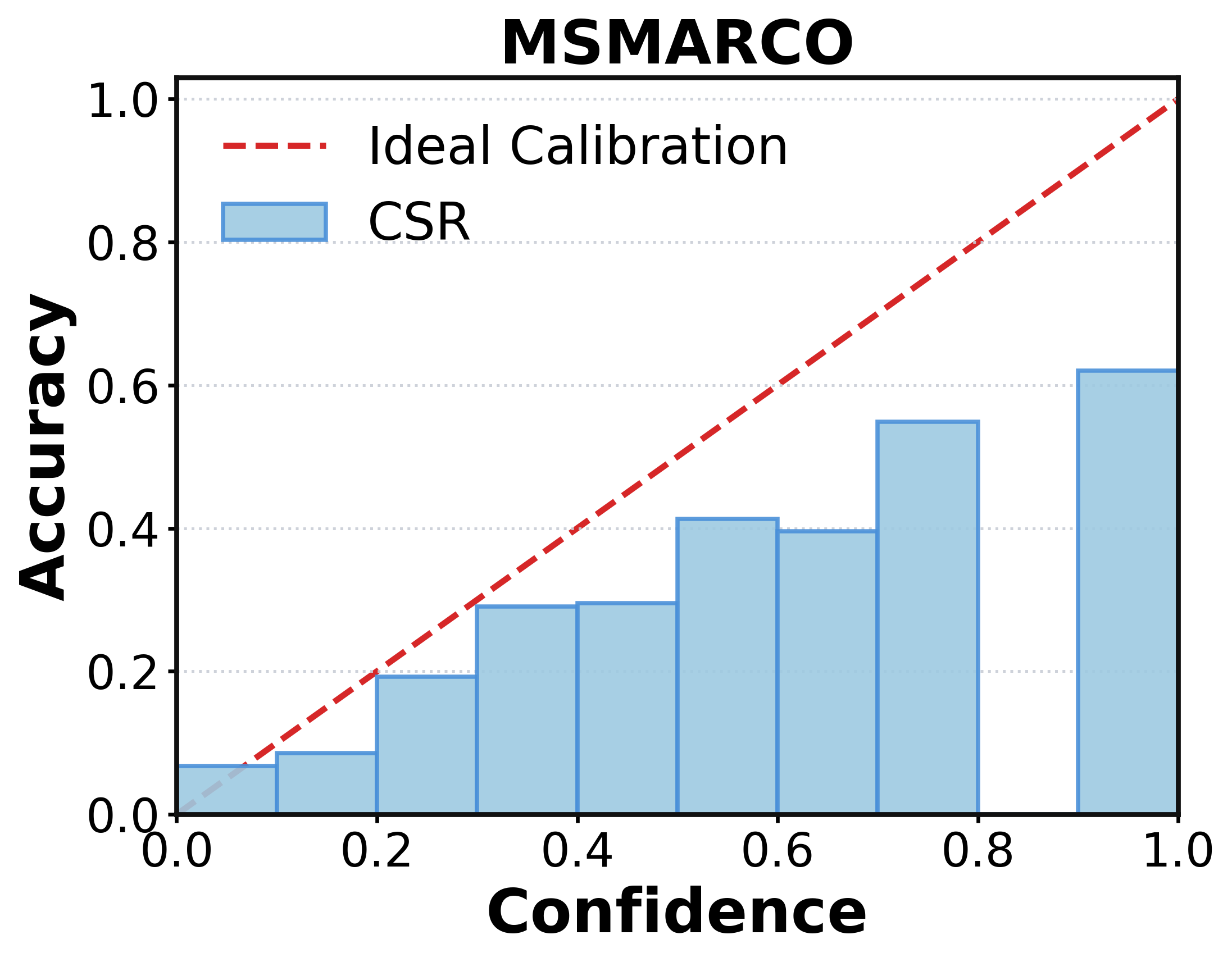}
\end{subfigure}
\caption{Reliability diagrams on Qwen for TriviaQA (top) and MSMARCO (bottom), both out of distribution. Columns are Base, RLVR, RD, RLCR, and CSR\@. Each bar is the accuracy of the questions whose confidence falls in a bin of width $0.1$.}
\label{fig:reliability}
\end{figure}

Fig.~\ref{fig:reliability} shows reliability diagrams on TriviaQA and MSMARCO, the two OOD datasets not shown in Fig.~\ref{fig:reliability_main}. On TriviaQA, CSR's bars stay close to the diagonal across the confidence range, consistent with the lowest ECE of the five methods, $0.0855$, while Base and RLVR fall well below the diagonal from the middle bins upward. On MSMARCO, every method is overconfident in its highest-confidence bins, but CSR's accuracy still rises steadily with confidence, which matches its highest AUROC on this dataset, $0.8001$.

\subsection{Transfer to larger backbones}
\label{app:scale_14b}

\textbf{Setup.} The backbones in Table~\ref{tab:main_results} are all in the 7--8B range, so we repeat the protocol on Qwen3-14B and Qwen3-32B~\citep{yang2025qwen3}, roughly two and four times that scale. Thinking mode is disabled in training and evaluation for every method except RLCR, whose output format places its reasoning in \texttt{\textless think\textgreater} tags and so requires it. All methods follow the training and evaluation settings of Appendix~\ref{app:exp_setup}, except that on Qwen3-14B CSR's calibration weight ranges from $\lambda_{\min}=0.5$ to $\lambda_{\max}=0.7$ instead of from $0.8$ to $1.2$, because with the original range the calibration reward lowered ECE further but also lowered accuracy. Qwen3-32B uses the original range.

\begin{table}[!t]
\centering
\small
\setlength{\tabcolsep}{3.0pt}
\renewcommand{\arraystretch}{1.1}
\caption{Transfer to Qwen3-14B and Qwen3-32B across HotpotQA and three out-of-distribution datasets. \textbf{Bold} marks the best and \underline{underline} the second-best value in each column within each model block.}
\label{tab:csr_qwen3_14b}
\resizebox{\textwidth}{!}{%
\begin{tabular}{l|l c c c|c c c|c c c|c c c|c c c}
\toprule
\multicolumn{1}{l}{} & \multicolumn{1}{l}{} & \multicolumn{3}{c}{\textbf{In-distribution}} & \multicolumn{3}{c}{} & \multicolumn{3}{c}{\textbf{Out-of-distribution}} & \multicolumn{3}{c}{} & \multicolumn{3}{c}{}\\
\cmidrule(lr){3-5}\cmidrule(lr){6-14}
\multicolumn{1}{l}{} & \multicolumn{1}{l}{} & \multicolumn{3}{c}{\textbf{HotpotQA}} & \multicolumn{3}{c}{\textbf{TriviaQA}} & \multicolumn{3}{c}{\textbf{MSMARCO}} & \multicolumn{3}{c}{\textbf{NQ-Open}} & \multicolumn{3}{c}{\textbf{Average}}\\
\cmidrule(lr){3-5}\cmidrule(lr){6-8}\cmidrule(lr){9-11}\cmidrule(lr){12-14}\cmidrule(lr){15-17}
\textbf{Model} & \textbf{Method} & Acc$\uparrow$ & ECE$\downarrow$ & AUROC$\uparrow$ & Acc$\uparrow$ & ECE$\downarrow$ & AUROC$\uparrow$ & Acc$\uparrow$ & ECE$\downarrow$ & AUROC$\uparrow$ & Acc$\uparrow$ & ECE$\downarrow$ & AUROC$\uparrow$ & Acc$\uparrow$ & ECE$\downarrow$ & AUROC$\uparrow$\\
\midrule
\multirow{5}{*}{\shortstack{Qwen3\\14B}}
& Base  & 0.2404 & 0.3528 & 0.8019 & \underline{0.6043} & 0.1820 & 0.8017 & \underline{0.3808} & \underline{0.3000} & 0.6859 & 0.3666 & \underline{0.2585} & \underline{0.7467} & 0.3980 & 0.2733 & 0.7590 \\
& RLVR  & \underline{0.2897} & 0.3220 & \underline{0.8362} & 0.6035 & 0.1918 & 0.7832 & 0.3783 & 0.3107 & 0.6933 & 0.3632 & 0.2817 & 0.7374 & \underline{0.4087} & 0.2766 & \underline{0.7625} \\
& RD    & \textbf{0.2950} & \underline{0.2080} & 0.7506 & 0.5910 & \underline{0.1300} & 0.8081 & 0.3420 & 0.3134 & \underline{0.7304} & 0.3180 & 0.3469 & 0.6886 & 0.3865 & \underline{0.2496} & 0.7444 \\
& RLCR  & 0.2640 & 0.4384 & 0.7349 & \textbf{0.6690} & 0.1750 & \underline{0.8260} & \textbf{0.4090} & 0.3726 & 0.7284 & \textbf{0.4230} & 0.3899 & 0.6951 & \textbf{0.4413} & 0.3440 & 0.7461 \\
\rowcolor{gray!12} \cellcolor{white}& CSR   & 0.2717 & \textbf{0.1430} & \textbf{0.8999} & 0.6006 & \textbf{0.0869} & \textbf{0.8718} & 0.3790 & \textbf{0.2296} & \textbf{0.7355} & \underline{0.3691} & \textbf{0.1359} & \textbf{0.8004} & 0.4051 & \textbf{0.1488} & \textbf{0.8269} \\
\midrule
\multirow{5}{*}{\shortstack{Qwen3\\32B}}
& Base  & 0.2386 & 0.2697 & 0.8201 & 0.6451 & 0.1061 & \underline{0.8571} & 0.3885 & 0.2386 & \underline{0.7525} & 0.3894 & 0.1688 & 0.7791 & 0.4154 & 0.1958 & 0.8022 \\
& RLVR  & \textbf{0.3115} & 0.2255 & \underline{0.8488} & 0.6546 & \underline{0.1025} & 0.8513 & 0.3822 & 0.2470 & 0.7430 & \underline{0.3946} & 0.1721 & \underline{0.7824} & \underline{0.4357} & 0.1868 & \underline{0.8064} \\
& RD    & \underline{0.3000} & \underline{0.1451} & 0.7712 & \underline{0.6560} & 0.2488 & 0.7612 & \underline{0.3970} & \textbf{0.1029} & 0.7010 & 0.3800 & \textbf{0.0880} & 0.6680 & 0.4333 & \underline{0.1462} & 0.7254 \\
& RLCR  & 0.2750 & 0.5315 & 0.7349 & \textbf{0.7080} & 0.1957 & 0.7612 & \textbf{0.4150} & 0.4520 & 0.6991 & \textbf{0.4400} & 0.4485 & 0.6692 & \textbf{0.4595} & 0.4069 & 0.7161 \\
\rowcolor{gray!12} \cellcolor{white}& CSR   & 0.2980 & \textbf{0.1300} & \textbf{0.9023} & 0.6514 & \textbf{0.0642} & \textbf{0.8748} & 0.3793 & \underline{0.2071} & \textbf{0.7638} & 0.3938 & \underline{0.1187} & \textbf{0.7843} & 0.4306 & \textbf{0.1300} & \textbf{0.8313} \\
\bottomrule
\end{tabular}%
}
\end{table}

\textbf{Results.} The main trends of Table~\ref{tab:main_results} carry over to both larger backbones (Table~\ref{tab:csr_qwen3_14b}). On Qwen3-14B, CSR attains the lowest ECE and the highest AUROC on every dataset, and it raises average AUROC from $0.7590$ for Base to $0.8269$. On Qwen3-32B, CSR has the highest AUROC on every dataset, and it raises average AUROC from $0.8022$ for Base to $0.8313$. On HotpotQA, it lowers ECE from $0.2697$ for Base and $0.2255$ for RLVR to $0.1300$. CSR also has the lowest average ECE, although RD is lower on MSMARCO and NQ-Open. On both backbones, RLCR has the highest average accuracy, measured on a single greedy answer, but more than twice CSR's average ECE, and RLVR again stays close to Base in average ECE and AUROC.

\subsection{Sample efficiency under matched training budgets}
\label{app:sample_efficiency}
\textbf{Setup.} To separate CSR's effect from the size of its training budget, we evaluate every method on HotpotQA with Qwen2.5-7B-Instruct at matched budgets of 1k to 10k examples. The intermediate budgets are checkpoints of each method's 10k run, so the $\lambda(t)$ and learning-rate schedules are those of the full run, and all other settings follow Appendix~\ref{app:exp_setup}. The CSR curve comes from the same run as the sigmoid arm in Fig.~\ref{fig:ablation_row}(b), and the 10k points of the baselines are the Qwen HotpotQA cells of Table~\ref{tab:main_results}.

\begin{figure}[!t]
\centering
\includegraphics[width=0.32\textwidth]{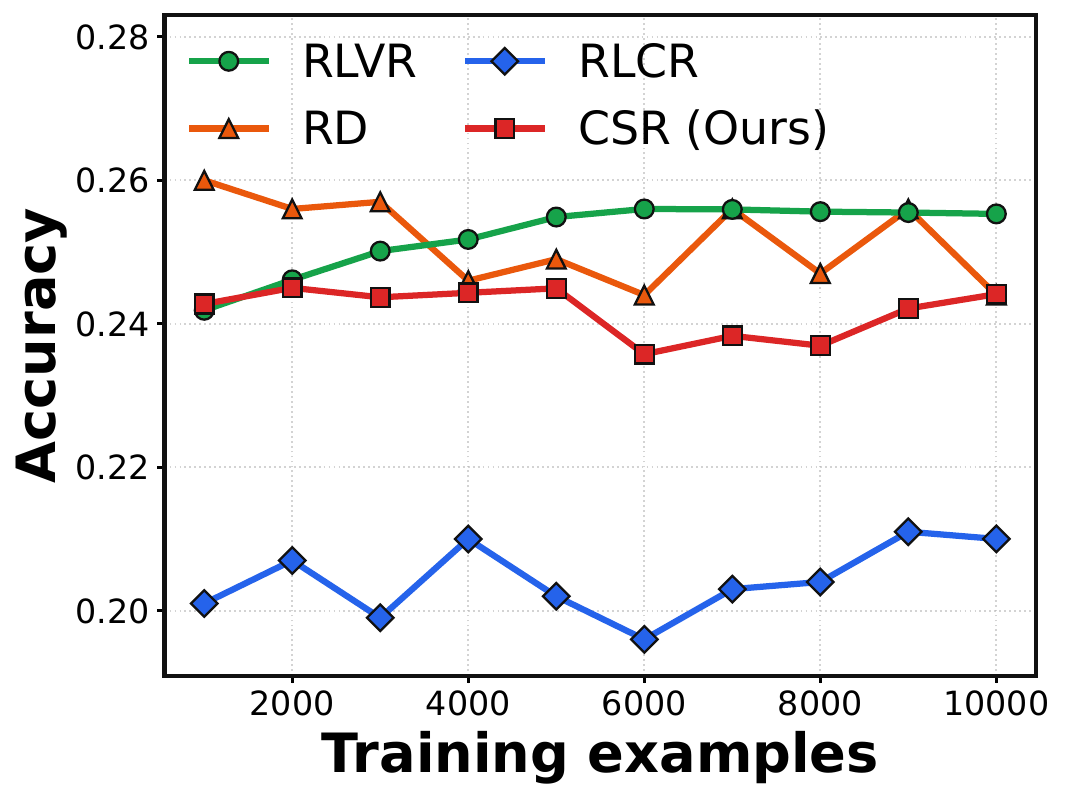}\hfill
\includegraphics[width=0.32\textwidth]{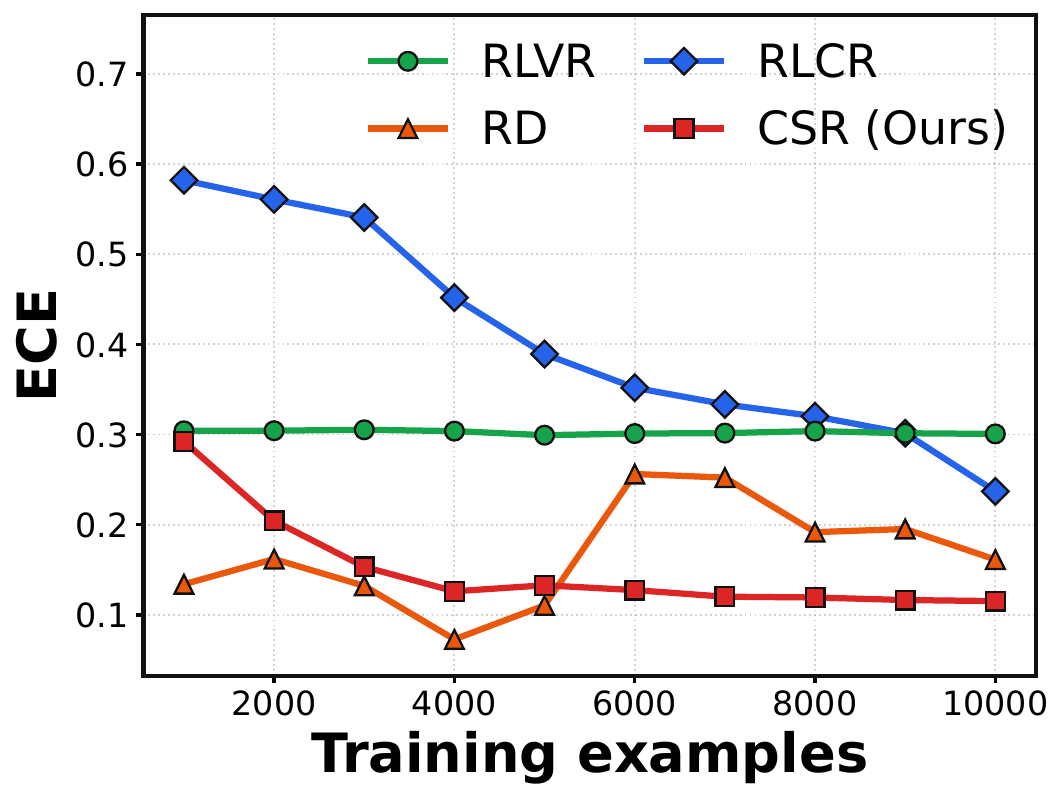}\hfill
\includegraphics[width=0.32\textwidth]{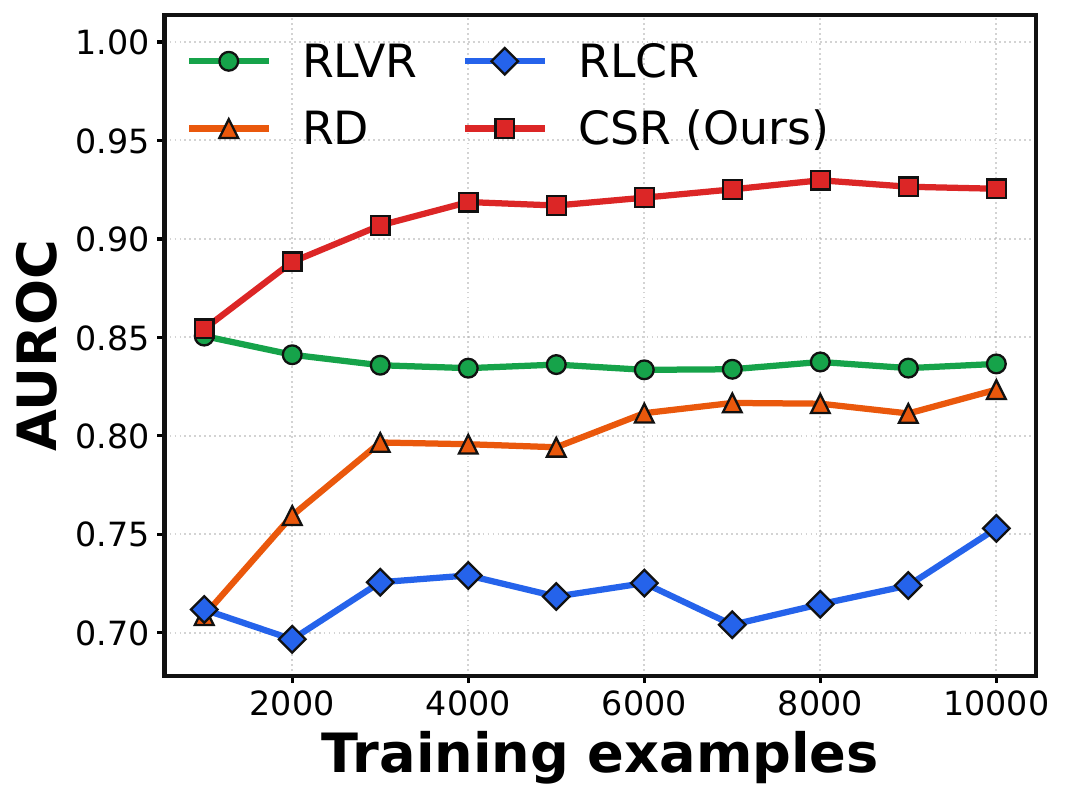}
\caption{Sample efficiency on HotpotQA with Qwen2.5-7B-Instruct at matched training budgets of 1k to 10k examples. Left to right: accuracy, ECE, and AUROC.}
\label{fig:budget_row}
\end{figure}

\textbf{Results.} CSR attains the highest AUROC at every budget and the lowest ECE from 6k on (Fig.~\ref{fig:budget_row}). Its gain arrives early: its ECE falls from about $0.29$ at 1k to about $0.13$ at 4k, after which its curves flatten. RLVR's accuracy instead rises slightly with more data, from about $0.24$ to $0.26$, while its ECE stays near $0.30$. CSR's accuracy stays near $0.24$, so the extra data mainly improves calibration for CSR and, modestly, accuracy for RLVR\@. RD has the lowest ECE up to 5k but ends above CSR, while its AUROC rises and its accuracy falls. RLCR lowers ECE, but its AUROC stays below CSR's and RLVR's.

\textbf{Practical guidance on training data.} Within the 10k run, most of CSR's gain is reached by about 4k examples: from 4k to 10k, its ECE falls by only about $0.01$ and its AUROC rises by less than $0.01$. These are checkpoints of a single run, so the curve suggests that a few thousand questions (roughly 4k to 6k) suffice on a task of similar difficulty. Which questions are used also matters. The correctness term gives a group-relative signal only when a rollout group mixes correct and incorrect answers, and only then can the calibration reward separate agreement among correct answers from agreement among incorrect ones. We therefore recommend corpora on which the backbone is right on some rollouts and wrong on others, with reliably verifiable answers. All our experiments use short-answer QA. We expect similar gains in other domains where rollout groups mix correct and incorrect answers that can be reliably verified, and smaller gains where they rarely do, while long-form generation and code would additionally need an equivalence judge that compares longer outputs.

\subsection{Token cost across methods}
\label{app:token_efficiency}
\begin{table}[!t]
\centering
\small
\setlength{\tabcolsep}{3.0pt}
\renewcommand{\arraystretch}{1.1}
\caption{LLM-judge and F1 equivalence on all three backbones. The Llama rows repeat Table~\ref{tab:efficient_sc_llama}. \textbf{Bold} marks the best and \underline{underline} the second-best value in each column within each model block.}
\label{tab:efficient_sc}
\resizebox{\textwidth}{!}{%
\ADLactivate
\begin{tabular}{l|l cc|cc|cc|cc|cc}
\toprule
\multicolumn{1}{l}{} & & \multicolumn{2}{c}{\textbf{In-distribution}} & \multicolumn{2}{c}{} & \multicolumn{2}{c}{\textbf{Out-of-distribution}} & \multicolumn{2}{c}{} & \multicolumn{2}{c}{}\\
\cmidrule(lr){3-4}\cmidrule(lr){5-10}
\multicolumn{1}{l}{} & & \multicolumn{2}{c}{\textbf{HotpotQA}} & \multicolumn{2}{c}{\textbf{TriviaQA}} & \multicolumn{2}{c}{\textbf{MSMARCO}} & \multicolumn{2}{c}{\textbf{NQ-Open}} & \multicolumn{2}{c}{\textbf{Average}}\\
\cmidrule(lr){3-4}\cmidrule(lr){5-6}\cmidrule(lr){7-8}\cmidrule(lr){9-10}\cmidrule(lr){11-12}
\textbf{Model} & \textbf{Method} & ECE$\downarrow$ & AUROC$\uparrow$ & ECE$\downarrow$ & AUROC$\uparrow$ & ECE$\downarrow$ & AUROC$\uparrow$ & ECE$\downarrow$ & AUROC$\uparrow$ & ECE$\downarrow$ & AUROC$\uparrow$\\
\midrule
& RD        & 0.1764 & 0.8316 & \underline{0.0746} & 0.7996 & 0.3302 & 0.7298 & 0.2081 & 0.7464 & 0.1973 & 0.7769 \\
& RLCR      & 0.1757 & 0.6923 & 0.0943 & 0.7337 & 0.2340 & 0.6816 & 0.1497 & 0.7013 & 0.1634 & 0.7022 \\
\noalign{\vskip 0.2ex}\cdashline{2-12}\noalign{\vskip 0.2ex}
& Base-LLM  & 0.2949 & 0.7050 & 0.0772 & 0.8423 & 0.1996 & 0.7062 & 0.1186 & 0.7361 & 0.1726 & 0.7474 \\
& RLVR-LLM  & 0.2567 & 0.8089 & 0.0901 & 0.8268 & 0.2280 & \underline{0.7371} & 0.1686 & 0.6957 & 0.1859 & 0.7671 \\
\rowcolor{gray!12} \cellcolor{white}& CSR-LLM   & \textbf{0.0978} & \textbf{0.9133} & \textbf{0.0670} & \textbf{0.8976} & \underline{0.1571} & \textbf{0.7594} & \textbf{0.1100} & \underline{0.7670} & \textbf{0.1080} & \textbf{0.8343} \\
\noalign{\vskip 0.2ex}\cdashline{2-12}\noalign{\vskip 0.2ex}
& Base-F1   & 0.2713 & 0.6681 & 0.1212 & 0.7929 & 0.1612 & 0.6094 & 0.1223 & 0.6992 & 0.1690 & 0.6924 \\
& RLVR-F1   & 0.2727 & 0.7770 & 0.0899 & 0.7819 & 0.2296 & 0.6670 & 0.1798 & 0.6771 & 0.1930 & 0.7258 \\
\rowcolor{gray!12} \cellcolor{white}\multirow{-8}{*}{Llama}& CSR-F1    & \underline{0.1240} & \underline{0.9008} & 0.1005 & \underline{0.8498} & \textbf{0.1042} & 0.7223 & \underline{0.1109} & \textbf{0.7735} & \underline{0.1099} & \underline{0.8116} \\
\midrule
& RD        & 0.1613 & 0.8235 & 0.0876 & 0.8013 & \textbf{0.1008} & 0.6531 & 0.1380 & 0.7606 & 0.1219 & 0.7596 \\
& RLCR      & 0.2370 & 0.7530 & 0.0894 & 0.7293 & 0.2031 & 0.7026 & 0.2671 & 0.6255 & 0.1991 & 0.7026 \\
\noalign{\vskip 0.2ex}\cdashline{2-12}\noalign{\vskip 0.2ex}
& Base-LLM  & 0.3050 & 0.8413 & 0.1881 & 0.8281 & 0.2945 & 0.7037 & 0.2487 & 0.7773 & 0.2591 & 0.7876 \\
& RLVR-LLM  & 0.3007 & 0.8365 & 0.1933 & 0.8241 & 0.2871 & \underline{0.7094} & 0.2492 & 0.7594 & 0.2576 & 0.7824 \\
\rowcolor{gray!12} \cellcolor{white}& CSR-LLM   & \underline{0.1152} & \textbf{0.9260} & \underline{0.0855} & \textbf{0.9094} & \underline{0.1560} & \textbf{0.8001} & \underline{0.0866} & \textbf{0.8348} & \underline{0.1108} & \textbf{0.8676} \\
\noalign{\vskip 0.2ex}\cdashline{2-12}\noalign{\vskip 0.2ex}
& Base-F1   & 0.2650 & 0.8206 & 0.1500 & 0.8166 & 0.1862 & 0.6079 & 0.1870 & 0.7489 & 0.1970 & 0.7485 \\
& RLVR-F1   & 0.2684 & 0.8250 & 0.1485 & 0.8157 & 0.1868 & 0.6126 & 0.1756 & 0.7364 & 0.1948 & 0.7474 \\
\rowcolor{gray!12} \cellcolor{white}\multirow{-8}{*}{Qwen}& CSR-F1    & \textbf{0.1132} & \underline{0.9079} & \textbf{0.0531} & \underline{0.9019} & 0.1761 & 0.6509 & \textbf{0.0659} & \underline{0.8031} & \textbf{0.1021} & \underline{0.8159} \\
\midrule
& RD        & \underline{0.0994} & \textbf{0.9065} & 0.1061 & 0.8118 & 0.1879 & 0.7014 & 0.1395 & \textbf{0.7897} & 0.1332 & \underline{0.8024} \\
& RLCR      & 0.2117 & 0.6715 & \textbf{0.0611} & 0.6913 & 0.2669 & 0.6529 & 0.1942 & 0.6856 & 0.1835 & 0.6753 \\
\noalign{\vskip 0.2ex}\cdashline{2-12}\noalign{\vskip 0.2ex}
& Base-LLM  & 0.2563 & 0.7745 & 0.0952 & 0.7860 & 0.3011 & \underline{0.7087} & 0.1507 & 0.7035 & 0.2008 & 0.7432 \\
& RLVR-LLM  & 0.2533 & 0.7936 & 0.1126 & 0.7887 & 0.2939 & 0.7003 & 0.1635 & 0.7079 & 0.2058 & 0.7476 \\
\rowcolor{gray!12} \cellcolor{white}& CSR-LLM   & \textbf{0.0966} & \underline{0.8710} & \underline{0.0751} & \textbf{0.8530} & 0.2513 & \textbf{0.7351} & \textbf{0.0974} & \underline{0.7606} & \textbf{0.1301} & \textbf{0.8049} \\
\noalign{\vskip 0.2ex}\cdashline{2-12}\noalign{\vskip 0.2ex}
& Base-F1   & 0.3901 & 0.6533 & 0.1692 & 0.6571 & 0.1946 & 0.6000 & 0.1722 & 0.6677 & 0.2315 & 0.6445 \\
& RLVR-F1   & 0.3477 & 0.6991 & 0.1211 & 0.7179 & \underline{0.1829} & 0.5914 & 0.1900 & 0.6558 & 0.2104 & 0.6660 \\
\rowcolor{gray!12} \cellcolor{white}\multirow{-8}{*}{Mistral}& CSR-F1    & 0.1098 & 0.8523 & 0.1522 & \underline{0.8127} & \textbf{0.1407} & 0.6041 & \underline{0.1235} & 0.7041 & \underline{0.1316} & 0.7433 \\
\bottomrule
\end{tabular}%
}
\end{table}

Table~\ref{tab:main_results} reports the average total token cost per question alongside the calibration metrics. RD has the lowest cost on Llama and Mistral because it decodes a single short answer, whereas Base, RLVR, and CSR decode $K=16$. RLCR costs more than CSR on every backbone, because its prompt is long, its output format has many required fields, and it generates a reasoning trace before answering. The gap is largest on Qwen, where RLCR uses $641.07$ tokens per question and CSR uses $198.54$, so RLCR costs about $3.2\times$ as much. On Llama and Mistral the gap is smaller, with RLCR costing about $15\%$ to $18\%$ more than CSR\@.

Each CSR rollout is an answer-only completion comparable in length to a Base or RLVR generation, so CSR's total per-question cost scales with $K$. Tok counts policy tokens only. The $\binom{K}{2}$ equivalence judgments that Base, RLVR, and CSR use at inference are accounted for in Appendix~\ref{app:efficient_sc}.

\subsection{Token-level F1 as a surrogate for the semantic-equivalence judge}
\label{app:efficient_sc}
\textbf{Judging cost.} With $K=8$ rollouts, each training question requires 8 correctness judgments against the gold answer and $\binom{K}{2}=28$ pairwise judgments, giving at most 36 judge calls per question and at most $360{,}000$ calls across a full 10k-example epoch. Evaluation uses $K=16$, so $\binom{16}{2}+16=136$ judgments per question and $136{,}000$ calls per dataset at $n=1{,}000$. In a timed training run on Llama-3.1-8B-Instruct, judging accounted for about one third ($33.2\%$) of the wall-clock. At deployment, the $K$ correctness judgments against gold answers are unnecessary, but semantic entropy with the LLM judge still uses $\binom{K}{2}$ equivalence judgments to cluster the sampled answers. For short-answer tasks, the F1 surrogate below removes those remaining LLM calls by replacing the semantic-equivalence test used to form the clusters.

\textbf{F1 equivalence.} We approximate semantic equivalence by lexical overlap. For two sampled answers $Y^{(i)}$ and $Y^{(j)}$, we define an approximate equivalence label by thresholding their token-level F1:
\begin{equation}
\tilde s_{ji}=\mathbf{1}\!\left[\mathrm{F1}\!\left(Y^{(j)},Y^{(i)}\right)\ge \tau\right].
\end{equation}
We replace the original semantic-equivalence signal in the semantic calibration reward with $\tilde s_{ji}$ (in place of $s_{ji}$ in Eq.~\eqref{eq:rlsr}) and use the same rule to form semantic clusters at evaluation. This keeps the reward structure while removing pairwise LLM-judge calls from both training and deployment.

\textbf{Scope.} Token-level F1 is a low-cost surrogate \emph{for short, entity-style answers}, where semantically equivalent answers usually share most of their tokens, and it is not intended to replace the judge for long or compositional answers. Its characteristic failure is under-crediting paraphrase and alias equivalence. This matches the AUROC gap of Sec.~\ref{sec:f1_vs_llm_judge}.

\textbf{Training and evaluation details.} All other settings follow the main experiments, with the threshold of Sec.~\ref{sec:f1_vs_llm_judge} fixed across datasets. Base-F1 and RLVR-F1 use F1 only at evaluation because their training objectives contain no semantic-equivalence term.

\textbf{Results.} The F1-equivalence variant preserves the main qualitative trend of CSR (Table~\ref{tab:efficient_sc}). Against Base-F1 and RLVR-F1, CSR-F1 has higher AUROC in all 12 model-dataset comparisons and lower ECE in 10 of 12, the exceptions being TriviaQA for Llama and Mistral against RLVR-F1. Relative to CSR-LLM, AUROC is lower in 11 of 12 cells, all but Llama on NQ-Open.

\subsection{Parsing and abstention rates}
\label{app:parsing_rate}
\begin{table}[!t]
\centering
\small
\setlength{\tabcolsep}{2.5pt}
\renewcommand{\arraystretch}{1.05}
\caption{Format-valid rate (Valid) and abstention rate (Abstain) on the runs of Table~\ref{tab:main_results}. Valid is the fraction of answers with a parsed, in-range confidence for RD and RLCR, and the answer-extraction rate for Base, RLVR, and CSR, which emit no confidence. Abstain is the fraction of responses that are abstentions, over $16{,}000$ rollouts per cell for Base, RLVR, and CSR and $1{,}000$ greedy answers for RD and RLCR. \textbf{Bold} marks the best value in each column within each model block.}
\label{tab:parsing_rate_verbal_methods}
\begin{tabular}{l|l c c|c c|c c|c c}
\toprule
\multicolumn{1}{l}{} & \multicolumn{1}{l}{} & \multicolumn{2}{c}{\textbf{HotpotQA}} & \multicolumn{2}{c}{\textbf{TriviaQA}} & \multicolumn{2}{c}{\textbf{MSMARCO}} & \multicolumn{2}{c}{\textbf{NQ-Open}}\\
\cmidrule(lr){3-4}\cmidrule(lr){5-6}\cmidrule(lr){7-8}\cmidrule(lr){9-10}
\textbf{Model} & \textbf{Method} & Valid$\uparrow$ & Abstain$\downarrow$ & Valid$\uparrow$ & Abstain$\downarrow$ & Valid$\uparrow$ & Abstain$\downarrow$ & Valid$\uparrow$ & Abstain$\downarrow$\\
\midrule
\multirow{5}{*}{Llama}
& Base  & \textbf{1.000} & 0.126 & \textbf{1.000} & 0.011 & \textbf{1.000} & 0.004 & \textbf{1.000} & 0.017\\
& RLVR  & \textbf{1.000} & 0.053 & \textbf{1.000} & 0.004 & \textbf{1.000} & 0.002 & \textbf{1.000} & 0.010\\
& RD    & 0.999 & 0.069 & \textbf{1.000} & 0.008 & 0.998 & 0.004 & 0.994 & 0.005\\
& RLCR  & 0.969 & 0.043 & 0.941 & 0.007 & 0.982 & 0.008 & 0.952 & 0.038\\
\rowcolor{gray!12} \cellcolor{white}& CSR   & \textbf{1.000} & \textbf{0.001} & \textbf{1.000} & \textbf{0.001} & \textbf{1.000} & \textbf{0.000} & \textbf{1.000} & \textbf{0.001}\\
\midrule
\multirow{5}{*}{Qwen}
& Base  & \textbf{1.000} & 0.003 & \textbf{1.000} & 0.001 & \textbf{1.000} & \textbf{0.000} & \textbf{1.000} & 0.001\\
& RLVR  & \textbf{1.000} & 0.003 & \textbf{1.000} & 0.001 & \textbf{1.000} & \textbf{0.000} & \textbf{1.000} & 0.001\\
& RD    & \textbf{1.000} & 0.093 & \textbf{1.000} & 0.010 & 0.999 & 0.013 & 0.996 & 0.037\\
& RLCR  & 0.995 & 0.128 & 0.996 & 0.008 & \textbf{1.000} & 0.024 & 0.999 & 0.029\\
\rowcolor{gray!12} \cellcolor{white}& CSR   & \textbf{1.000} & \textbf{0.001} & \textbf{1.000} & \textbf{0.000} & \textbf{1.000} & \textbf{0.000} & \textbf{1.000} & \textbf{0.000}\\
\midrule
\multirow{5}{*}{Mistral}
& Base  & \textbf{1.000} & 0.002 & \textbf{1.000} & \textbf{0.000} & \textbf{1.000} & 0.001 & \textbf{1.000} & 0.004\\
& RLVR  & \textbf{1.000} & 0.001 & \textbf{1.000} & \textbf{0.000} & \textbf{1.000} & \textbf{0.000} & \textbf{1.000} & 0.003\\
& RD    & \textbf{1.000} & 0.345 & \textbf{1.000} & 0.099 & 0.997 & 0.121 & 0.998 & 0.276\\
& RLCR  & 0.999 & 0.042 & 0.999 & \textbf{0.000} & \textbf{1.000} & 0.005 & \textbf{1.000} & 0.008\\
\rowcolor{gray!12} \cellcolor{white}& CSR   & \textbf{1.000} & \textbf{0.000} & \textbf{1.000} & \textbf{0.000} & \textbf{1.000} & \textbf{0.000} & \textbf{1.000} & \textbf{0.000}\\
\bottomrule
\end{tabular}
\end{table}

The verbalized baselines, RD and RLCR, require the model to emit a numeric confidence in a fixed format alongside its answer (a \texttt{Confidence:} field on a $0$--$10$ scale, rescaled to $[0,1]$, for RD and \texttt{\textless confidence\textgreater} tags for RLCR)\@. Format failures occur when a generation exhausts its token budget before emitting the required fields or emits no recoverable confidence (Table~\ref{tab:parsing_rate_verbal_methods}).

RD and RLCR emit a parsable confidence on at least $94\%$ of questions in every cell. Unparsable confidences are scored $c(x)=0$ and affect up to $5.9\%$ of RLCR's questions (Llama on TriviaQA, Valid $0.941$) and at most $0.6\%$ of RD's, so they are rare but can shift RLCR's metrics in the affected cells of Table~\ref{tab:main_results}. Abstention accounts for part of RD's lower accuracy on Mistral. We count a response as an abstention when its extracted answer, lowercased and stripped of punctuation, is an abstention phrase such as ``unknown'' or ``I don't know'', or begins with a refusal such as ``I'm sorry''. A bare ``none'' is not counted, since it can be the correct answer. Abstentions are scored as incorrect. Mistral RD abstains on $34.5\%$ of HotpotQA and $27.6\%$ of NQ-Open questions, whereas CSR abstains on at most $0.1\%$ of responses in any cell and Base on at most $1.7\%$, except on HotpotQA with Llama, where $12.6\%$ of its rollouts are refusals. RLCR abstains on $12.8\%$ of HotpotQA questions with Qwen and on at most $4.3\%$ elsewhere.

Mistral RD's abstention is consistent with the form of its reward. The logarithmic scoring rule scores the stated confidence against the outcome, so an abstention stated with a confidence of $0$ earns the same score as the best-scored incorrect answer. Llama and Mistral both learn to state $0$, on $32.6\%$ and $47.2\%$ of HotpotQA questions, but Llama abstains on only $6.9\%$, so it still writes an answer in most of these cases. CSR gives abstention no such advantage, because an abstention earns nothing from its correctness term.

\raggedbottom
\tcbset{height fixed for=first and middle}
\clubpenalties 2 10000 0
\widowpenalties 2 10000 0
\renewcommand{\cmark}{\unskip~\ding{51}}\renewcommand{\xmark}{\unskip~\ding{55}}
\section{Qualitative rollout examples}
\label{app:visualization}

We provide qualitative rollout-level samples from CSR, Base, RLVR, RD, and RLCR on Llama-3.1-8B-Instruct, generated with the prompts in Appendix~\ref{app:system_prompt}. Within each box, every numbered line $[k]$ shows one sampled rollout, annotated with \cmark{} or \xmark{} according to the LLM judge against the gold answer(s). For Base, RLVR, and CSR, whose semantic confidence comes from $K=16$ rollouts as in Table~\ref{tab:main_results}, we display all sixteen rollouts. For RD and RLCR we display the single greedy output with its verbalized confidence. Correct answers should receive high confidence (Appendix~\ref{app:vis_consistent}) and wrong answers low confidence (Appendix~\ref{app:vis_dispersed}). Bold text only highlights the answers.

\subsection{Correct and consistent rollout groups}
\label{app:vis_consistent}
When CSR answers a question correctly, its rollouts concentrate in one semantic cluster and the semantic confidence is high, reaching $1.000$ when every rollout falls in that cluster. The two examples below show that one cluster is not the same as one string: the rollouts differ in surface form and the semantic-equivalence judge merges them by meaning, so the confidence reflects agreement in meaning rather than a repeated output.

\subsubsection{HotpotQA (in-distribution)}
\label{app:vis_consistent_hotpotqa}
\begin{tcolorbox}[breakable,
  title=Question,
  colback=gray!10,
  colframe=black,
  colbacktitle=black,
  coltitle=white,
  fonttitle=\bfseries,
  boxrule=1pt,
  arc=2mm,
  left=8pt, right=8pt, top=4pt, bottom=4pt,
  before skip=8pt, after skip=8pt,
]
Charles Haughey held what position when he dissolved the list of members who were elected to the lower house of the Oireachtas of Ireland on 25 May 1989?\\
\textbf{Ground Truth Answer:} Taoiseach of Ireland
\end{tcolorbox}

The question asks for Haughey's office. CSR answers it in fourteen distinct strings, from the bare ``Taoiseach (Prime Minister)'' to a full restatement of the question (``Charles Haughey held the position of Taoiseach (Prime Minister) when he dissolved \ldots''). The semantic-equivalence judge places all sixteen rollouts in one cluster, so its accuracy and confidence are both $1.000$. Base and RLVR reach the same values with seven distinct strings each, and six of RLVR's sixteen rollouts are the same sentence. The verbalized baselines are also correct, with RD and RLCR both answering ``Taoiseach''.

\begin{tcolorbox}[breakable,
  title={CSR (Ours) generated answer},
  colframe=CSRRed,
  colbacktitle=CSRRed,
  colback=CSRRed!8,
  coltitle=white,
  fonttitle=\bfseries,
  boxrule=1pt,
  arc=2mm,
  left=8pt, right=8pt, top=4pt, bottom=4pt,
  before skip=8pt, after skip=3pt,
]
\ttfamily\raggedright\fontsize{10pt}{10.72pt}\selectfont
{[0]} \boldanswer{Taoiseach} (\boldanswer{Prime Minister}) \cmark\par
{[1]} He was the \boldanswer{Taoiseach} (\boldanswer{Prime Minister} of Ireland) \cmark\par
{[2]} He was the \boldanswer{Taoiseach} (\boldanswer{Prime Minister}). \cmark\par
{[3]} \boldanswer{Prime Minister} (\boldanswer{Taoiseach}) \cmark\par
{[4]} Charles Haughey was the \boldanswer{Taoiseach} (\boldanswer{Prime Minister}) of Ireland. \cmark\par
{[5]} The \boldanswer{Taoiseach} (\boldanswer{Prime Minister}). \cmark\par
{[6]} Charles Haughey was the \boldanswer{Taoiseach} (\boldanswer{Prime Minister}) when he dissolved the list of members who were elected to the lower house of the Oireachtas of Ireland on 25 May 1989. \cmark\par
{[7]} Charles Haughey was the \boldanswer{Taoiseach} (\boldanswer{Prime Minister}) of Ireland when he dissolved the list of members who were elected to the lower house of the Oireachtas of Ireland on 25 May 1989. \cmark\par
{[8]} He was the \boldanswer{Taoiseach} of Ireland \cmark\par
{[9]} He was the \boldanswer{Taoiseach} (\boldanswer{Prime Minister}) of Ireland. \cmark\par
{[10]} Charles Haughey held the position of \boldanswer{Taoiseach} (\boldanswer{Prime Minister}) when he dissolved the list of members who were elected to the lower house of the Oireachtas of Ireland on 25 May 1989. \cmark\par
{[11]} Charles Haughey was the \boldanswer{Taoiseach} (\boldanswer{Prime Minister}) when he dissolved the list of members who were elected to the lower house of the Oireachtas of Ireland on 25 May 1989. \cmark\par
{[12]} \boldanswer{Taoiseach} (\boldanswer{Prime Minister} of Ireland) \cmark\par
{[13]} \boldanswer{Taoiseach} (\boldanswer{Prime Minister}). \cmark\par
{[14]} He held the position of \boldanswer{Taoiseach} (\boldanswer{Prime Minister}). \cmark\par
{[15]} \boldanswer{Taoiseach} (\boldanswer{Prime Minister} of Ireland) \cmark
\metrics{1.000}{1.000}
\end{tcolorbox}

\begin{tcolorbox}[breakable,
  title={Base generated answer},
  colframe=BaseBlue,
  colbacktitle=BaseBlue,
  colback=BaseBlue!8,
  coltitle=white,
  fonttitle=\bfseries,
  boxrule=1pt,
  arc=2mm,
  left=8pt, right=8pt, top=4pt, bottom=4pt,
  before skip=3pt, after skip=3pt,
]
\ttfamily\raggedright\fontsize{10pt}{10.72pt}\selectfont
{[0]} Charles Haughey was the \boldanswer{Taoiseach} (\boldanswer{Prime Minister}) of Ireland when he dissolved the list of members who were elected to the lower house of the Oireachtas of Ireland on 25 May 1989. \cmark\par
{[1]} Charles Haughey was the \boldanswer{Taoiseach} (\boldanswer{Prime Minister}) of Ireland when he dissolved the list of members who were elected to the lower house of the Oireachtas of Ireland on 25 May 1989. \cmark\par
{[2]} Charles Haughey was the \boldanswer{Taoiseach} (\boldanswer{Prime Minister}) when he dissolved the list of members who were elected to the lower house of the Oireachtas of Ireland on 25 May 1989. \cmark\par
{[3]} Charles Haughey was the \boldanswer{Taoiseach} (\boldanswer{Prime Minister}) of Ireland. \cmark\par
{[4]} Charles Haughey was the \boldanswer{Taoiseach} (\boldanswer{Prime Minister}) of Ireland when he dissolved the list of members who were elected to the lower house of the Oireachtas of Ireland on 25 May 1989. \cmark\par
{[5]} Charles Haughey was the \boldanswer{Taoiseach} (\boldanswer{Prime Minister}) of Ireland. \cmark\par
{[6]} Charles Haughey was the \boldanswer{Taoiseach} (\boldanswer{Prime Minister}) when he dissolved the list of members who were elected to the lower house of the Oireachtas of Ireland on 25 May 1989. \cmark\par
{[7]} Charles Haughey was the \boldanswer{Taoiseach} (\boldanswer{Prime Minister}) of Ireland when he dissolved the list of members who were elected to the lower house of the Oireachtas of Ireland on 25 May 1989. \cmark\par
{[8]} Charles Haughey was the \boldanswer{Taoiseach} (\boldanswer{Prime Minister} of Ireland) when he dissolved the list of members who were elected to the lower house of the Oireachtas of Ireland on 25 May 1989. \cmark\par
{[9]} Charles Haughey was the \boldanswer{Taoiseach} (\boldanswer{Prime Minister}) when he dissolved the list of members who were elected to the lower house of the Oireachtas of Ireland on 25 May 1989. \cmark\par
{[10]} Charles Haughey was the \boldanswer{Taoiseach} (\boldanswer{Prime Minister}) when he dissolved the list of members who were elected to the lower house of the Oireachtas of Ireland on 25 May 1989. \cmark\par
{[11]} Charles Haughey was the \boldanswer{Taoiseach} (\boldanswer{Prime Minister}) when he dissolved the list of members who were elected to the lower house of the Oireachtas of Ireland on 25 May 1989. \cmark\par
{[12]} Charles Haughey was the \boldanswer{Taoiseach} (\boldanswer{Prime Minister} of Ireland) when he dissolved the list of members who were elected to the lower house of the Oireachtas of Ireland on 25 May 1989. \cmark\par
{[13]} Charles Haughey was the \boldanswer{Taoiseach}. \cmark\par
{[14]} Charles Haughey held the position of \boldanswer{Taoiseach} (\boldanswer{Prime Minister}) when he dissolved the list of members who were elected to the lower house of the Oireachtas of Ireland on 25 May 1989. \cmark\par
{[15]} Charles Haughey was the \boldanswer{Taoiseach} of Ireland. \cmark
\metrics{1.000}{1.000}
\end{tcolorbox}

\begin{tcolorbox}[breakable,
  title={RLVR generated answer},
  colframe=RLVRGreen,
  colbacktitle=RLVRGreen,
  colback=RLVRGreen!8,
  coltitle=white,
  fonttitle=\bfseries,
  boxrule=1pt,
  arc=2mm,
  left=8pt, right=8pt, top=4pt, bottom=4pt,
  before skip=3pt, after skip=3pt,
]
\ttfamily\raggedright\fontsize{10pt}{10.72pt}\selectfont
{[0]} \boldanswer{Taoiseach}. \cmark\par
{[1]} Charles Haughey was the \boldanswer{Taoiseach} (\boldanswer{Prime Minister}) of Ireland. \cmark\par
{[2]} Charles Haughey was the \boldanswer{Taoiseach} (\boldanswer{Prime Minister}) of Ireland. \cmark\par
{[3]} Charles Haughey was the \boldanswer{Taoiseach} (\boldanswer{Prime Minister}) of Ireland. \cmark\par
{[4]} Charles Haughey was the \boldanswer{Taoiseach} (\boldanswer{Prime Minister}) of Ireland. \cmark\par
{[5]} The \boldanswer{Prime Minister} of Ireland. \cmark\par
{[6]} Charles Haughey was the \boldanswer{Taoiseach} (\boldanswer{Prime Minister}) when he dissolved the list of members who were elected to the lower house of the Oireachtas of Ireland on 25 May 1989. \cmark\par
{[7]} Charles Haughey was the \boldanswer{Taoiseach} (\boldanswer{Prime Minister}) of Ireland when he dissolved the list of members who were elected to the lower house of the Oireachtas of Ireland on 25 May 1989. \cmark\par
{[8]} Charles Haughey was the \boldanswer{Taoiseach} (\boldanswer{Prime Minister} of Ireland). \cmark\par
{[9]} Charles Haughey was the \boldanswer{Taoiseach} (\boldanswer{Prime Minister}) of Ireland. \cmark\par
{[10]} Charles Haughey was the \boldanswer{Taoiseach} (\boldanswer{Prime Minister}) of Ireland. \cmark\par
{[11]} Charles Haughey was the \boldanswer{Taoiseach} (\boldanswer{Prime Minister}) when he dissolved the list of members who were elected to the lower house of the Oireachtas of Ireland on 25 May 1989. \cmark\par
{[12]} Charles Haughey was the \boldanswer{Taoiseach}. \cmark\par
{[13]} Charles Haughey was the \boldanswer{Taoiseach}. \cmark\par
{[14]} Charles Haughey was the \boldanswer{Taoiseach}. \cmark\par
{[15]} Charles Haughey was the \boldanswer{Taoiseach} (\boldanswer{Prime Minister}) of Ireland when he dissolved the list of members who were elected to the lower house of the Oireachtas of Ireland on 25 May 1989. \cmark
\metrics{1.000}{1.000}
\end{tcolorbox}

\begin{tcolorbox}[breakable,
  title={RD generated answer},
  colframe=RDGold,
  colbacktitle=RDGold,
  colback=RDGold!8,
  coltitle=white,
  fonttitle=\bfseries,
  boxrule=1pt,
  arc=2mm,
  left=8pt, right=8pt, top=4pt, bottom=4pt,
  before skip=3pt, after skip=3pt,
]
\ttfamily\raggedright\fontsize{10pt}{10.72pt}\selectfont
{[0]} Answer: \boldanswer{Taoiseach}, Confidence: 8 \cmark
\metrics{1.000}{0.800}
\end{tcolorbox}

\begin{tcolorbox}[breakable,
  title={RLCR generated answer},
  colframe=RLCRPurple,
  colbacktitle=RLCRPurple,
  colback=RLCRPurple!8,
  coltitle=white,
  fonttitle=\bfseries,
  boxrule=1pt,
  arc=2mm,
  left=8pt, right=8pt, top=4pt, bottom=4pt,
  before skip=3pt, after skip=3pt,
]
\ttfamily\raggedright\fontsize{10pt}{10.72pt}\selectfont
{[0]} \textless{}think\textgreater{} To solve this problem, I'll first recall that Charles Haughey was the Taoiseach (Prime Minister) of Ireland from 1981 to 1982 and again from 1987 to 1992. I'll then consider the date given in the question, which is 25 May 1989. I'll think about the fact that the question is asking about the dissolution of the lower house of the Oireachtas, which is the D\'ail \'Eireann. I'll recall that the Taoiseach has the power to dissolve the D\'ail and call a general election. Given that Charles Haughey was the Taoiseach at the time, I'll reason that he was likely the one who dissolved the list of members who were elected to the lower house of the Oireachtas on 25 May 1989. \textless{}/think\textgreater{} \textless{}answer\textgreater{}\boldanswer{Taoiseach}\textless{}/answer\textgreater{} \textless{}analysis\textgreater{} There are a few uncertainties in this solution. Firstly, I'm assuming that Charles Haughey was still the Taoiseach in May 1989, but I'm not certain about his exact tenure as Taoiseach. Secondly, I'm assuming that the dissolution of the list of members was a direct action by the Taoiseach, but it's possible that there were other factors at play. Additionally, there could be alternative approaches to solving this problem, such as looking at the specific legislation or events surrounding the dissolution of the D\'ail in 1989. \textless{}/analysis\textgreater{} \textless{}confidence\textgreater{} \boldanswer{0.62} \textless{}/confidence\textgreater{} \cmark
\metrics{1.000}{0.620}
\end{tcolorbox}

\subsubsection{NQ-Open (out-of-distribution)}
\label{app:vis_consistent_nqopen}
\begin{tcolorbox}[breakable,
  title=Question,
  colback=gray!10,
  colframe=black,
  colbacktitle=black,
  coltitle=white,
  fonttitle=\bfseries,
  boxrule=1pt,
  arc=2mm,
  left=8pt, right=8pt, top=4pt, bottom=4pt,
  before skip=8pt, after skip=8pt,
]
how many seasons of the smurfs are there\\
\textbf{Ground Truth Answer:} 9
\end{tcolorbox}

The same behavior holds out of distribution, where the rollouts range from the bare number to a full sentence and still form a single cluster. Base answers correctly on fifteen of its sixteen rollouts, and the one that invents a second series and assigns nine seasons to each forms a second cluster and lowers its confidence to $0.792$, so the dispersion tracks the rollout it gets wrong.

\begin{tcolorbox}[breakable,
  title={CSR (Ours) generated answer},
  colframe=CSRRed,
  colbacktitle=CSRRed,
  colback=CSRRed!8,
  coltitle=white,
  fonttitle=\bfseries,
  boxrule=1pt,
  arc=2mm,
  left=8pt, right=8pt, top=4pt, bottom=4pt,
  before skip=8pt, after skip=3pt,
]
\ttfamily\raggedright\fontsize{10pt}{10.72pt}\selectfont
{[0]} There are \boldanswer{9 seasons} of the Smurfs animated television series, \cmark\par
{[1]} There are \boldanswer{9 seasons} of the Smurfs. \cmark\par
{[2]} There are \boldanswer{9 seasons} of the Smurfs (Smurfs), \cmark\par
{[3]} There are \boldanswer{9 seasons} of the Smurfs. \cmark\par
{[4]} There are \boldanswer{9 seasons} of the animated television series `The Smurfs', \cmark\par
{[5]} \boldanswer{9} \cmark\par
{[6]} \boldanswer{9} \cmark\par
{[7]} \boldanswer{9 seasons} \cmark\par
{[8]} There are \boldanswer{9 seasons} of the Smurfs. \cmark\par
{[9]} \boldanswer{9 seasons} \cmark\par
{[10]} There are \boldanswer{9 seasons} of the Smurfs animated television series. \cmark\par
{[11]} There are \boldanswer{9 seasons} of the Smurfs TV series. \cmark\par
{[12]} There are \boldanswer{9 seasons} of the original Smurfs cartoon series that aired from 1981 to 1989. \cmark\par
{[13]} There are \boldanswer{9 seasons} of The Smurfs \cmark\par
{[14]} \boldanswer{9} \cmark\par
{[15]} There are \boldanswer{9 seasons} of the Smurfs. \cmark
\metrics{1.000}{1.000}
\end{tcolorbox}

\begin{tcolorbox}[breakable,
  title={Base generated answer},
  colframe=BaseBlue,
  colbacktitle=BaseBlue,
  colback=BaseBlue!8,
  coltitle=white,
  fonttitle=\bfseries,
  boxrule=1pt,
  arc=2mm,
  left=8pt, right=8pt, top=4pt, bottom=4pt,
  before skip=3pt, after skip=3pt,
]
\ttfamily\raggedright\fontsize{10pt}{10.72pt}\selectfont
{[0]} There are \boldanswer{9 seasons} of the classic animated television series ``The Smurfs'' (also known as ``Smurf'' or ``Smurfs''), which aired from 1981 to 1990. \cmark\par
{[1]} There are \boldanswer{9 seasons} of the Smurfs. \cmark\par
{[2]} There are \boldanswer{9 seasons} of the classic animated series ``The Smurfs'' (also known as ``Smurf'' or ``Smurfs'') \cmark\par
{[3]} There are \boldanswer{9 seasons} of the Smurfs. \cmark\par
{[4]} There are \boldanswer{9 seasons} of the animated TV series ``The Smurfs''. \cmark\par
{[5]} There are \boldanswer{9 seasons} of the original Smurfs animated series, which aired from 1981 to 1990. \cmark\par
{[6]} There are \boldanswer{9 seasons} of the Smurfs. \cmark\par
{[7]} There are \boldanswer{9 seasons} of the animated TV series ``The Smurfs''. \cmark\par
{[8]} There are \boldanswer{9 seasons} of the Smurfs TV show. \cmark\par
{[9]} There are \boldanswer{9 seasons} of the original Smurfs animated series (1981-1990). \cmark\par
{[10]} There are \boldanswer{9 seasons} of the Smurfs animated television series that aired from 1981 to 1990. \cmark\par
{[11]} There are \boldanswer{9 seasons} of the animated television series ``The Smurfs'' (also known as ``The Smurfs'' or ``Smurfs''). \cmark\par
{[12]} There are \boldanswer{9 seasons} of the original Smurfs cartoon series that aired from 1981 to 1990. \cmark\par
{[13]} There are \boldanswer{9 seasons} of the Smurfs animated television series, which aired from 1981 to 1990. \cmark\par
{[14]} There are \boldanswer{9 seasons} of the original animated series ``The Smurfs'' (1981-1990), and a total of \boldanswer{9 seasons} of the new CGI-animated series ``The Smurfs'' (2011-2019). \xmark\par
{[15]} There are \boldanswer{9 seasons} of the original Smurfs cartoon series. \cmark
\metrics{0.938}{0.792}
\end{tcolorbox}

\begin{tcolorbox}[breakable,
  title={RLVR generated answer},
  colframe=RLVRGreen,
  colbacktitle=RLVRGreen,
  colback=RLVRGreen!8,
  coltitle=white,
  fonttitle=\bfseries,
  boxrule=1pt,
  arc=2mm,
  left=8pt, right=8pt, top=4pt, bottom=4pt,
  before skip=3pt, after skip=3pt,
]
\ttfamily\raggedright\fontsize{10pt}{10.72pt}\selectfont
{[0]} There are \boldanswer{9 seasons} of the Smurfs. \cmark\par
{[1]} There are \boldanswer{9 seasons} of the Smurfs. \cmark\par
{[2]} There are \boldanswer{9 seasons} of the Smurfs. \cmark\par
{[3]} There are \boldanswer{9 seasons} of the Smurfs. \cmark\par
{[4]} There are \boldanswer{9 seasons} of the Smurfs. \cmark\par
{[5]} There are \boldanswer{9 seasons} of the original Smurfs animated TV series. \cmark\par
{[6]} There are \boldanswer{9 seasons} of the Smurfs. \cmark\par
{[7]} There are \boldanswer{9 seasons} of the Smurfs. \cmark\par
{[8]} There are \boldanswer{9 seasons} of the Smurfs. \cmark\par
{[9]} There are \boldanswer{9 seasons} of the original Smurfs cartoon series. \cmark\par
{[10]} There are \boldanswer{9 seasons} of the Smurfs. \cmark\par
{[11]} There are \boldanswer{9 seasons} of the Smurfs. \cmark\par
{[12]} There are \boldanswer{9 seasons} of the Smurfs (1981-1990) \cmark\par
{[13]} There are \boldanswer{9 seasons} of the Smurfs. \cmark\par
{[14]} There are \boldanswer{9 seasons} of the Smurfs. \cmark\par
{[15]} There are \boldanswer{9 seasons} of the Smurfs. \cmark
\metrics{1.000}{1.000}
\end{tcolorbox}

\begin{tcolorbox}[breakable,
  title={RD generated answer},
  colframe=RDGold,
  colbacktitle=RDGold,
  colback=RDGold!8,
  coltitle=white,
  fonttitle=\bfseries,
  boxrule=1pt,
  arc=2mm,
  left=8pt, right=8pt, top=4pt, bottom=4pt,
  before skip=3pt, after skip=3pt,
]
\ttfamily\raggedright\fontsize{10pt}{10.72pt}\selectfont
{[0]} Answer: \boldanswer{9}, Confidence: 6 \cmark
\metrics{1.000}{0.600}
\end{tcolorbox}

\begin{tcolorbox}[breakable,
  title={RLCR generated answer},
  colframe=RLCRPurple,
  colbacktitle=RLCRPurple,
  colback=RLCRPurple!8,
  coltitle=white,
  fonttitle=\bfseries,
  boxrule=1pt,
  arc=2mm,
  left=8pt, right=8pt, top=4pt, bottom=4pt,
  before skip=3pt, after skip=3pt,
]
\ttfamily\raggedright\fontsize{10pt}{10.72pt}\selectfont
{[0]} \textless{}think\textgreater{} I'm thinking about the popular animated TV show ``The Smurfs'' and trying to recall the information about its seasons. I'm searching through my knowledge database to find the correct information. I'm considering the show's initial run from 1981 to 1990 and any potential revivals or spin-offs. I'm also thinking about the show's production history and the typical length of seasons for animated TV series. \textless{}/think\textgreater{} \textless{}answer\textgreater{}{\rmfamily\textbf{There are 9 seasons of the original ``The Smurfs'' TV series}}\textless{}/answer\textgreater{} \textless{}analysis\textgreater{} Uncertainties:\\ 1. I'm assuming the question is referring to the original ``The Smurfs'' TV series, but it's possible the user is asking about a specific spin-off or revival, which could have a different number of seasons.\\ 2. I'm relying on my knowledge of the show's initial run, but there might be some ambiguity about the exact dates of the show's production and broadcast.\\ 3. I'm considering the typical length of seasons for animated TV series, but there could be exceptions or variations in the case of ``The Smurfs''.\\ 4. I'm not accounting for any potential differences in season length or episode count between the original series and any spin-offs or revivals. \textless{}/analysis\textgreater{} \textless{}confidence\textgreater{} \boldanswer{0.62} \textless{}/confidence\textgreater{} \cmark
\metrics{1.000}{0.620}
\end{tcolorbox}

\subsection{Incorrect and dispersed rollout groups}
\label{app:vis_dispersed}
The complementary case is a question CSR cannot answer. The rollouts then spread over many semantic clusters and the semantic confidence falls, which signals the error. The failure this guards against is a confident hallucination: a policy that commits to one wrong answer reports high confidence, which the calibration reward penalizes by rewarding incorrect rollouts for disagreeing. In both examples below no method answers correctly, and the methods differ only in how clearly their confidence says so.

\subsubsection{HotpotQA (in-distribution)}
\label{app:vis_dispersed_hotpotqa}
\begin{tcolorbox}[breakable,
  title=Question,
  colback=gray!10,
  colframe=black,
  colbacktitle=black,
  coltitle=white,
  fonttitle=\bfseries,
  boxrule=1pt,
  arc=2mm,
  left=8pt, right=8pt, top=4pt, bottom=4pt,
  before skip=8pt, after skip=8pt,
]
What memorial park located at John J. Pershing Drive in North Omaha has, among other amenities, a baseball park?\\
\textbf{Ground Truth Answer:} Dodge Memorial Park
\end{tcolorbox}

CSR spreads its sixteen rollouts over eleven semantic clusters and reports $0.105$, whereas RLVR collapses the same guesses onto four clusters and reports $0.313$.

\begin{tcolorbox}[breakable,
  title={CSR (Ours) generated answer},
  colframe=CSRRed,
  colbacktitle=CSRRed,
  colback=CSRRed!8,
  coltitle=white,
  fonttitle=\bfseries,
  boxrule=1pt,
  arc=2mm,
  left=8pt, right=8pt, top=4pt, bottom=4pt,
  before skip=8pt, after skip=3pt,
]
\ttfamily\raggedright\fontsize{10pt}{10.72pt}\selectfont
{[0]} \boldanswer{Fontenberry Park} \xmark\par
{[1]} \boldanswer{Fontenberry Memorial Park} \xmark\par
{[2]} \boldanswer{Lauritzen Gardens} \xmark\par
{[3]} {\rmfamily\textbf{The memorial park at John J. Pershing Drive in North Omaha is Fontenelle Park.}} \xmark\par
{[4]} \boldanswer{Fontenello Park} \xmark\par
{[5]} \boldanswer{Bubbling Springs Memorial Park} \xmark\par
{[6]} \boldanswer{Bemis Park.} \xmark\par
{[7]} \boldanswer{Fontenot Memorial Park} \xmark\par
{[8]} \boldanswer{Bemis Park} \xmark\par
{[9]} \boldanswer{Fontenberry Park} \xmark\par
{[10]} \boldanswer{Fontenelle Park} \xmark\par
{[11]} \boldanswer{Lake Manawa Memorial Park} \xmark\par
{[12]} {\rmfamily\textbf{The memorial park you are referring to is Elmwood Park.}} \xmark\par
{[13]} \boldanswer{Fontenberry Park} \xmark\par
{[14]} \boldanswer{Standing Bear Park} \xmark\par
{[15]} \boldanswer{Miller Park} \xmark
\metrics{0.000}{0.105}
\end{tcolorbox}

\begin{tcolorbox}[breakable,
  title={Base generated answer},
  colframe=BaseBlue,
  colbacktitle=BaseBlue,
  colback=BaseBlue!8,
  coltitle=white,
  fonttitle=\bfseries,
  boxrule=1pt,
  arc=2mm,
  left=8pt, right=8pt, top=4pt, bottom=4pt,
  before skip=3pt, after skip=3pt,
]
\ttfamily\raggedright\fontsize{10pt}{10.72pt}\selectfont
{[0]} \boldanswer{Fontenberry Park} \xmark\par
{[1]} \boldanswer{Fontenberry Park} \xmark\par
{[2]} \boldanswer{Fontenello Park} \xmark\par
{[3]} \boldanswer{Lake Manawa Park} \xmark\par
{[4]} \boldanswer{Fontenello Park} \xmark\par
{[5]} \boldanswer{Benson Memorial Park.} \xmark\par
{[6]} \boldanswer{Fontenberry Park.} \xmark\par
{[7]} \boldanswer{Fontenot Memorial Park} \xmark\par
{[8]} \boldanswer{Bemis Park} \xmark\par
{[9]} \boldanswer{Fontenberry Park} \xmark\par
{[10]} \boldanswer{Fontenello Memorial Park} \xmark\par
{[11]} \boldanswer{Lake Manawa Memorial Park} \xmark\par
{[12]} \boldanswer{Fontenot Park} \xmark\par
{[13]} \boldanswer{Fontenberry Park} \xmark\par
{[14]} \boldanswer{Fontenberry Memorial Park.} \xmark\par
{[15]} \boldanswer{Fontenello Park.} \xmark
\metrics{0.000}{0.206}
\end{tcolorbox}

\begin{tcolorbox}[breakable,
  title={RLVR generated answer},
  colframe=RLVRGreen,
  colbacktitle=RLVRGreen,
  colback=RLVRGreen!8,
  coltitle=white,
  fonttitle=\bfseries,
  boxrule=1pt,
  arc=2mm,
  left=8pt, right=8pt, top=4pt, bottom=4pt,
  before skip=3pt, after skip=3pt,
]
\ttfamily\raggedright\fontsize{10pt}{10.72pt}\selectfont
{[0]} \boldanswer{Fontenberry Park} \xmark\par
{[1]} \boldanswer{Fontenberry Park} \xmark\par
{[2]} \boldanswer{Fontenello Memorial Park} \xmark\par
{[3]} \boldanswer{Fontenberry Park.} \xmark\par
{[4]} \boldanswer{Fontenello Park} \xmark\par
{[5]} \boldanswer{Fontenot Park} \xmark\par
{[6]} \boldanswer{Fontenberry Park.} \xmark\par
{[7]} \boldanswer{Fontenello Memorial Park} \xmark\par
{[8]} \boldanswer{Fontenello Memorial Park} \xmark\par
{[9]} \boldanswer{Fontenberry Park} \xmark\par
{[10]} \boldanswer{Fontenello Memorial Park.} \xmark\par
{[11]} \boldanswer{Lake Manawa Memorial Park} \xmark\par
{[12]} \boldanswer{Fontenot Memorial Park.} \xmark\par
{[13]} \boldanswer{Fontenberry Park} \xmark\par
{[14]} \boldanswer{Fontenberry Memorial Park.} \xmark\par
{[15]} \boldanswer{Fontenello Park.} \xmark
\metrics{0.000}{0.313}
\end{tcolorbox}

\begin{tcolorbox}[breakable,
  title={RD generated answer},
  colframe=RDGold,
  colbacktitle=RDGold,
  colback=RDGold!8,
  coltitle=white,
  fonttitle=\bfseries,
  boxrule=1pt,
  arc=2mm,
  left=8pt, right=8pt, top=4pt, bottom=4pt,
  before skip=3pt, after skip=3pt,
]
\ttfamily\raggedright\fontsize{10pt}{10.72pt}\selectfont
{[0]} Answer: \boldanswer{Fontenello Park}, Confidence: 0 \xmark
\metrics{0.000}{0.000}
\end{tcolorbox}

\begin{tcolorbox}[breakable,
  title={RLCR generated answer},
  colframe=RLCRPurple,
  colbacktitle=RLCRPurple,
  colback=RLCRPurple!8,
  coltitle=white,
  fonttitle=\bfseries,
  boxrule=1pt,
  arc=2mm,
  left=8pt, right=8pt, top=4pt, bottom=4pt,
  before skip=3pt, after skip=3pt,
]
\ttfamily\raggedright\fontsize{10pt}{10.72pt}\selectfont
{[0]} \textless{}think\textgreater{} I'm thinking about the memorial park located at John J. Pershing Drive in North Omaha. I'm considering the amenities it might have, including a baseball park. I'm searching through my knowledge about parks in Omaha and their features. I'm looking for a match between the location and the amenities. I'm thinking about the possibility of the park being a well-known memorial park in Omaha. I'm recalling the names of parks in Omaha and their locations. I'm trying to find a connection between the park and the baseball field. \textless{}/think\textgreater{} \textless{}answer\textgreater{}\boldanswer{Fontenello Park}\textless{}/answer\textgreater{} \textless{}analysis\textgreater{} There are several uncertainties in my reasoning process. Firstly, I'm relying on my knowledge about Omaha's parks, which might not be exhaustive. Secondly, I'm assuming that the park I'm thinking of is a well-known memorial park, but there could be other parks with similar features. Thirdly, I'm not considering the possibility of the park being a smaller or lesser-known park that still has a baseball field. Lastly, I'm not accounting for the possibility of the park being renamed or redeveloped since my knowledge cutoff. \textless{}/analysis\textgreater{} \textless{}confidence\textgreater{} \boldanswer{0.52} \textless{}/confidence\textgreater{} \xmark
\metrics{0.000}{0.520}
\end{tcolorbox}

\subsubsection{NQ-Open (out-of-distribution)}
\label{app:vis_dispersed_nqopen}
\begin{tcolorbox}[breakable,
  title=Question,
  colback=gray!10,
  colframe=black,
  colbacktitle=black,
  coltitle=white,
  fonttitle=\bfseries,
  boxrule=1pt,
  arc=2mm,
  left=8pt, right=8pt, top=4pt, bottom=4pt,
  before skip=8pt, after skip=8pt,
]
who said one man's vulgarity is another's lyric\\
\textbf{Ground Truth Answer:} Justice Harlan
\end{tcolorbox}

Out of distribution the gap widens. RLVR returns the same wrong name in all sixteen rollouts, forms one cluster, and so reports $1.000$. CSR spreads over eleven clusters and reports $0.113$, so its confidence exposes the error.

\begin{tcolorbox}[breakable,
  title={CSR (Ours) generated answer},
  colframe=CSRRed,
  colbacktitle=CSRRed,
  colback=CSRRed!8,
  coltitle=white,
  fonttitle=\bfseries,
  boxrule=1pt,
  arc=2mm,
  left=8pt, right=8pt, top=4pt, bottom=4pt,
  before skip=8pt, after skip=3pt,
]
\ttfamily\raggedright\fontsize{10pt}{10.72pt}\selectfont
{[0]} \boldanswer{George Carlin} \xmark\par
{[1]} {\rmfamily\textbf{This quote is attributed to Robert A. Heinlein, an American science fiction author.}} \xmark\par
{[2]} \boldanswer{Lillian Hellman} \xmark\par
{[3]} \boldanswer{E.B. White} \xmark\par
{[4]} \boldanswer{Alfred Alvarez} \xmark\par
{[5]} \boldanswer{Alexander Woollcott} \xmark\par
{[6]} \boldanswer{George Carlin} \xmark\par
{[7]} \boldanswer{George Carlin} \xmark\par
{[8]} \boldanswer{George Carlin} \xmark\par
{[9]} \boldanswer{H.L. Mencken} \xmark\par
{[10]} \boldanswer{Oscar Wilde} \xmark\par
{[11]} \boldanswer{George Carlin} \xmark\par
{[12]} \boldanswer{Aldous Huxley} \xmark\par
{[13]} \boldanswer{Justice Oliver Wendell Holmes Jr.} \xmark\par
{[14]} \boldanswer{H.L. Mencken} \xmark\par
{[15]} \boldanswer{Frederic Raphael} \xmark
\metrics{0.000}{0.113}
\end{tcolorbox}

\begin{tcolorbox}[breakable,
  title={Base generated answer},
  colframe=BaseBlue,
  colbacktitle=BaseBlue,
  colback=BaseBlue!8,
  coltitle=white,
  fonttitle=\bfseries,
  boxrule=1pt,
  arc=2mm,
  left=8pt, right=8pt, top=4pt, bottom=4pt,
  before skip=3pt, after skip=3pt,
]
\ttfamily\raggedright\fontsize{10pt}{10.72pt}\selectfont
{[0]} \boldanswer{George Carlin} \xmark\par
{[1]} {\rmfamily\textbf{This quote is often attributed to Oscar Wilde.}} \xmark\par
{[2]} \boldanswer{George Carlin.} \xmark\par
{[3]} \boldanswer{George Carlin.} \xmark\par
{[4]} \boldanswer{Oscar Wilde} \xmark\par
{[5]} \boldanswer{Lawrence J. Peter.} \xmark\par
{[6]} \boldanswer{George Carlin.} \xmark\par
{[7]} \boldanswer{George Carlin} \xmark\par
{[8]} \boldanswer{George Carlin} \xmark\par
{[9]} \boldanswer{H.L. Mencken} \xmark\par
{[10]} \boldanswer{Oscar Wilde} \xmark\par
{[11]} \boldanswer{George Carlin.} \xmark\par
{[12]} \boldanswer{George Carlin} \xmark\par
{[13]} \boldanswer{George Carlin} \xmark\par
{[14]} \boldanswer{H.L. Mencken.} \xmark\par
{[15]} \boldanswer{George Carlin} \xmark
\metrics{0.000}{0.353}
\end{tcolorbox}

\begin{tcolorbox}[breakable,
  title={RLVR generated answer},
  colframe=RLVRGreen,
  colbacktitle=RLVRGreen,
  colback=RLVRGreen!8,
  coltitle=white,
  fonttitle=\bfseries,
  boxrule=1pt,
  arc=2mm,
  left=8pt, right=8pt, top=4pt, bottom=4pt,
  before skip=3pt, after skip=3pt,
]
\ttfamily\raggedright\fontsize{10pt}{10.72pt}\selectfont
{[0]} \boldanswer{George Carlin} \xmark\par
{[1]} \boldanswer{George Carlin} \xmark\par
{[2]} \boldanswer{George Carlin.} \xmark\par
{[3]} \boldanswer{George Carlin.} \xmark\par
{[4]} \boldanswer{George Carlin} \xmark\par
{[5]} \boldanswer{George Carlin} \xmark\par
{[6]} \boldanswer{George Carlin.} \xmark\par
{[7]} \boldanswer{George Carlin} \xmark\par
{[8]} \boldanswer{George Carlin} \xmark\par
{[9]} \boldanswer{George Carlin} \xmark\par
{[10]} \boldanswer{George Carlin} \xmark\par
{[11]} \boldanswer{George Carlin.} \xmark\par
{[12]} \boldanswer{George Carlin} \xmark\par
{[13]} \boldanswer{George Carlin} \xmark\par
{[14]} \boldanswer{George Carlin} \xmark\par
{[15]} \boldanswer{George Carlin} \xmark
\metrics{0.000}{1.000}
\end{tcolorbox}

\begin{tcolorbox}[breakable,
  title={RD generated answer},
  colframe=RDGold,
  colbacktitle=RDGold,
  colback=RDGold!8,
  coltitle=white,
  fonttitle=\bfseries,
  boxrule=1pt,
  arc=2mm,
  left=8pt, right=8pt, top=4pt, bottom=4pt,
  before skip=3pt, after skip=3pt,
]
\ttfamily\raggedright\fontsize{10pt}{10.72pt}\selectfont
{[0]} Answer: \boldanswer{George Carlin}, Confidence: 7 \xmark
\metrics{0.000}{0.700}
\end{tcolorbox}

\begin{tcolorbox}[breakable,
  title={RLCR generated answer},
  colframe=RLCRPurple,
  colbacktitle=RLCRPurple,
  colback=RLCRPurple!8,
  coltitle=white,
  fonttitle=\bfseries,
  boxrule=1pt,
  arc=2mm,
  left=8pt, right=8pt, top=4pt, bottom=4pt,
  before skip=3pt, after skip=3pt,
]
\ttfamily\raggedright\fontsize{10pt}{10.72pt}\selectfont
{[0]} \textless{}think\textgreater{} I'm thinking about the origin of this quote, which seems to be a phrase that highlights the subjective nature of what is considered ``vulgar'' or ``lyric.'' It appears to be a quote from an American author, and I'm trying to recall the exact words and context in which it was said. I'm considering the possible sources of this quote, such as literature, music, or even a famous speech. I'm also thinking about the possible authors who might have coined this phrase, such as Mark Twain or H.L. Mencken. I'm weighing the likelihood of this quote being a common saying or a unique phrase from a specific work. \textless{}/think\textgreater{} \textless{}answer\textgreater{}\boldanswer{Oscar Wilde}\textless{}/answer\textgreater{} \textless{}analysis\textgreater{} One uncertainty is that I'm relying on my training data to identify the quote, and there's a possibility that the quote might be misattributed or not accurately represented in my training data. Another uncertainty is that there might be multiple authors who have used similar phrases, making it difficult to pinpoint the exact origin of this quote. Additionally, the quote's meaning and context could be open to interpretation, leading to different perspectives on what is considered ``vulgar'' or ``lyric.'' \textless{}/analysis\textgreater{} \textless{}confidence\textgreater{} \boldanswer{0.62} \textless{}/confidence\textgreater{} \xmark
\metrics{0.000}{0.620}
\end{tcolorbox}

\newpage
\section*{NeurIPS Paper Checklist}

\begin{enumerate}

\item {\bf Claims}
    \item[] Question: Do the main claims made in the abstract and introduction accurately reflect the paper's contributions and scope?
    \item[] Answer: \answerYes{} 
    \item[] Justification: The abstract and introduction state the claims, which are supported by the results in Section~\ref{sec:main_results} and Appendix~\ref{app:supp_analyses}.

\item {\bf Limitations}
    \item[] Question: Does the paper discuss the limitations of the work performed by the authors?
    \item[] Answer: \answerYes{} 
    \item[] Justification: Section~\ref{sec:conclusion} discusses the limitations.

\item {\bf Theory assumptions and proofs}
    \item[] Question: For each theoretical result, does the paper provide the full set of assumptions and a complete (and correct) proof?
    \item[] Answer: \answerYes{} 
    \item[] Justification: We give the proof of Proposition~\ref{prop:meanfield} in Appendix~\ref{app:meanfield_proof}.

    \item {\bf Experimental result reproducibility}
    \item[] Question: Does the paper fully disclose all the information needed to reproduce the main experimental results of the paper to the extent that it affects the main claims and/or conclusions of the paper (regardless of whether the code and data are provided or not)?
    \item[] Answer: \answerYes{} 
    \item[] Justification: We provide complete experiment setup, training and evaluation configurations in Appendix~\ref{app:exp_details}.

\newpage
\item {\bf Open access to data and code}
    \item[] Question: Does the paper provide open access to the data and code, with sufficient instructions to faithfully reproduce the main experimental results, as described in supplemental material?
    \item[] Answer: \answerYes{} 
    \item[] Justification: We use publicly available datasets and provide code at \url{https://github.com/Rose-STL-Lab/Semantic_Calibration}. The repository includes the implementation, evaluation scripts, and configuration files needed to reproduce the main experimental results.

\item {\bf Experimental setting/details}
    \item[] Question: Does the paper specify all the training and test details (e.g., data splits, hyperparameters, how they were chosen, type of optimizer) necessary to understand the results?
    \item[] Answer: \answerYes{} 
    \item[] Justification: Section~\ref{sec:exp_setup} summarizes the setup, and Appendix~\ref{app:exp_setup} gives the full training and evaluation details.

\item {\bf Experiment statistical significance}
    \item[] Question: Does the paper report error bars suitably and correctly defined or other appropriate information about the statistical significance of the experiments?
    \item[] Answer: \answerNo{} 
    \item[] Justification: We do not report error bars because repeating every RL training run across three backbones and four trained methods is computationally expensive (Appendix~\ref{app:efficient_sc}). Base, RLVR, and CSR are evaluated on $16$ sampled rollouts per question, while RD and RLCR use a single greedy answer.

\item {\bf Experiments compute resources}
    \item[] Question: For each experiment, does the paper provide sufficient information on the computer resources (type of compute workers, memory, time of execution) needed to reproduce the experiments?
    \item[] Answer: \answerYes{} 
    \item[] Justification: Every training and evaluation run uses one NVIDIA A100 80GB GPU (Appendix~\ref{app:exp_setup}). Depending on the backbone, a full training run on HotpotQA takes about 9 to 32 GPU-hours for CSR, 7 to 21 for RLVR, 8 to 11 for RD, and 18 to 37 for RLCR.    
    
\item {\bf Code of ethics}
    \item[] Question: Does the research conducted in the paper conform, in every respect, with the NeurIPS Code of Ethics \url{https://neurips.cc/public/EthicsGuidelines}?
    \item[] Answer: \answerYes{} 
    \item[] Justification: This paper adheres to the NeurIPS Code of Ethics.

\item {\bf Broader impacts}
    \item[] Question: Does the paper discuss both potential positive societal impacts and negative societal impacts of the work performed?
    \item[] Answer: \answerYes{} 
    \item[] Justification: Better-calibrated confidence can help users decide when to trust a model's answer, for example in medical or legal question answering. A negative impact is over-reliance: a calibrated score can still be wrong, especially out of distribution (e.g., MSMARCO in Table~\ref{tab:main_results}), so it should not be treated as a guarantee in high-stakes settings.
    
\item {\bf Safeguards}
    \item[] Question: Does the paper describe safeguards that have been put in place for responsible release of data or models that have a high risk for misuse (e.g., pre-trained language models, image generators, or scraped datasets)?
    \item[] Answer: \answerNA{} 
    \item[] Justification: This work uses publicly available datasets and pre-trained models.

\item {\bf Licenses for existing assets}
    \item[] Question: Are the creators or original owners of assets (e.g., code, data, models), used in the paper, properly credited and are the license and terms of use explicitly mentioned and properly respected?
    \item[] Answer: \answerYes{} 
    \item[] Justification: We cite the source of every asset. Datasets: HotpotQA (CC BY-SA 4.0), TriviaQA (Apache 2.0), MS MARCO (non-commercial research terms), Natural Questions/NQ-Open (CC BY-SA 3.0), MuSiQue (CC BY 4.0), and 2WikiMultiHopQA (Apache 2.0). Models: Llama-3.1-8B-Instruct (Llama 3.1 Community License), and Qwen2.5-7B-Instruct, Qwen3-14B, Qwen3-32B, and Mistral-7B-Instruct-v0.3 (Apache 2.0). GPT-4.1 nano is used under the OpenAI API terms. Baseline code: RLCR (MIT) and Rewarding Doubt (public repository, no license specified). Libraries: TRL, Unsloth, and vLLM (Apache 2.0).

\item {\bf New assets}
    \item[] Question: Are new assets introduced in the paper well documented and is the documentation provided alongside the assets?
    \item[] Answer: \answerYes{} 
    \item[] Justification: We release our code under the MIT license at \url{https://github.com/Rose-STL-Lab/Semantic_Calibration}, with a README documenting installation, training, evaluation, and the configurations used in the paper.

\item {\bf Crowdsourcing and research with human subjects}
    \item[] Question: For crowdsourcing experiments and research with human subjects, does the paper include the full text of instructions given to participants and screenshots, if applicable, as well as details about compensation (if any)? 
    \item[] Answer: \answerNA{} 
    \item[] Justification: This paper does not involve crowdsourcing nor research with human subjects.

\item {\bf Institutional review board (IRB) approvals or equivalent for research with human subjects}
    \item[] Question: Does the paper describe potential risks incurred by study participants, whether such risks were disclosed to the subjects, and whether Institutional Review Board (IRB) approvals (or an equivalent approval/review based on the requirements of your country or institution) were obtained?
    \item[] Answer: \answerNA{} 
    \item[] Justification: This paper does not involve crowdsourcing nor research with human subjects.

\item {\bf Declaration of LLM usage}
    \item[] Question: Does the paper describe the usage of LLMs if it is an important, original, or non-standard component of the core methods in this research? Note that if the LLM is used only for writing, editing, or formatting purposes and does \emph{not} impact the core methodology, scientific rigor, or originality of the research, declaration is not required.
    \item[] Answer: \answerYes{} 
    \item[] Justification: An LLM (GPT-4.1 nano) is a core component of the method: it judges the correctness of each rollout and the semantic equivalence of rollout pairs, both in the training reward and in the confidence estimate (Section~\ref{sec:exp_setup}, Appendices~\ref{app:metrics} and~\ref{app:system_prompt}).

\end{enumerate}

\end{document}